%% file: main.tex
\documentclass[12pt]{report}
\usepackage[utf8]{inputenc}
\usepackage{a4}
\usepackage[none]{hyphenat} 
\usepackage{parskip} 
\usepackage{afterpage} 
\usepackage{makeidx}
\usepackage[numbers]{natbib}
\usepackage{graphicx}
\usepackage{picins} 
\usepackage{siunitx} 

\usepackage{url} 
\usepackage{setspace}
\usepackage{titlesec}
\usepackage{dsfont} 
\usepackage{tabularx}
\usepackage{floatflt} 

\usepackage{enumitem} 
\usepackage{amsmath,amssymb}
\usepackage[format=default,font=footnotesize,labelfont=bf]{caption}
\usepackage{amsthm} 
\usepackage{listings} 
\usepackage{color}
\usepackage{algpseudocode} 
\usepackage{algorithm} 
\usepackage{epigraph}
\usepackage{float} 
\usepackage{placeins}
\usepackage[table]{xcolor} 
\usepackage{multirow} 

\usepackage{todonotes}
\usepackage{booktabs} 
\usepackage{tikz}
\usetikzlibrary{arrows.meta}
\usepackage[edges]{forest}

\forestset{
  mytree/.style={
    for tree={
      base=bottom,
      grow=south,
      child anchor=north,
      align=center,
      font=\Huge,                    
      l sep+=1.5cm,                   
      s sep+=0.4cm,                   
      edge+={thick, -{Latex}},        
      edge label+/.style={font=\Huge}, 
      straight edge/.style={
        edge path={
          \noexpand\path[\forestoption{edge}] 
          (!u.parent anchor) -- (.child anchor);
        }
      },
      if n children=0{
        tier=word, draw, thick, rectangle
      }{
        draw, diamond, thick, aspect=2.2
      },
    },
  },
}

\usepackage{afterpage}
\usepackage{pdflscape}

\usepackage{adjustbox}
\usepackage{rotating}

\DeclareMathOperator{\doo}{do}

\usepackage{sectsty}
\allsectionsfont{\raggedright}

\usetikzlibrary{arrows,shapes.arrows,shapes.geometric,
shapes.multipart,decorations.pathmorphing,positioning,
swigs}

\titleformat{\paragraph}[hang]{\normalfont\bfseries}{\theparagraph}{.5em}{}

\makeindex
\input{figures/tumlogo}

\begin{document}

\hoffset=5mm
\thispagestyle{empty}

\begin{center}
	\bigskip \bigskip \bigskip 
	\oTUM{6.0cm} \\
	\vspace*{0.8cm}
	{\huge \bf Technische Universität} \\
	\bigskip
	{\huge \bf München} \\
	\bigskip \bigskip \bigskip
	{\huge \bf Fakultät für Informatik} \\
	\bigskip \bigskip \bigskip
	{\Large \bf Bachelor's Thesis in Information Systems} \\
	\bigskip \bigskip \bigskip \bigskip \bigskip
	{\Large Bounding Causal Effects and Counterfactuals} \\        
	\bigskip \bigskip \bigskip \bigskip
	{\Large Tobias Anton Maringgele} \\    
	\bigskip
	\begin{figure}[ht]
	\centering \includegraphics[width=0.2\linewidth]{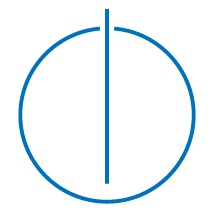}
	\end{figure}
	\bigskip 
\end{center}

\vfill

\newpage
\hoffset=5mm
\thispagestyle{empty}

\begin{center}
	\bigskip \bigskip \bigskip 
	\oTUM{6.0cm} \\
	\vspace*{0.8cm}
	{\huge \bf Technical University of} \\
	\bigskip
	{\huge \bf Munich} \\
	\bigskip \bigskip \bigskip
	{\huge \bf Department of Informatics} \\
	\bigskip \bigskip \bigskip
	{\Large \bf Bachelor's Thesis in Information Systems} \\
	\bigskip \bigskip \bigskip \bigskip \bigskip
	{\Large Bounding Causal Effects and Counterfactuals} \\
	\bigskip \bigskip \bigskip
    {\Large  Schranken kausaler und kontrafaktischer Effekte} \\     
	\bigskip
\end{center}
\vfill

\begin{tabular}{ll}
{\Large \bf Author:} & {\Large Tobias Anton Maringgele} \\\\
{\Large \bf Supervisor:} & {\Large Prof. Dr. Jalal Etesami} \\\\
{\Large \bf Advisor:} & {\Large Prof. Dr. Jalal Etesami} \\\\
{\Large \bf Submission:} & {\Large 18.08.2025}
\end{tabular}

\newpage	
\thispagestyle{empty}
\hoffset=0mm
\vspace*{\fill}
\noindent I assure the single handed composition of this bachelor's thesis only supported by declared resources.\\\\
Munich, 18.08.2025
\begin{figure}[H]
  \includegraphics[width=0.2\textwidth]{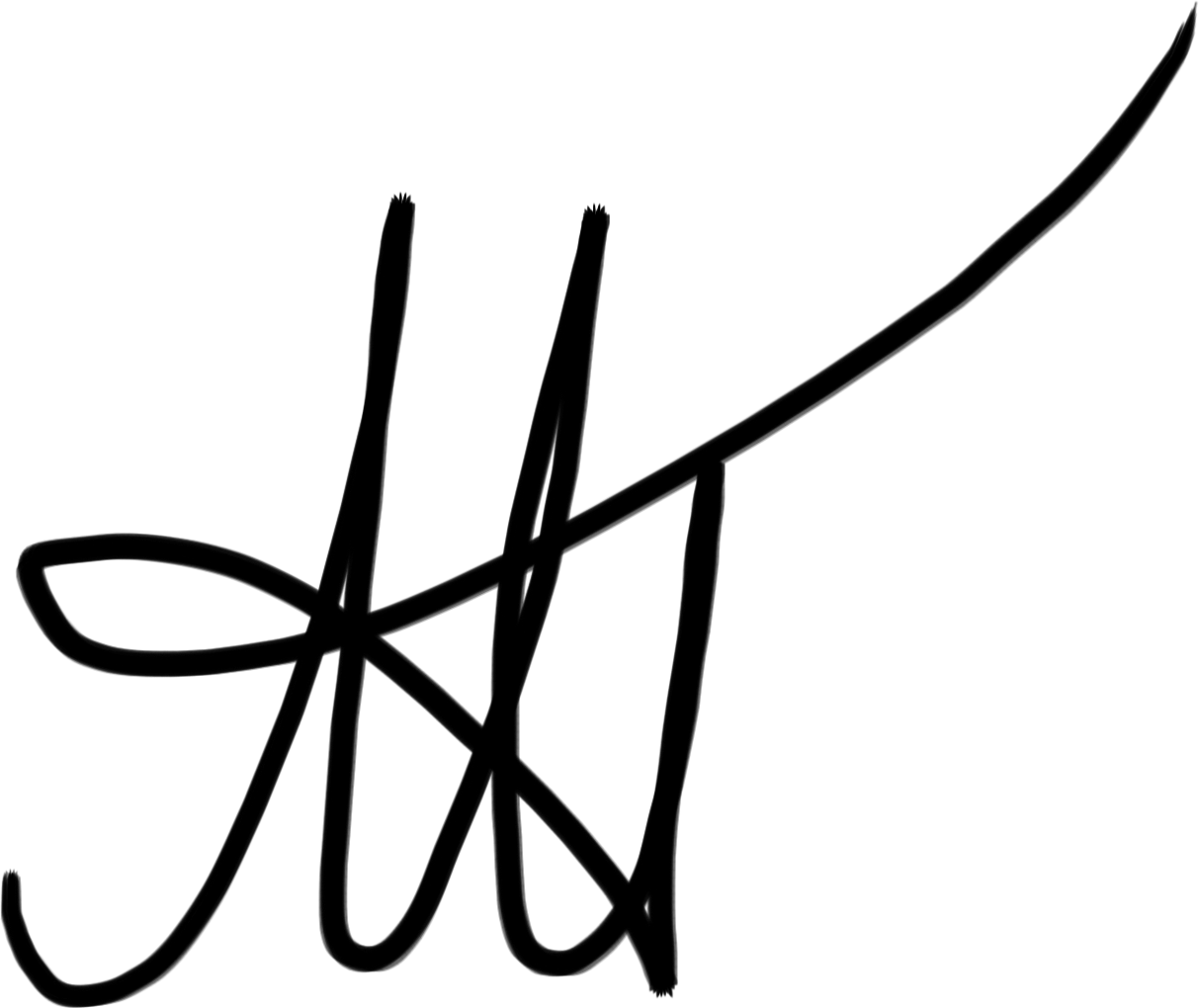}
\end{figure}
\noindent \textit{(Tobias Anton Maringgele)}

\newpage
\thispagestyle{empty}
\null

\newpage
\thispagestyle{empty}
\hoffset=0mm
\section*{Abstract}	
\begin{spacing}{1.2}
\input{abstracte}
\end{spacing}

\newpage
\thispagestyle{empty}
\hoffset=0mm
\section*{Inhaltsangabe}
\begin{spacing}{1.2}
\input{abstractd}
\end{spacing}

\newpage
\setcounter{page}{1}
\hoffset=0mm
\bibliographystyle{wmaainf} 
\setcounter{tocdepth}{3}
\setcounter{secnumdepth}{3}
\fboxsep 0mm

\tableofcontents

\newpage
\setlength{\baselineskip}{3ex}

\begin{spacing}{1.15}
        \include{chapters/ch-introduction}

        \include{chapters/ch-problem_setting}
        \include{chapters/ch-methods}
        \include{chapters/ch-simulation_and_experiment}

        \include{chapters/ch-discussion}
\end{spacing}
\newpage
\thispagestyle{empty}
\null

\newpage
\addcontentsline{toc}{chapter}{List of figures}
\listoffigures

\input{chapters/appendix}

\newpage
\thispagestyle{empty}
\null

\bibliography{references}

\end{document}

%% file: figures/tumlogo.tex
\def\TUM#1{%
\dimen1=#1\dimen1=.1143\dimen1%
\dimen2=#1\dimen2=.419\dimen2%
\dimen3=#1\dimen3=.0857\dimen3%
\dimen4=\dimen1\advance\dimen4 by\dimen2%
\setbox0=\vbox{\hrule width\dimen3 height\dimen1 depth0pt\vskip\dimen2}%
\setbox1=\vbox{\hrule width\dimen1 height\dimen4 depth0pt}%
\setbox2=\vbox{\hrule width\dimen3 height\dimen1 depth0pt}%
\setbox3=\hbox{\copy0\copy1\copy0\copy1\box2\copy1\copy0\copy1\box0\box1}%
\leavevmode\vbox{\box3}}
\def\oTUM#1{%
\dimen1=#1\dimen1=.1143\dimen1%
\dimen2=#1\dimen2=.419\dimen2%
\dimen3=#1\dimen3=.0857\dimen3%
\dimen0=#1\dimen0=.018\dimen0%
\dimen4=\dimen1\advance\dimen4 by-\dimen0%
\setbox1=\vbox{\hrule width\dimen0 height\dimen4 depth0pt}%
\advance\dimen4 by\dimen2%
\setbox8=\vbox{\hrule width\dimen0 height\dimen4 depth0pt}%
\advance\dimen4 by-\dimen2\advance\dimen4 by-\dimen0%
\setbox4=\vbox{\hrule width\dimen4 height\dimen0 depth0pt}%
\advance\dimen4 by\dimen1\advance\dimen4 by\dimen3%
\setbox6=\vbox{\hrule width\dimen4 height\dimen0 depth0pt}%
\advance\dimen4 by\dimen3\advance\dimen4 by\dimen0%
\setbox9=\vbox{\hrule width\dimen4 height\dimen0 depth0pt}%
\advance\dimen4 by\dimen1%
\setbox7=\vbox{\hrule width\dimen4 height\dimen0 depth0pt}%
\dimen4=\dimen3%
\setbox5=\vbox{\hrule width\dimen4 height\dimen0 depth0pt}%
\advance\dimen4 by-\dimen0%
\setbox2=\vbox{\hrule width\dimen4 height\dimen0 depth0pt}%
\dimen4=\dimen2\advance\dimen4 by\dimen0%
\setbox3=\vbox{\hrule width\dimen0 height\dimen4 depth0pt}%
\setbox0=\vbox{\hbox{\box9\lower\dimen2\copy3\lower\dimen2\copy5%
\lower\dimen2\copy3\box7}\kern-\dimen2\nointerlineskip%
\hbox{\raise\dimen2\box1\raise\dimen2\box2\copy3\copy4\copy3%
\raise\dimen2\copy5\copy3\box6\copy3\raise\dimen2\copy5\copy3\copy4\copy3%
\raise\dimen2\box5\box3\box4\box8}}%
\leavevmode\box0}

%% file: abstracte.tex
Causal inference often hinges on strong assumptions—such as no unmeasured confounding or perfect compliance—that are rarely satisfied in practice. Partial identification offers a principled alternative: instead of relying on unverifiable assumptions to estimate causal effects precisely, it derives bounds that reflect the uncertainty inherent in the data. Despite its theoretical appeal, partial identification remains underutilized in applied work, in part due to the fragmented nature of existing methods and the lack of practical guidance.

This thesis addresses these challenges by systematically comparing a diverse set of bounding algorithms across multiple causal scenarios. We implement, extend, and unify state-of-the-art methods—including symbolic, optimization-based, and information-theoretic approaches—within a common evaluation framework. In particular, we propose an extension of the entropy-bounded method by Jiang et al.~\cite{jiang_approximate_2023}, making it applicable to counterfactual queries such as the Probability of Necessity and Sufficiency (PNS).

Our empirical study spans thousands of randomized simulations involving both discrete and continuous data-generating processes. We assess each method in terms of bound tightness, computational efficiency, and robustness to assumption violations. To support practitioners, we distill our findings into a practical decision tree for algorithm selection and train a machine learning model in an attempt to predict the best-performing method based on observable data characteristics.

All implementations are released as part of an open-source Python package, \texttt{CausalBoundingEngine}, which enables users to apply and compare bounding methods through a unified interface. We hope this work contributes to the growing toolkit for causal inference under uncertainty and lowers the barrier to using partial identification in empirical research.

%% file: abstractd.tex
Kausale Inferenz beruht häufig auf starken Annahmen – etwa dem Fehlen unbeobachteter Einflussfaktoren oder der perfekten Durchführung von Experimenten –, welche in der Praxis selten erfüllt sind. Partielle Identifikation bietet hierfür eine methodisch saubere Alternative: Anstatt sich auf schwer überprüfbare Voraussetzungen zu stützen, liefert sie Intervallschätzungen, welche die inhärente Unsicherheit der Daten explizit abbilden. Trotz ihrer theoretischen Relevanz findet sie bislang kaum praktische Anwendung – nicht zuletzt aufgrund der Vielzahl verstreuter Methoden und fehlender Orientierungshilfen.

Diese Arbeit begegnet diesem Defizit durch einen systematischen Vergleich verschiedener Verfahren zur Schrankenbestimmung kausaler Effekte über eine Bandbreite an Szenarien hinweg. Wir implementieren, erweitern und vereinheitlichen moderne Ansätze – darunter analytische, optimierungsbasierte und informationstheoretische Methoden – in einem gemeinsamen Rahmen zur Auswertung. Besonders hervorzuheben ist unsere Erweiterung der von Jiang et al.~\cite{jiang_approximate_2023} vorgeschlagenen entropiebasierten Methode, die nun auch auf kontrafaktische Fragestellungen wie die Wahrscheinlichkeit von Notwendigkeit und Hinreichendheit (PNS) anwendbar ist.

Unsere empirische Untersuchung basiert auf tausenden zufallsbasierten Simulationen mit sowohl diskreten als auch kontinuierlichen Datengenerierungsprozessen. Für jedes Verfahren analysieren wir die Enge der resultierenden Schranken, den Rechenaufwand sowie die Robustheit gegenüber Modellverletzungen. Zur Unterstützung praktischer Anwendungen leiten wir aus den Ergebnissen einen Entscheidungsbaum zur Methodenwahl ab und entwickeln ein Vorhersagemodell, das anhand beobachtbarer Merkmale – wie Entropie oder wechselseitiger Information – die jeweils geeignetste Methode bestimmt.

Die Verfahren wurden im Rahmen eines quelloffenen Softwarepakets (\texttt{CausalBoundingEngine}) implementiert, das eine standardisierte Schnittstelle für Anwendung und Vergleich bereitstellt. Diese Arbeit verfolgt das Ziel, kausale Inferenz unter Unsicherheit methodisch zu stärken und die partielle Identifikation als praktikables Instrument in der empirischen Forschung weiter zu etablieren.

%% file: chapters/ch-introduction.tex
\chapter{Introduction}
\label{ch-introduction}
    
\epigraph{Before something is being used, it is not being used.}{An example of a tautology. \\Jakob Zeitler}

\section{Motivation and the Case for Partial Identification}

Modern empirical research often aspires to precise causal estimation, seeking a single ``true'' effect size for a treatment or policy. However, real-world data limitations---such as unmeasured confounders, selection biases, and missing observations---pose fundamental identification challenges. In practice, obtaining a point estimate of a causal effect typically requires making strong, untestable assumptions (e.g., assuming no hidden bias or perfectly measured variables). If those assumptions are wrong, the point estimate can be misleading or even completely invalid. 

An alternative approach is to acknowledge the limits of the data and embrace uncertainty in our conclusions. Rather than demanding a precise but potentially unreliable estimate, we can ask: what range of effect sizes is consistent with the data under weaker assumptions? This approach is known as \emph{partial identification}, and it can yield meaningful insights without relying on untenable assumptions~\cite{mullahy_embracing_2021}. In many cases, accepting a bit more uncertainty in the results is preferable to basing ``precise'' conclusions on heroic assumptions that may not hold~\cite{mullahy_embracing_2021}. As Charles Manski famously observed, there is a trade-off between informativeness and credibility in inference---the more we narrow an estimate via assumptions, the more we risk believing something \emph{precisely wrong} rather than \emph{roughly right}~\cite{manski_identification_2007, manski_public_2013}. Partial identification operates on the more credible end of this trade-off, deliberately using minimal assumptions to avoid overconfidence.

It does not attempt to pinpoint a singular value for a causal effect; instead, partial identification determines a bounded interval (or set) of plausible values for that effect, given the information available. The width of this interval reflects the amount of ambiguity in the data after accounting for only the most defensible assumptions. If we are willing to assume more (for example, assume no unmeasured confounding or monotonic responses), the interval may shrink---potentially collapsing to a point if the assumptions suffice for point identification. 

But crucially, even when we cannot credibly make such strong assumptions, partial identification still allows us to learn something: it tells us how large or small the effect could be, in the best and worst cases consistent with the data. This style of analysis makes the assumptions \emph{transparent}---we can start with very weak assumptions (yielding perhaps wide bounds) and then see how the bounds tighten as we explore slightly stronger assumptions. 

In contrast, the traditional point-estimation approach often hides its assumptions behind a veneer of certainty, reporting a single number (and maybe a narrow confidence interval) that assumes a fully identified model. Partial identification flips this script, prioritizing \emph{robustness} of conclusions over narrowness. As a result, even somewhat wide bounds are informative: they may reveal, for instance, that an effect could be positive or negative (sign-uncertain) or, alternatively, that the effect is definitely positive but could be anywhere from small to fairly large. Policymakers and scientists can use such results to understand best- and worst-case scenarios, and to decide if a conclusion is actionable or if more data/assumptions are needed~\cite{mullahy_embracing_2021}. 

In summary, partial identification broadens the analytical perspective by focusing on what can be learned with fewer leaps of faith, rather than insisting on a precise answer that might rest on a shaky foundation~\cite{manski_public_2013}.

\section{Bounding Counterfactuals}
While partial identification of causal effects provides robustness, its real power is most evident when applied to richer, counterfactual queries. The \emph{Probability of Necessity and Sufficiency} (PNS) is one such query that can reveal insights hidden from the Average Treatment Effect (ATE).

To illustrate this, consider a hypothetical scenario in which a new educational program is tested and the ATE is found to be essentially zero---suggesting that, on average, the program did not change student performance. Based on this alone, one might conclude the program had no effect and consider scrapping it. However, a partial identification analysis of a counterfactual quantity like the \emph{Probability of Necessity and Sufficiency} (PNS) could tell a different story: even with an average effect close to zero, it may be that a substantial fraction of students actually benefited from the program while others were harmed, resulting in the overall effect canceling out. For instance, the bounds on the PNS might indicate that between 12\% and 23\% of individuals experienced a positive effect due to the intervention. 

This kind of nuanced insight is invisible to the ATE---the average hides the distribution of individual-level responses. By contrast, bounding a counterfactual quantity such as the PNS reveals the possibility that a non-negligible subgroup was helped by the treatment, which could justify refining or targeting the program rather than discarding it entirely. Thus, even though PNS cannot be precisely estimated without strong assumptions, its partially identified bounds can provide valuable information for decision-makers.

In summary, partial identification methods can shine where traditional analyses might mislead---they can confirm whether an effect is possibly large for some even when it is zero on average, helping avoid false dismissals of potentially impactful treatments.

\section{Historical Background and Methodological Landscape}
Partial identification as a formal framework is not new---its intellectual origins trace back at least to the early 20\textsuperscript{th} century in statistics and economics~\cite{ho_partial_2015}. Pioneering work in the late 1980s and 1990s by Charles Manski and others helped unify and popularize partial identification methods for various problems, from estimating treatment effects with missing data to analyzing discrete choice models under minimal assumptions~\cite{manski_nonparametric_1990,  manski_identification_2007, ho_partial_2015}.

Over the past few decades, a rich literature has developed, and many new bounding techniques have been proposed in recent years across different disciplines. For example, researchers have derived sharp bounds for causal effects under instrumental variable models with imperfect instruments~\cite{balke_bounds_1997}, bounds for counterfactual probabilities like PNS~\cite{tian_probabilities_2000, zaffalon_bounding_2022}, and bounds in settings with a continuous outcome variable~\cite{zhang_bounding_2021}.

As a result, we now have a \emph{toolbox} of partial identification methods that can be applied to a wide range of causal inference scenarios. Manski's comprehensive reviews and books~\cite{manski_identification_2007, manski_public_2013} catalogue much of this development, and the ongoing surge of research (spanning fields like econometrics, epidemiology, and machine learning) continues to expand the toolkit.

\section{Barriers to Adoption}

Paradoxically, despite these methodological advances, partial identification techniques have seen limited uptake in practice outside of certain research communities. In fields like medicine and public health, for instance, the basic ideas of partial identification have only rarely been applied until very recently~\cite{mullahy_embracing_2021}. One reason is that many analysts default to well-known point-identification strategies, perhaps unaware that partial identification is an option when standard assumptions are violated.

But there are also more practical barriers. First, with so many new bounding methods developed in the abstract, there is a lack of empirical guidance about when which method works best. The literature currently contains few head-to-head comparisons of different partial identification approaches on the same problem, making it hard for practitioners to know which method would be most informative or efficient in their particular setting. Each method often comes with its own theoretical assumptions and mathematical complexity, and its performance can depend on context (e.g.\ some bounds might be tighter in one scenario, while another method yields tighter bounds elsewhere). Without comparative studies or empirical benchmarks, an applied researcher might be hesitant to try partial identification at all, not knowing whether it will give useful results or which approach to choose.

Second, and perhaps even more crucially, many of these methods lack accessible software implementations. Unlike classical techniques (where one can run a regression or use a statistical package), partial identification methods often require custom coding and significant effort. Only recently have some researchers begun to release code for specific methods~\cite{ho_partial_2015, jonzon_accessible_2024}, and even then, the code might be specialized for a particular scenario or method, or not well-documented for general users.

In short, the current landscape is fragmented: there's a plethora of theoretical results, but they are not consolidated into easy-to-use tools. This combination of little practical guidance and limited tooling likely contributes to why practitioners seldom use partial identification as part of their standard analytical toolkit~\cite{mullahy_embracing_2021, ho_partial_2015}. It is arguably a classic case of a powerful idea remaining on the sidelines because the barriers to entry are too high for non-specialists.

\section{Contributions}

This thesis is an effort to bridge the gap between theoretical developments in partial identification and their practical application. We aim to make bounding methods more approachable and actionable for researchers and decision-makers by consolidating state-of-the-art algorithms and offering clear empirical guidance on their use. By doing so, we hope to lower the barriers that have so far limited the adoption of partial identification in causal inference.

\paragraph{Methodological Integration.}
We collect, implement, and standardize a suite of bounding algorithms for estimating the Average Treatment Effect (ATE) and the Probability of Necessity and Sufficiency (PNS). These include classical analytical methods (e.g., \texttt{manski}, \texttt{tianpearl}), optimization-based techniques (\texttt{autobound}, \texttt{zhangbareinboim}, \texttt{causaloptim}), and more recent innovations such as entropy-constrained bounds (\texttt{entropybounds}) and modified Expectation-Maximization (EM) approaches (\texttt{zaffalonbounds}).
In particular, we extend the \texttt{entropybounds} framework proposed by Jiang et al.~\cite{jiang_approximate_2023} beyond interventional queries to support counterfactual quantities such as the PNS. This extension enables tighter bounds in settings with weak confounding, even for non-identifiable targets on the third rung of Pearl’s causal hierarchy.

\paragraph{Empirical Evaluation.} 
We conduct a systematic evaluation of these algorithms across four canonical causal scenarios, including both discrete and continuous settings. Using a large-scale simulation framework with randomized structural causal models, we measure the tightness of bounds, failure rates, and runtime across thousands of instances. To our knowledge, this constitutes one of the most extensive empirical evaluations of recent causal bounding methods to date.

\paragraph{Tools for Practitioners.} 
We develop an open-source Python package, \texttt{CausalBoundingEngine}\footnote{\url{https://pypi.org/project/causalboundingengine/}}, that implements all bounding algorithms in a single coherent framework. The package provides a unified interface for applying and comparing methods, supports automated pre-processing, and handles both binary and continuous variables where applicable. In addition to this software toolkit, we present two practical tools: (1) a decision tree that helps analysts select an appropriate bounding algorithm, and (2) a Random Forest classifier trained to predict the empirically best-performing algorithm using only observable statistics such as entropy and mutual information of the observed variables.

\medskip

By achieving these objectives, this thesis contributes both a better understanding of partial identification techniques and a practical resource for applying them. The comparative study clarifies the tradeoffs and assumptions underlying different approaches, helping to inform both theory and practice. Meanwhile, the software suite serves as a lasting tool that others can build upon, encouraging experimentation and broader adoption. Ultimately, we hope this work helps shift partial identification from a theoretical curiosity to a mainstream method—one that enables rigorous causal inference even when full identification is out of reach.

\section{Thesis Outline}

The remainder of this thesis is organized as follows.

\textbf{Chapter 2} provides the theoretical background on causal effect identification and formally introduces the concept of partial identification. It reviews the relevant literature and foundational techniques that form the basis for the bounding algorithms presented in the subsequent chapters. In addition, we formally specify the central research question and outline the objectives of the thesis.

\textbf{Chapter 3} presents the collection of bounding algorithms implemented in this study. For each method, we outline the underlying assumptions, provide the mathematical formulation, and describe relevant implementation details.

\textbf{Chapter 4} presents the design and results of our simulation experiments. We introduce a formal framework for conducting the simulations, describe the synthetic data-generating processes, specify the parameters varied, and define the evaluation metrics used to compare algorithms. The empirical results are then reported across multiple settings and causal queries.

\textbf{Chapter 5} presents a discussion and interpretation of the results. We analyze how each algorithm performs under varying conditions, examine trade-offs between bound tightness and validity, and offer an intuitive decision guide for practitioners. In addition, we introduce a data-driven notion of the \emph{best} algorithm for a given scenario and use it to train a predictive model. The chapter concludes with a summary of the main contributions and an outlook on future research directions.

Together, these chapters aim to equip readers with both the theoretical foundations and practical tools needed to incorporate partial identification into causal analyses—helping to bridge the gap between methodological research and applied practice.

%% file: chapters/ch-problem_setting.tex
\chapter{Problem Setting}
\label{ch-problem_setting}

This chapter formalizes the core ideas introduced in Chapter~\ref{ch-introduction} and establishes the theoretical foundations necessary for the algorithmic approaches presented in Chapter~\ref{ch-considered_algorithms}. We begin by situating the task of causal inference within Pearl’s ladder of causation, highlighting the increasing complexity from associational to interventional and counterfactual reasoning. To express these concepts precisely, we introduce structural causal models (SCMs) and the potential outcomes framework, which serve as the formal backbone for defining causal quantities such as the Average Treatment Effect (ATE) and the Probability of Necessity and Sufficiency (PNS). We then discuss the concepts of identifiability and partial identification, motivating the use of bounding techniques when full identification is infeasible. The chapter concludes with a summary of the notation and terminology used throughout the thesis, followed by a clear statement of the central research question and objectives.

\input{chapters/ch-problem_setting/preliminaries}

\section{Research Question and Objectives}
\subsection{Research Question}
Causal inference allows us to move beyond correlation and reason about the effects of interventions. However, in many real-world scenarios, point identification of causal and counterfactual quantities is impossible due to issues such as unobserved confounding. In such settings, one must resort to \emph{partial identification} techniques that provide bounds on the quantity of interest rather than a precise estimate. These bounding methods delineate the range within which the true causal effects or counterfactual probabilities may lie given the available information.

This thesis investigates the following central research question:

\begin{quote}\itshape
How do different algorithms for bounding causal effects and counterfactuals perform under varying data assumptions and experimental conditions?
\end{quote}

In order to address this question, the study focuses on a systematic comparison of multiple bounding methods through simulation. Notably, no real-world data are analyzed in this thesis; instead, a range of synthetic datasets is constructed to represent diverse scenarios. This simulation-based approach allows controlled variation of assumptions and conditions (e.g., the strength of unobserved confounding or the type of causal query) as well as the comparison of the algorithm results to a known ground truth. Ultimately, the aim is to evaluate the strengths and limitations of different bounding approaches and to provide insight into when and how each method can be applied most effectively in practice.

\subsection{Objectives}
To answer this research question, the thesis pursues the following objectives:

\begin{enumerate}
    \item \textbf{Literature Review and Methodology Selection} 
    \begin{itemize}
        \item Survey state-of-the-art techniques for bounding causal effects and counterfactuals.
        \item Identify the assumptions and applicable scenarios for each approach, and select a set of bounding algorithms for further investigation.
    \end{itemize}
    
    \item \textbf{Implementation of Bounding Algorithms} 
    \begin{itemize}
        \item Implement the selected algorithms in a common framework using Python, including both existing methods and original extensions.
        \item Develop a unified interface to apply all methods consistently across datasets.
    \end{itemize}
    
    \item \textbf{Design of Simulated Causal Scenarios} 
    \begin{itemize}
        \item Define a comprehensive set of simulation scenarios with synthetic data, covering:
        \begin{itemize}
            \item different outcome types (binary and continuous),
            \item varied structural causal models with randomized functions,
            \item both interventional queries (e.g., ATE) and counterfactual queries (e.g., PNS).
        \end{itemize}
    \end{itemize}
    
    \item \textbf{Simulation-Based Evaluation} 
    \begin{itemize}
        \item Apply the bounding algorithms to each simulated scenario.
        \item Evaluate the methods in terms of:
        \begin{itemize}
            \item bound tightness (informativeness),
            \item computational efficiency
            \item and validity (coverage of the true effect).
        \end{itemize}
    \end{itemize}
    
    \item \textbf{Comparative Analysis of Methods} 
    \begin{itemize}
        \item Compare algorithm performance across scenarios to identify systematic patterns and trade-offs.
        \item Investigate which methods perform best under which conditions, and how assumptions impact bound quality.
    \end{itemize}

    \item \textbf{Decision Support and Predictive Modeling}
    \begin{itemize}
        \item Develop a decision tree that helps practitioners choose an appropriate bounding method based on scenario characteristics.
        \item If possible, train a machine learning model to predict the best-performing algorithm using observable features.
    \end{itemize}
    
    \item \textbf{Open-Source Software Package} 
    \begin{itemize}
        \item Develop and release the open-source Python package \texttt{CausalBoundingEngine}, integrating all bounding algorithms in a unified framework.
        \item Provide documentation and usage examples to facilitate reproducibility and application to new data.
    \end{itemize}
\end{enumerate}

%% file: chapters/ch-problem_setting/preliminaries.tex
\section{Preliminaries} 
\subsection{Pearl's Causal Ladder}
Judea Pearl's \emph{ladder of causation} organizes causal reasoning into three hierarchical levels  \cite{pearl_causality_2000}. The lowest level is \textbf{association}, which deals with pure observation and correlation (pattern recognition). At this level, one can answer queries of the form “What is $P(Y=y \mid X=x)$?” using the joint distribution of observed variables. Modern machine learning excels at this level by finding statistical associations in data. However, association alone cannot distinguish correlation from causation. 

The second level is \textbf{intervention}, which answers questions about cause and effect by considering external manipulations. Intervention-level queries have the form “What is $P(Y=y \mid \doo(X=x))$?”, asking about the probability of an outcome $Y=y$ when we \emph{force} $X$ to value $x$. Here $\doo(X=x)$ denotes an ideal intervention setting $X$ to $x$ (sometimes thought of as performing a randomized experiment on $X$). Answering such questions requires more than observational data; we either need controlled experiments or a causal model that permits computing the effect of interventions  \cite{pearl_causality_2000}. Intervention reasoning allows us to predict consequences of actions (e.g., determining if a new drug (X) \emph{causes} improvement in health (Y) by administering it). 

The top rung is \textbf{counterfactuals}, which involve reasoning about hypothetical alternate scenarios for individual cases. A counterfactual query asks: “Given what we observed, what would have happened to $Y$ if $X$ had been different?”. For example, “Given that a particular patient took the drug and recovered ($X=1, Y=1$ in actuality), would they have recovered if they had \emph{not} taken the drug ($X=0$)?”. Such queries can be written as $P(Y_{X=x}=y \mid X=x', Y=y')$, meaning the probability that $Y$ would be $y$ if $X$ were $x$, given that in fact $X$ was $x'$ and $Y$ was observed to be $y'$ in reality. 

Counterfactuals are the most challenging type of query because they require modeling two “worlds” at once (the observed world and the hypothetical world) and finding a way to reconcile them. In Pearl's framework, answering counterfactuals demands a fully specified causal model (including assumptions about the underlying mechanisms) so that one can determine, for an individual case, what \emph{would} happen under an alternate action. In summary, each step up the causal ladder (association $\to$ intervention $\to$ counterfactual) requires strictly stronger assumptions or data, but yields richer information. Our problem setting, which involves \emph{bounding causal effects and counterfactuals}, squarely deals with the top two rungs of this ladder – making explicit when full identification is not possible and how we can still learn about causal and counterfactual quantities in those cases. 

\subsection{Structural Causal Models (SCMs)}
\label{prelim-SCM}
To formally reason about interventions and counterfactuals, we use the language of \emph{structural causal models (SCMs)}  \cite{pearl_causality_2000}. An SCM provides a mathematical description of a data-generating process in terms of cause-and-effect relationships. Formally, an SCM is a tuple $(U, V, F)$ where $U$ is a set of exogenous (unobserved) background variables, $V$ is a set of endogenous (observed) variables, and $F$ is a set of functions ${f_V : \mathrm{Pa}(V), U \rightarrow V }$ that determine how each $V_i \in V$ is produced. Specifically, for each endogenous variable $V_i \in V$, $V_i := f_{V_i}(\text{Pa}_i, U_i)$ where $\text{Pa}_i \subseteq V \setminus {V_i}$ are the parents (direct causes) of $V_i$ in the model, and $U_i \subseteq U$ is the subset of exogenous factors influencing $V_i$. The relationships $f_V$ are typically deterministic functions; all uncertainty in the system is encoded by the exogenous variables $U$, which are assumed to have an underlying probability distribution $P(U)$. 

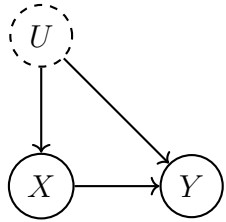
\begin{figure}[h!]\centering
\begin{tikzpicture}[->, node distance=2cm, thick]
\node[draw, circle] (X) {$X$};
\node[draw, circle, above of=X, dashed] (U) {$U$};
\node[draw, circle, right of=X] (Y) {$Y$};
\draw[->] (U) -- (X);
\draw[->] (U) -- (Y);
\draw[->] (X) -- (Y);
\end{tikzpicture}
\caption{A simple causal graph (DAG) representing a structural causal model. Here $X$ is a treatment variable and $Y$ an outcome. An unobserved (indicated by the dashed border) factor $U$ influences both, making $U$ a confounder that creates spurious associations between $X$ and $Y$.}
\label{fig:scm-example}
\end{figure}

The causal structure of an SCM can be represented by a directed acyclic graph (DAG) in which nodes correspond to variables (both observed and unobserved) and directed edges represent direct causal influence via the functions $F$. Figure~\ref{fig:scm-example} illustrates a simple causal DAG. Each observed node (e.g., $X$ or $Y$) has arrows coming from its parent causes. Unobserved variables (like $U$ in the figure) can act as unmeasured common causes (confounders) of multiple observed variables. The absence of an arrow encodes an assumption of no direct causal influence. This graphical representation is not just a qualitative sketch: it encodes conditional independence relations and guides formal causal reasoning (e.g., via $d$-separation and the causal Markov condition). 

SCMs provide the formal semantics for interventions and counterfactuals. Given a \emph{fully specified} SCM—that is, one in which every structural function $F$ and the joint distribution $P(U)$ of the exogenous variables are known—an \textbf{intervention} $\doo(X=x)$ is modeled by modifying the structural equation for $X$ to set $X$ exogenously to $x$ (cutting off its original parent inputs). The model then allows us to compute the resulting distribution of other variables, in particular $Y$.
We denote by $P(Y=y \mid \doo(X=x))$ the interventional distribution of $Y$ under action $X=x$. In an SCM, this can be computed by effectively “fixing” $X=x$ in the equations $F$ and evaluating $Y$.

The fully specified SCM also yields counterfactual quantities through the three steps of
\emph{abduction, action, and prediction} \cite{pearl_causality_2000}.
In the \emph{abduction} step, we take the unit’s observed values—say
\(X = x_{\mathrm{obs}}\) and \(Y = y_{\mathrm{obs}}\)—and condition the prior
\(P(U)\) to obtain the posterior
\(P\!\bigl(U \mid X = x_{\mathrm{obs}},\, Y = y_{\mathrm{obs}}\bigr)\).
In the \emph{action} step, we intervene by replacing the structural
equation for \(X\) with the constant assignment \(X = x\).
In the \emph{prediction} step, we propagate the (posterior) exogenous
variables \(U\) through the modified model to determine the resulting
distribution of \(Y\).

Thus, the counterfactual \(Y_{X = x}\) for a unit with observed history
\((X, Y) = (x_{\mathrm{obs}}, y_{\mathrm{obs}})\) is defined by:
\begin{enumerate}
    \item conditioning on \((x_{\mathrm{obs}}, y_{\mathrm{obs}})\) to infer \(U\),
    \item setting \(X = x\), and
    \item solving for \(Y\) under the altered equations.
\end{enumerate}

The power of SCMs is that they allow such counterfactual reasoning: two “parallel worlds” (one real, one hypothetical) are connected via the shared exogenous variables that encode the unit’s intrinsic properties. Without a fully specified SCM (including those exogenous factors), one generally cannot answer counterfactual questions. 

\subsection{Potential Outcomes and Causal Effects}
An alternative but equivalent framework to describe interventions is the \textbf{potential outcomes} notation (also known as the Neyman-Rubin causal model)  \cite{rubin_estimating_1974}. We denote by $Y_{X=x}$ (often shortened to $Y_x$) the \emph{potential outcome} random variable representing the value that outcome $Y$ would take if $X$ were set to $x$ by an intervention. In a binary treatment setting ($X \in \{0,1\}$), each unit or individual has two potential outcomes: $Y_{0}$ and $Y_{1}$ (shorthand for $Y_{X=0}$ and $Y_{X=1}$). At most one of these can be observed for any given unit (e.g., in reality either $X=0$ or $X=1$ happens, not both), but we can reason about them theoretically or infer their distribution under assumptions. The potential outcome framework is closely connected to the SCM: in fact, under the assumptions of an SCM, $P(Y_x = y)$ is identical to the causal query $P(Y=y \mid \doo(X=x))$. We will freely use the notation $P(Y_{x}=y)$ and $P(Y=y \mid \doo(X=x))$ interchangeably, as they represent the same quantity under the semantics of a given causal model. 

A fundamental causal effect measure is the \textbf{Average Treatment Effect (ATE)}. In the binary case, $X=0$ and $X=1$ might represent absence vs. presence of some treatment or intervention. The ATE is defined as the difference in the average outcome if everyone in the population received the treatment versus if no one did. Using potential outcomes, we can write the ATE as:
\begin{equation}
\label{eq:ate}
\mathrm{ATE} \;=\; P\big(Y_{X=1}=1\big)\;-\;P\big(Y_{X=0}=1\big),
\end{equation}

assuming $Y$ is an indicator (binary) outcome for simplicity. In words, $\mathrm{ATE} = \mathbb{E}[Y_1] - \mathbb{E}[Y_0]$, the difference in the probability (or rate) of the desired outcome $Y=1$ under intervention $X=1$ versus $X=0$.\footnote{Since ATE compares two \emph{interventional} distributions and does not
require counterfactual conjunctions, it is a \emph{rung 2} (intervention-level) quantity on Pearl’s ladder.}

More generally, if $Y$ is numeric, $\mathrm{ATE} = \mathbb{E}[Y \mid \doo(X=1)] - \mathbb{E}[Y \mid \doo(X=0)]$. We emphasize this is a \emph{population-level causal effect}: it averages over all individuals. If $X$ is not binary (say continuous or multi-valued), one can define analogous causal effects, such as contrasts between two intervention regimes ($\mathbb{E}[Y_{x_1} - Y_{x_0}]$ for two specific doses $x_1$ vs $x_0$), or a \emph{marginal structural effect} like $\frac{d}{dx}\mathbb{E}[Y \mid \doo(X=x)]$ at a certain point. In this thesis, we primarily consider binary treatments for clarity, but the concepts can be extended to multi-valued $X$ with appropriate definitions. 

Another important quantity, which ventures into the counterfactual realm (Pearl’s third rung), is the \textbf{Probability of Necessity and Sufficiency (PNS)}  \cite{pearl_causality_2000}. PNS is defined (for binary $X$ and $Y$) as the probability that $X$ was both a necessary and sufficient cause for $Y$ in an individual instance. Formally, it is the probability that $Y$ would be 1 if $X$ is set to 1, \emph{and} $Y$ would be 0 if $X$ is set to 0. Using potential outcomes:
\begin{equation}
\label{eq:pns}
\mathrm{PNS} \;=\; P\big(Y_{X=1}=1 \;\wedge\; Y_{X=0}=0\big),
\end{equation}
This captures the intuitive notion of “how often does $X$ make the difference for $Y$?” In a treatment context, PNS answers: for a random individual, what is the probability that the outcome occurs \emph{only if} the treatment is given (and would not occur otherwise). 

Notice that PNS is inherently a \emph{counterfactual} probability, involving the conjunction of two potential outcomes for the same unit. This is in contrast to ATE, which is a difference of marginal probabilities $P(Y_1=1)$ and $P(Y_0=1)$. ATE can be identified from population data under relatively mild assumptions (for example, in a randomized trial or with appropriate adjustment for confounders, one can estimate $P(Y_1=1)$ and $P(Y_0=1)$). PNS, on the other hand, is much harder to identify: no individual reveals both $Y_{0}$ and $Y_{1}$, so we cannot directly observe the event ${Y_{1}=1 \wedge Y_{0}=0}$ for any single person. Even with experimental data, PNS generally cannot be pinned down exactly without further strong assumptions. We will see that when a target quantity like PNS is \textbf{non-identifiable}, one must resort to alternative strategies such as bounding its possible values. 

\subsection{Response Types in Binary Models}
\label{prelim-responsetypes}
When both treatment $X$ and outcome $Y$ are binary, it is often useful to conceptualize each individual as having one of a set of \textit{response types} characterized by their pair of potential outcomes $(Y_0, Y_1)$ \cite{pearl_causality_2000}. There are four possible response types for any given unit:
\begin{itemize}\setlength{\itemsep}{0pt}
\item \textbf{Never-Taker (Always $0$):} $Y_0 = 0$ and $Y_1 = 0$. The outcome is 0 regardless of whether $X=0$ or $X=1$. (The treatment has no effect and the outcome is always negative.)
\item \textbf{Always-Taker (Always $1$):} $Y_0 = 1$ and $Y_1 = 1$. The outcome is 1 regardless of treatment. ($X$ has no effect; the outcome would happen anyway.)
\item \textbf{Responder (Positive Effect):} $Y_0 = 0$, $Y_1 = 1$. The outcome is 0 if not treated, but 1 if treated. This individual benefits from the treatment (for them, $X$ is a necessary and sufficient cause of $Y$).
\item \textbf{Contrarian (Negative Effect):} $Y_0 = 1$, $Y_1 = 0$. The outcome is 1 if untreated, but 0 if treated. The treatment harms or prevents the outcome for this individual (a rare or pathological case in many scenarios; sometimes ruled out by monotonicity assumptions).
\end{itemize}
These response types provide an intuitive way to express counterfactual queries. For instance, $\mathrm{PNS} = P(Y_0=0, Y_1=1)$ is exactly the proportion of \emph{Responder}-type individuals in the population. Similarly, one can see that the Average Treatment Effect can be related to the balance of positive vs. negative response types: since $P(Y_1=1) = P(\text{Always-1}) + P(\text{Responder})$ and $P(Y_0=1) = P(\text{Always-1}) + P(\text{Contrarian})$, it follows that

\[
\mathrm{ATE} 
= P(\text{Responder}) - P(\text{Contrarian}) \,.
\]
Thus, $\mathrm{ATE}$ is positive if there are more Responders than Contrarians in the population, zero if those groups balance out, and negative if harm outweighs benefit on average. The response-type view underscores that causal effects can vary across individuals: ATE is an average that might mask this heterogeneity. In contrast, counterfactual measures like PNS attempt to quantify an individual-level effect (the probability an individual is of the Responder type). Generally, we cannot observe an individual’s type directly (because we never see both $Y_0$ and $Y_1$ for the same person), but reasoning in terms of response types helps in formulating identification strategies and bounds. 

\subsection{Identifiability and the Need for Bounding}
In causal inference, a target quantity (causal parameter) is said to be \textbf{identifiable} (from given data and assumptions) if it can be expressed as a function of the available information (typically the observational data distribution, possibly augmented with experimental data, and the known causal graph structure) \cite{pearl_causality_2000}. Identifiability means that, despite not observing all counterfactuals, the quantity has the same value in all causal models that are consistent with our known inputs.  For example, under the assumption of no unmeasured confounding, the ATE is identifiable from observational data – one can compute it via $P(Y=1 \mid \doo(X=1)) - P(Y=1 \mid \doo(X=0))$, and if we can observe the confounders, this reduces to estimable probabilities like $P(Y=1 \mid X=1, \text{adjusted for confounders})$ etc.

In contrast, \textbf{non-identifiable} queries are those for which multiple different values are compatible with what we know. PNS (see Equation~\ref{eq:pns}) is a canonical example – given only observational or even experimental distributions of $X$ and $Y$, PNS cannot be derived exactly because we do not know how to pair up outcomes across the two potential worlds for each individual. There can exist two extreme models, both fitting the data, one in which all individuals who responded under $X=1$ would not have responded under $X=0$ (maximizing PNS), and another in which the individuals who respond under $X=1$ are exactly those who also would have responded under $X=0$ (minimizing PNS). If nothing else is assumed, the true PNS could be anywhere between those extremes, making it unidentifiable. 

When a causal query is not identifiable, we shift our goal to \textbf{partial identification}: rather than pinpointing a single value, we derive a range of possible values (interval or bounds) that the quantity can take, given the data and assumed model structure. \textbf{Bounding} refers to finding logical or mathematical limits on a causal quantity using the information at hand. The bounds reflect our uncertainty due to non-identifiability: the true value is guaranteed to lie within the bounds as long as our model assumptions and data constraints are valid. A classic example comes from Manski's work on evaluating treatment effects with minimal assumptions  \cite{manski_nonparametric_1990}. Manski derived so-called “worst-case” bounds for the ATE of a binary treatment on a binary outcome without assuming any confounders are observed. In the simplest case, using only the raw outcome rates among treated and untreated groups, one can show the ATE is bounded between
\begin{align*}\quad
& P(Y=1,  X=1) - P(Y=1, X=0) - P(X=1) \\
\text{and} \quad
& P(Y=1,  X=1) - P(Y=1, X=0)  + P(X=0),
\end{align*}
which represent the most extreme possibilities consistent with the observed proportions. These are often called Manski bounds; they make virtually no assumptions about the data-generating process beyond what is observed. In practice, additional assumptions (like confounding strength or structural knowledge) can tighten the bounds. Bounding techniques have been further developed in many settings, as we will see in Chapter~\ref{ch-considered_algorithms}. The general approach is typically to enumerate all logically possible consistent assignments to the unobserved potential outcomes or latent variables, and then find the extreme values of the target query. This often reduces to an optimization problem subject to constraints given by known probabilities. 

\subsection{Sharpness of Bounds}
\label{prelim-sharpness}

When presenting bounds for a causal quantity, an important concept is \textbf{sharpness}. Bounds are said to be \emph{sharp} if they are the tightest possible given the assumptions and available data. A bound that is not sharp may be unnecessarily wide, meaning it fails to reflect the full informativeness of the assumptions.

Sharpness is particularly relevant when comparing different bounding algorithms. A method that returns non-sharp bounds is suboptimal in principle, as it either overstates uncertainty or does not fully leverage the available constraints. In theory, sharpness is often established by reformulating the causal query and assumptions as a constrained optimization problem, where the computed bounds correspond to the minimum and maximum of the estimand over the feasible region \cite{duarte_automated_2023, sachs_general_2023}.

However, in simulation settings like those used in this thesis, sharpness must be interpreted with care. In some cases, bounds that are \emph{provably sharp under their assumptions} may still perform poorly in practice—e.g., by being invalid (not containing the ground truth), overly narrow, or less informative (wider) than bounds from other methods.

This discrepancy can partly be explained by characteristics of the data-generating processes, which will be introduced in Chapter~\ref{ch-simulation_and_experiment}. Importantly, while sharpness remains a valuable theoretical benchmark, it may not always be a reliable indicator of empirical performance in simulations. Our evaluation framework accounts for this by assessing not only whether bounds are theoretically sharp, but also whether they remain empirically valid and informative under varying conditions and potential assumption violations (e.g., random noise).

\subsection{Notation and Terminology}
Before proceeding, we summarize the notation and conventions that will be used throughout this thesis:
\begin{itemize}\setlength{\itemsep}{0pt}
\item We use capital letters ($X, Y, Z$, etc.) to denote random variables, and lowercase ($x, y, z$) for their values. We may sometimes write $P(x)$, in which case it will be clear from context, that we mean $P(X=x)$.
\item \textbf{Observations vs Interventions:} $P(Y=y \mid X=x)$ represents a conditional probability from observational data (seeing $X=x$), whereas $P(Y=y \mid \doo(X=x))$ denotes an interventional probability (setting $X=x$ deliberately). We read $\doo(X=x)$ as “do $X=x$”.
\item \textbf{Potential Outcomes:} $Y_{X=x}$ (shorthand $Y_x$) denotes the potential outcome of $Y$ under intervention $X=x$. For example, $Y_1$ and $Y_0$ are the two potential outcomes for $Y$ corresponding to $X=1$ or $X=0$. We sometimes write statements like $Y_x = y$ to describe the event that $Y$ would equal $y$ if $X$ were set to $x$. Joint statements involving multiple potential outcomes (for the same unit) will be used, e.g. $Y_{1}=1 \wedge Y_{0}=0$ corresponds to the \textit{Responder} type as described above.
\item \textbf{Distributions:} We write $\text{Uni}(a, b)$ for the uniform distribution over $[a, b]$, $\text{Bern}(p)$ for the Bernoulli distribution with success probability $p$, and $\mathcal{N}(\mu, \sigma^2)$ for the normal distribution with mean $\mu$ and variance $\sigma^2$.

\item \textbf{SCM Components:} In discussing structural causal models, we represent variables as \emph{nodes} in a directed acyclic graph (DAG), with arrows indicating direct causal relationships. The notation $\text{Pa}(X)$ (or $\mathrm{Parents}(X)$) denotes the set of parent nodes of variable $X$ and $f_X$ denotes the function  $f_X:\text{Pa}(X) \to X$ for the same variable, if not specified otherwise. Exogenous (unobserved) variables are denoted by $U$ and typically have no parents.

To indicate which variables are unobserved in a DAG, we use a visual convention: nodes with a \emph{dashed border} represent unobserved or latent variables.

\item \textbf{Causal Effect Symbols:} The Average Treatment Effect is abbreviated as ATE. Probability of Necessity and Sufficiency is PNS.

\item \textbf{Identifiability and Bounds:} We will say a causal query is “identifiable” if it can be expressed in terms of observable distributions; otherwise, it is “non-identifiable”. For non-identifiable queries, we will present bounds (interval estimates). If needed, we will use notation like $[\text{LB}, \text{UB}]$ to denote a bound from below by LB and above by UB.

\item \textbf{Entropy and Mutual Information:}  
We write \( H(X) \) to denote the (Shannon) entropy of a discrete random variable \( X \), computed in bits as
\[
H(X) = -\sum_{x} P(x) \log_2 P(x),
\]
where the sum ranges over all values \( x \) in the support of \( X \).

The joint entropy of two variables \( X \) and \( Y \) is defined as
\[
H(X, Y) = -\sum_{x, y} P(x, y) \log_2 P(x, y).
\]

The \emph{mutual information} between \( X \) and \( Y \), denoted \( I(X; Y) \), quantifies the information shared between them:
\[
I(X; Y) = \sum_{x, y} P(x, y) \log_2 \frac{P(x, y)}{P(x) P(y)}.
\]
Equivalently, \( I(X; Y) = H(X) + H(Y) - H(X, Y) \).
\end{itemize} 
\medskip
\noindent
All the above concepts and notation will be used consistently in the subsequent chapters. Having established the preliminaries, we now move on to the specific problem setting of bounding causal effects and counterfactuals in more detail.

%% file: chapters/ch-methods.tex
\chapter{Considered Algorithms}
\label{ch-considered_algorithms}

This chapter presents the set of bounding algorithms considered in this thesis. Each method is introduced with a brief description of its underlying logic, assumptions, and the causal quantities it targets.

We begin with the more intuitive and well-established approaches and gradually proceed to more recent and sophisticated algorithms.

Where applicable, we also comment on implementation details relevant to the empirical evaluation. Parameter choices are discussed later in Section~\ref{sec:parameter-selection}. The selected algorithms span a diverse range of techniques—from symbolic bounding methods to optimization-based approaches—and are evaluated systematically in Chapter~\ref{ch-simulation_and_experiment}.

\input{chapters/ch-methods/manski}

\input{chapters/ch-methods/tianpearl}

\input{chapters/ch-methods/2SLS_OLS}

\input{chapters/ch-methods/causaloptim}

\input{chapters/ch-methods/autobound}

\input{chapters/ch-methods/entropybounds}

\input{chapters/ch-methods/zaffalon}

\input{chapters/ch-methods/zhangbareinboim}

\section{Overview of Considered Algorithms}
\label{sec:algorithm-overview}

Table~\ref{tab:algorithm-overview} summarizes all bounding algorithms evaluated in this thesis. For each method, we report the computational approach, the year the underlying theory was published, and the type of causal queries it can address (ATE, PNS, or both).

\begin{table}[h]
\centering
\caption{Summary of all considered bounding algorithms, including their computational method, year of theoretical publication, and supported causal queries (ATE and/or PNS).}
\label{tab:algorithm-overview}
\scriptsize
\begin{tabularx}{\textwidth}{lXcc}
\toprule
Algorithm & Method & Year & Queries \\
\midrule
\texttt{OLS}            & Linear regression (OLS) & 1805 & ATE \\
\addlinespace
\texttt{2SLS}           & Two-stage least squares (IV regression) & 1957 & ATE \\
\addlinespace
\texttt{manski}         & Symbolic bounds & 1990 & ATE \\
\addlinespace
\texttt{tianpearl}      & Symbolic bounds & 2000 & PNS \\
\addlinespace
\texttt{zhangbareinboim} & LP (continuous outcome) & 2021 & ATE \\
\addlinespace
\texttt{causaloptim}    & LP (symbolic bounds) & 2023 & ATE, PNS \\
\addlinespace
\texttt{zaffalonbounds} & Causal EM (EMCC) & 2021 & ATE, PNS \\
\addlinespace
\texttt{autobound}      & Polynomial optimization & 2023 & ATE, PNS \\
\addlinespace
\texttt{entropybounds}  & LP with entropy constraint & 2023 & ATE, PNS \\
\bottomrule
\end{tabularx}
\end{table}

%% file: chapters/ch-methods/manski.tex
\section{\texttt{manski}}
\label{alg-manski}

In his seminal 1990 paper, Charles F. Manski derived a simple and general bound on the Average Treatment Effect (ATE, as defined in Equation~\ref{eq:ate}) under minimal assumptions~\cite{manski_nonparametric_1990}. His approach applies to binary treatments \( X \in \{0,1\} \) and binary outcomes \( Y \in \{0,1\} \), and crucially makes no assumptions about the treatment assignment mechanism or the presence of unobserved confounding.

Manski’s method is based on rewriting the potential outcome distribution \( P(Y_1) \) using the law of total probability:
\[
P(Y_1) = P(Y_1 \mid X=1) \cdot P(X=1) + P(Y_1 \mid X=0) \cdot P(X=0)
\]
A similar expression holds for \( P(Y_0) \), and the ATE is defined as \( \text{ATE} = P(Y_1) - P(Y_0) \).

In this decomposition, the only unidentifiable terms are \( P(Y_1 \mid X=0) \) and \( P(Y_0 \mid X=1) \)—counterfactuals not observed in the data. The classic Manski bounds are obtained by assigning these unknown quantities their logical extremes (either 0 or 1), while keeping the observed conditional probabilities and treatment proportions fixed.

This yields the following sharp bounds on the ATE:
\begin{align*}
\mathrm{ATE} \in \Big[\, 
& P(Y=1 \mid X=1) \cdot P(X=1) - P(Y=1 \mid X=0) \cdot P(X=0) - P(X=1), \\
& P(Y=1 \mid X=1) \cdot P(X=1) + P(X=0) - P(Y=1 \mid X=0) \cdot P(X=0) 
\,\Big]
\end{align*}

These bounds represent the widest possible range of ATE values consistent with the observed data under no assumptions beyond consistency and the binary structure of \( X \) and \( Y \). In our simulations, we refer to them as \texttt{ATE\_manski}.

%% file: chapters/ch-methods/tianpearl.tex
\section{\texttt{tianpearl}}

In the year 2000, Jin Tian and Judea Pearl proposed symbolic bounds for the Probability of Necessity and Sufficiency (PNS, Equation~\ref{eq:pns}) \cite{tian_probabilities_2000}. These bounds are derived under minimal assumptions and apply to binary variables in the presence of unobserved confounding.

The \text{PNS} quantifies the probability that treatment \( X \) is both necessary and sufficient for the outcome \( Y \). In other words, it is the probability that a randomly selected individual is of the \emph{Responder} type (see Section~\ref{prelim-responsetypes}).

Without making any assumptions about the data-generating process Tian and Pearl showed that the \text{PNS} can still be bounded using only the observational joint distribution \( P(X, Y) \):
\[
0 \leq \text{PNS} \leq P(X=1, Y=1) + P(X=0, Y=0)
\]

These bounds follow from logical consistency and the structure of potential outcomes in a binary setting. While the lower bound is vacuous, the upper bound can sometimes be informative—especially when the observed data exhibits strong alignment between treatment and outcome. The intuition is that individuals with observed pairs \((X=1, Y=1)\) or \((X=0, Y=0)\) could plausibly be Responders, and the upper bound assumes the largest such proportion compatible with the data.

In our implementation, this method is referred to as \texttt{PNS\_tianpearl}.

%% file: chapters/ch-methods/2SLS_OLS.tex
\section{Confidence Intervals as Bounds}
\label{sec:methods-2SLS_OLS}

The following algorithms are heuristics that allow us to benchmark more principled partial identification strategies against a simple and widely-used tool from statistical inference: confidence intervals. Specifically, the methods \texttt{ATE\_OLS} and \texttt{ATE\_2SLS} estimate the Average Treatment Effect (ATE), as defined in Equation~\ref{eq:ate}, using linear regression techniques and interpret the resulting confidence intervals on the treatment coefficient as bounds on the same quantity.

To compute these intervals, we specify a desired confidence level, commonly denoted as \( 1 - \alpha \), where \( \alpha \in (0,1) \) is the significance level. 

\subsection{\texttt{OLS}}

This method is applied in a simple confounding scenario, where both the treatment \( X \) and outcome \( Y \) are influenced by an unobserved confounder \( U \):

\begin{center}
\begin{tikzpicture}[->, node distance=2cm, thick]
  \node[draw, circle] (X) {\( X \)};
  \node[draw, circle, right of=X] (Y) {\( Y \)};
  \node[draw, circle, dashed, above of=X] (U) {\( U \)};
  
  \draw (X) -- (Y);
  \draw (U) -- (X);
  \draw (U) -- (Y);
\end{tikzpicture}
\end{center}

We estimate a linear model using Ordinary Least Squares (OLS):
\[
Y_i = \alpha + X_i \beta + \varepsilon_i
\]
OLS provides both a point estimate for the treatment effect \( \hat{\beta} \) and a confidence interval at the specified level. We interpret this confidence interval as a bound on the \text{ATE}, although this interpretation is heuristic and not causally justified under unobserved confounding. Nonetheless, this approach offers a useful reference for how well (or poorly) naive regression captures treatment effects.

\subsection{\texttt{2SLS}}

This method applies to scenarios where an instrumental variable \( Z \) affects the treatment \( X \), which in turn affects the outcome \( Y \), while an unobserved confounder \( U \) influences both \( X \) and \( Y \):

\begin{center}
\begin{tikzpicture}[->, node distance=2cm, thick]
  \node[draw, circle, left of=X] (Z) {\( Z \)};
  \node[draw, circle] (X) {\( X \)};
  \node[draw, circle, right of=X] (Y) {\( Y \)};
  \node[draw, circle, dashed, above of=X] (U) {\( U \)};

  \draw (Z) -- (X);
  \draw (X) -- (Y);
  \draw (U) -- (X);
  \draw (U) -- (Y);
\end{tikzpicture}
\end{center}

To account for endogeneity due to unobserved confounding, we estimate the \text{ATE} using Two-Stage Least Squares (2SLS):

\begin{enumerate}
  \item \textbf{First stage:} Regress the treatment \( X \) on the instrument \( Z \):
  \[
  X_i = \gamma_0 + \gamma_1 Z_i + \eta_i
  \]
  This yields predicted values \( \hat{X}_i \) which isolate variation in \( X \) attributable to \( Z \) only.
  
  \item \textbf{Second stage:} Regress the outcome \( Y \) on the fitted values \( \hat{X} \):
  \[
  Y_i = \alpha + \beta \hat{X}_i + \varepsilon_i
  \]
\end{enumerate}

As with OLS, the 2SLS estimator yields a coefficient \( \hat{\beta}_{\text{2SLS}} \) and an associated confidence interval, which we again treat heuristically as a bound on the \text{ATE}. Unlike OLS, this method may be causally justified under the classical IV assumptions, provided the instrument \( Z \) is valid (relevant and exogenous).

%% file: chapters/ch-methods/causaloptim.tex
\section{\texttt{causaloptim}}

The \texttt{causaloptim} package\footnote{\url{https://github.com/sachsmc/causaloptim}} provides a general-purpose framework for deriving tight symbolic bounds on non-identifiable causal queries using linear programming (LP). Developed by Sachs et al. in 2023 \cite{sachs_general_2023, jonzon_accessible_2024}, it builds on the foundational work of Balke and Pearl~\cite{balke_counterfactual_1994}. The method takes as input a causal diagram (DAG) over categorical variables and a target causal query—such as a counterfactual probability or an interventional expression.

The output consists of symbolic expressions for the bounds—functions of observable probabilities—that can be evaluated on any compatible dataset. (This is why it does not take actual observational data as input.)

If the DAG and query fulfill certain conditions, these symbolic bounds are guaranteed to be sharp (See Section~\ref{prelim-sharpness}).\footnote{These conditions are fulfilled by all our binary scenarios (and all our queries), so in theory, the bounds produced by \texttt{causaloptim} should be sharp.} 

This method internally uses two algorithms:

The first algorithm takes the DAG as input and constructs a system of linear equations that symbolically relate the observable probabilities to the distribution over possible response types (see Section~\ref{prelim-responsetypes}). These equations define the feasible region of the resulting linear program.\footnote{The original \texttt{R} package also allows users to incorporate additional assumptions (e.g., monotonicity), but we omit this functionality, as none of the scenarios in our study involve such assumptions.}

The second algorithm takes the target query—which may involve interventions, counterfactuals, or observed quantities—and represents it as a linear objective function over the response-type variables.

The resulting LP is solved using a vertex enumeration algorithm. Specifically, \texttt{causaloptim} relies on the double description method~\cite{Motz53} to enumerate all vertices of the dual polytope, and then applies the strong duality theorem to recover the optimal solution to the primal problem. This yields two expressions—one for the lower bound and one for the upper bound—each formulated in terms of observable quantities.

This overall approach generalizes the linear programming strategy, enabling automated symbolic bound derivation with a guarantee of sharpness across a wide class of causal DAGs and queries.

In our implementation, we use \texttt{causaloptim} to compute bounds for both the Average Treatment Effect (ATE, Equation~\ref{eq:ate}) and the Probability of Necessity and Sufficiency (PNS, Equation~\ref{eq:pns}), referred to respectively as \texttt{ATE\_causaloptim} and \texttt{PNS\_causaloptim}.

%% file: chapters/ch-methods/autobound.tex
\section{\texttt{autobound}}
\label{alg-autobound}

The \texttt{autobound} package\footnote{\url{https://www.tandfonline.com/doi/suppl/10.1080/01621459.2023.2216909}} implements the causal bounding framework developed by Duarte et al. in 2023 \cite{duarte_automated_2023}, providing a general-purpose tool for deriving bounds on non-identifiable causal queries in discrete settings. The method accepts a causal DAG, a target estimand, a set of assumptions, and observed data, and then returns the tightest possible range of compatible values—i.e., sharp bounds—via constrained polynomial optimization.

Causal queries are translated into polynomial expressions over \emph{response type variables}, and bounded by solving a polynomial program subject to constraints implied by the causal model and the observed data.

To illustrate how this is done, it is easiest to work through a little example:

\subsection{Example}
\label{alg-autobound-example}

Consider a simple confounded binary setting where all variables \( U, X, Y \in \{0,1\} \), with causal structure \( U \rightarrow X \rightarrow Y \) and \( U \rightarrow Y \). While we do not observe the confounder $U$, we do observe the following joint distribution $P(X,Y)$:
\[
\begin{array}{c|cc}
  & Y=0 & Y=1 \\
\hline
X=0 & 0.3 & 0.2 \\
X=1 & 0.1 & 0.4 \\
\end{array}
\]

We now introduce a representation where each unit in the population is characterized by two components:
\begin{enumerate}
    \item its observed treatment assignment \( X \in \{0,1\} \), and  
    \item its unobserved response type, defined by the pair of potential outcomes \( (Y_0, Y_1) \in \{0,1\}^2 \).
\end{enumerate}

This yields a total of \( 2 \cdot 4 = 8 \) possibilities for each unit, which we index as \( q_0, \dots, q_7 \). Each \( q_i \) represents the probability of a unit having a specific combination of treatment assignment and response type. For example:
\begin{itemize}
    \item \( q_0 \): units with \( (Y_1 = 0, Y_0 = 0) \) and \( X=0 \),
    \item \( q_1 \): units with \( (Y_1 = 0, Y_0 = 0) \) and \( X=1 \),
    \item \( q_2 \): units with \( (Y_1 = 0, Y_0 = 1) \) and \( X=0 \),
    \item \dots
    \item \( q_7 \): units with \( X=1 \) and \( (Y_0 = 1, Y_1 = 1) \).
\end{itemize}

Although we do not observe these types directly, we observe \( P(X,Y) \) and can relate it to the \( q_i \) using the following condition:
\[
Y = 
\begin{cases}
Y_0 & \text{if } X=0 \\
Y_1 & \text{if } X=1
\end{cases}
\]
This allows us to express the observed data as linear constraints on the \( q_i \). For example, the probability of observing \( X=0 \) and \( Y=0 \) corresponds to the total mass of all units with \( X=0 \) whose \( Y_0 = 0 \), i.e., \( q_0 + q_4 \). Applying this logic across all cells of the observed distribution leads to the following constraints:
\[
\begin{aligned}
q_0 + q_4 &= 0.3 &\quad& \text{(X=0, Y=0)} \\
q_2 + q_6 &= 0.2 &\quad& \text{(X=0, Y=1)} \\
q_1 + q_3 &= 0.1 &\quad& \text{(X=1, Y=0)} \\
q_5 + q_7 &= 0.4 &\quad& \text{(X=1, Y=1)} \\
\sum_{i=0}^{7} q_i &= 1 &\quad& \text{(probabilities sum to 1)} \\
q_i &\geq 0 &\quad& \text{for all } i
\end{aligned}
\]

The ATE becomes a linear function of the response-type marginals:
\[
\text{ATE} = P(Y_1) - P(Y_0) = (q_4 + q_5) - (q_2 + q_3)
\]

We now solve the following linear program:
\[
\begin{aligned}
\text{min/max } & (q_4 + q_5) - (q_2 + q_3) \\
\text{subject to } & \text{the above constraints}.
\end{aligned}
\]

Solving this program yields:
\[
\text{ATE} \in [-0.3,\ 0.7]
\]

We can also compute bounds for the Probability of Necessity and Sufficiency, we proceed analogously. The PNS is defined as:
\[
\text{PNS} = P(Y_1 = 1, Y_0 = 0)
\]
which corresponds precisely to the Responder type. In terms of our $q_i$, this becomes:
\[
\text{PNS} = q_4 + q_5
\]
So we simply change the objective of the linear program to:
\[
\text{min/max } q_4 + q_5
\]
All constraints remain unchanged.

This example captures the core logic behind the \texttt{autobound} framework: it reduces discrete causal inference problems to constrained optimization over latent response-type distributions. While we manually derived the objective and constraints for this toy scenario, the contribution of Duarte et al.~\cite{duarte_automated_2023} lies in generalizing this approach to arbitrary discrete causal models and queries. 

Their algorithm accepts any DAG, target estimand, and set of assumptions, and systematically compiles them into a polynomial program\footnote{In our example we arrived at a \emph{linear} program, but \texttt{autobound} also allows for scenarios where this is not the case.} whose solution yields sharp bounds—bounds that may collapse to a point estimate if the query is identifiable. In our study, we use this method to compute bounds for both the ATE and the PNS, which we refer to as \texttt{ATE\_autobound} and \texttt{PNS\_autobound}, respectively.

%% file: chapters/ch-methods/entropybounds.tex
\section{\texttt{entropybounds}}
\label{sec:methods-entropybounds}

Jiang et al., in their work from 2023, propose a linear programming approach for bounding causal effects under \emph{weak confounding}, where the unobserved confounder is assumed to have low entropy~\cite{jiang_approximate_2023}. The key idea is that low entropy implies limited dependence between treatment and outcome via the confounder, thereby shrinking the feasible set of compatible models and enabling tighter bounds than classical results.

Their method targets interventional probabilities of the form \( P(Y \mid \doo(X)) \), and thus allows one to bound the Average Treatment Effect by solving the program for \( P(Y_{X=1}) \) and \( P(Y_{X=0}) \). We refer to their implementation as \texttt{ATE\_entropybounds}\footnote{\url{https://github.com/ziwei-jiang/Approximate-Causal-Effect-Identification-under-Weak-Confounding}}. The relevant causal structure corresponds to the classical confounding setup, depicted in the DAG below:

\begin{center}
\begin{tikzpicture}[->, node distance=2cm, thick]
  \node[draw, circle] (X) {\( X \)};
  \node[draw, circle, right of=X] (Y) {\( Y \)};
  \node[draw, circle, dashed, above of=X] (U) {\( U \)};
  
  \draw (X) -- (Y);
  \draw (U) -- (X);
  \draw (U) -- (Y);
\end{tikzpicture}
\end{center}

Here, \(X\) and \(Y\) are categorical variables (i.e., finite and discrete).

We extend their approach to handle a rung 3 counterfactual quantity: the Probability of Necessity and Sufficiency, implemented in our code as \texttt{PNS\_entropybounds}.

Both \texttt{ATE\_entropybounds} and \texttt{PNS\_entropybounds} require a parameter \( \theta \in \mathbb{R}_+ \)\footnote{In scenarios where the confounder is binary, it holds that \(\theta \in [0,1]\).}, which is assumed to be an upper bound on the confounder entropy \(H(U)\). All information-theoretic quantities in this section are expressed in terms of \emph{bits}, and we use \(\log\) to denote the base-2 logarithm.

\subsection{\texttt{ATE\_entropybounds}}
\label{sec:alg-entropy-ATE}

Let the treatment $X$ take values in the finite set $\{x_1,...,x_n\}$ and let the outcome $Y$ take values in the finite set $\{y_1,...,y_m\}$.

The \texttt{ATE\_entropybounds} algorithm casts the task of finding bounds \(\text{LB}\) and \(\text{UB}\) such that 
\[
\text{LB} \leq P( Y = y_p \mid \doo(X = x_q)) \leq \text{UB}
\]
for some $q \in \{1,...,n\} $ and $p \in \{1,...,m\}$, as the following linear program:

\begin{align*}
\text{LB/UB} = \min/\max \quad & \sum_j b_{pj} P(x_j) \\
\text{s.t.} \quad & \sum_{i,j} b_{ij} P(x_j) = 1 \\
& b_{iq} P(x_q) = P(y_i, x_q) \quad \forall i \\
& 0 \leq b_{ij} \leq 1 \quad \forall i,j \\
& \sum_{i,j} b_{ij} P(x_j) \log\left( \frac{b_{ij}}{\sum_k b_{ik} P(x_k)} \right) \leq \theta
\end{align*}

Here, \( b_{ij} := P(Y_{x_q} = y_i \mid X = x_j) \) are the decision variables. The variables $j$ and $k$ are indexing over the set $\{1,...,n\}$, while $i$ is indexing over $\{1,...,m\}$.

The first three constraints ensure normalization and consistency with the observed data and closely resemble the setup of the \texttt{autobound} example, described in Section~\ref{alg-autobound-example}.

The final constraint—referred to as the \emph{entropy constraint}—is the key contribution of this method. We now explain how it is derived.

\paragraph{From SWIGs to Markov Chains.}

\begin{figure}[H]
\centering
\begin{tikzpicture}[->, node distance=2cm, thick]
  \node[draw, circle] (X) {\( X \)};
  \node[draw, circle, right of=X] (Y) {\( Y \)};
  \node[draw, circle, dashed, above of=X] (U) {\( U \)};
  
  \draw (X) -- (Y);
  \draw (U) -- (X);
  \draw (U) -- (Y);
\end{tikzpicture}
\caption{A DAG with treatment, outcome, and unobserved confounder}
\label{fig:ATE_entr_confDag}
\end{figure}
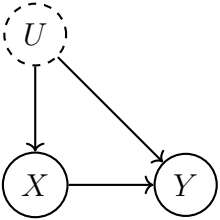

Let us first consider the standard DAG for this setting, shown in Figure~\ref{fig:ATE_entr_confDag}.

We note that both \(X\) and \(Y\) are deterministic functions of their parent variables:
\[
\begin{aligned}
X &= f_X(U) \\
Y &= f_Y(X, U)
\end{aligned}
\]

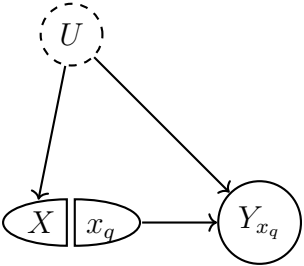
\begin{figure}[H]
\centering
\begin{tikzpicture}[->, node distance=2.5cm, thick]
  \node[name=Xq,shape=swig vsplit, swig vsplit={gap=3pt}]{
    \nodepart{left}{$X$}
    \nodepart{right}{$x_q$} };
  \node[draw, circle, right of=Xq] (Yx) {\( Y_{x_q} \)};
  \node[draw, circle, dashed, above of=Xq] (U) {\( U \)};
  
  \draw (Xq) -- (Yx);
  \draw (U) -- (Yx);
  \draw (U) -- (Xq.142);
\end{tikzpicture}
\caption{Single-World Intervention Graph (SWIG) under the intervention \( X = x_q \)}
\label{fig:SWIG_X_eq_xq}
\end{figure}

Next, we consider an alternate world in which the intervention \(\doo(X = x_q)\) has been performed. This is represented by a Single World Intervention Graph (SWIG), shown in Figure~\ref{fig:SWIG_X_eq_xq}.

In this setting, \( Y_{x_q} \) is now only a function of \(U\):
\[
Y_{x_q} = f_Y(x_q, U)
\]

Thus, both \(X\) (in the original DAG) and \(Y_{x_q}\) (in the SWIG) are deterministic functions of \(U\). Once \(U\) is fixed, both \(X\) and \(Y_{x_q}\) take on fixed values. It follows that
\[
Y_{x_q} \perp X \mid U
\]
which implies the Markov chain:
\[
X \to U \to Y_{x_q}
\]

\paragraph{Data Processing Inequality.}

The \emph{Data Processing Inequality} (DPI) from information theory states that for any Markov chain \(A \to B \to C\), the mutual information satisfies \( I(A; C) \leq I(A; B) \).

Applying the DPI to the chain \(X \to U \to Y_{x_q}\) yields:
\[
I(X; Y_{x_q}) \leq I(X; U)
\]

Using the identity \( I(X; U) = H(U) - H(U \mid X) \), and the fact that entropy is always non-negative, we obtain:
\[
I(X; U) \leq H(U)
\]

Combining this with the previous result gives:
\begin{equation}
\label{eq:entropy_bound}
I(X; Y_{x_q}) \leq H(U)
\end{equation}

\paragraph{Rewriting the Mutual Information.}

To complete the derivation of the entropy constraint, we express \( I(X; Y_{x_q}) \) in terms of the decision variables \( b_{ij} \). Applying the definition of mutual information, along with the chain rule and the law of total probability (in the denominator of the logarithm), we obtain:
\[
I(X; Y_{x_q}) = \sum_{i,j} P(Y_{x_q} = y_i \mid X = x_j) P(x_j) \log\left( \frac{P(Y_{x_q} = y_i \mid X = x_j) P(x_j)}{\sum_k P(Y_{x_q} = y_i \mid X = x_k) P(x_k) P(x_j)} \right)
\]

The terms \(P(x_j)\) cancel inside the logarithm, and substituting \( b_{ij} = P(Y_{x_q} = y_i \mid X = x_j) \) simplifies the expression:
\[
I(X; Y_{x_q}) = \sum_{i,j} b_{ij} P(x_j) \log\left( \frac{b_{ij}}{\sum_k b_{ik} P(x_k)} \right)
\]

Together with inequality~\ref{eq:entropy_bound} and the assumption that \(\theta\) serves as an upper bound on \( H(U) \), this results in the \emph{entropy constraint} used in the linear program of \texttt{ATE\_entropybounds}.

\subsection{\texttt{PNS\_entropybounds}}

The \texttt{PNS\_entropybounds} algorithm extends the \texttt{ATE\_entropybounds} framework to a \emph{counterfactual} quantity of rung 3, namely the Probability of Necessity and Sufficiency (PNS), defined in Equation~\ref{eq:pns} for binary treatment and outcome as
\[
\text{PNS} = P(Y_{X=1} = 1, Y_{X=0} = 0).
\]
Using a similar approach to that proposed by Jiang et al.\ for bounding the interventional quantity \( P(Y_{X=1}) \), the bounds for PNS can be tightened under the \emph{weak confounding} assumption, which limits the mutual information between \( X \) and the potential outcomes.

To this end, we define a vector of optimization variables \( q \in \mathbb{R}^8 \), where each entry \( q_i \) corresponds to the joint probability \( P(Y_1 = y_1, Y_0 = y_0, X = x) \) for a specific triple \( (y_1, y_0, x) \in \{0,1\}^3 \) (This is the same enumeration as in Section \ref{alg-autobound-example}). The mapping between indices \( i \) and value combinations is given in Table~\ref{tab:q_enumeration}.

\begin{table}[H]
\centering
\caption{Enumeration of \( q_i \) entries over all \((y_1, y_0, x) \in \{0,1\}^3\).}
\label{tab:q_enumeration}
\begin{tabular}{ccl}
\toprule
\textbf{Index \( i \)} & \textbf{\((y_1, y_0, x)\)} & \textbf{Interpretation} \\
\midrule
0 & (0, 0, 0) & No outcome under either treatment; untreated \\
1 & (0, 0, 1) & No outcome under either treatment; treated \\
2 & (0, 1, 0) & Outcome only if untreated; untreated \\
3 & (0, 1, 1) & Outcome only if untreated; treated \\
4 & (1, 0, 0) & Outcome only if treated; untreated \\
5 & (1, 0, 1) & Outcome only if treated; treated \\
6 & (1, 1, 0) & Outcome always; untreated \\
7 & (1, 1, 1) & Outcome always; treated \\
\bottomrule
\end{tabular}
\end{table}

The PNS then corresponds to the sum \( q_4+ q_5 \). We optimize over the vector \( q \) as follows:

\begin{align*}
\text{LB/UB} = \min/\max \quad & q_4 + q_5 \\
\text{s.t.} \quad & \sum_{i=0}^{7} q_i = 1 \\
& q_0 + q_4 = P(X=0, Y=0) \\
& q_2 + q_6 = P(X=0, Y=1) \\
& q_1 + q_3 = P(X=1, Y=0) \\
& q_5 + q_7 = P(Y=1, X=1) \\
& q_i \geq 0 \quad \forall i \\
& \sum_{i=0}^{7} q_i \log\left( \frac{q_i}{r_i} \right) \leq \theta 
\end{align*}

Here, \( r_i \) is defined as the product of the marginal distributions over potential outcomes and treatment:
\[
r_i := P(Y_1 = y_1^{(i)}, Y_0 = y_0^{(i)}) \cdot P(X = x^{(i)}),
\]
where \( (y_1^{(i)}, y_0^{(i)}, x^{(i)}) \) denotes the triple corresponding to index \( i \) as defined in Table~\ref{tab:q_enumeration}.\footnote{Hence, \( r_i \) is not a decision variable but a deterministic function of \( q \).}

As in \texttt{ATE\_entropybounds} (see Section~\ref{sec:alg-entropy-ATE}), the interesting part is the final constraint, which we again refer to as the \emph{entropy constraint}. The derivation is slightly different in this case, so we will walk through it step by step.

\paragraph{Joint Counterfactuals and the DPI.}

We begin again with the standard DAG (Figure~\ref{fig:ATE_entr_confDag}) and note that \( X \), as before, is a function of \( U \):
\[
X = f_X(U)
\]

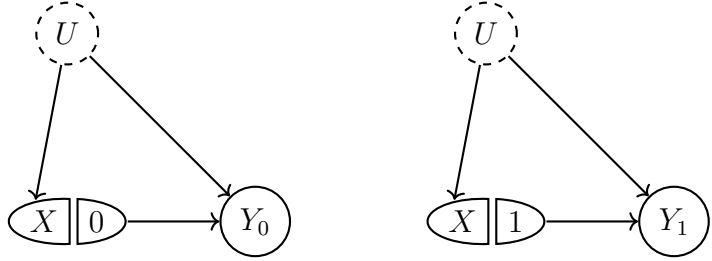
\begin{figure}[H]
\centering
\begin{tikzpicture}[->, node distance=2.5cm, thick]
  \node[name=Xq,shape=swig vsplit, swig vsplit={gap=3pt}]{
    \nodepart{left}{$X$}
    \nodepart{right}{$0$} };
  \node[draw, circle, right of=Xq] (Yx) {\( Y_{0} \)};
  \node[draw, circle, dashed, above of=Xq] (U) {\( U \)};
  \draw (Xq) -- (Yx);
  \draw (U) -- (Yx);
  \draw (U) -- (Xq.142);
\end{tikzpicture}
\hspace{1.5cm}
\begin{tikzpicture}[->, node distance=2.5cm, thick]
  \node[name=Xq,shape=swig vsplit, swig vsplit={gap=3pt}]{
    \nodepart{left}{$X$}
    \nodepart{right}{$1$} };
  \node[draw, circle, right of=Xq] (Yx) {\( Y_{1} \)};
  \node[draw, circle, dashed, above of=Xq] (U) {\( U \)};
  \draw (Xq) -- (Yx);
  \draw (U) -- (Yx);
  \draw (U) -- (Xq.142);
\end{tikzpicture}
\caption{Single-World Intervention Graphs (SWIGs) under the interventions \( X = 0 \) and \( X = 1 \)}
\label{fig:SWIG_X_eq_0}
\end{figure}

We now consider two interventions: \( \doo(X = 1) \) and \( \doo(X = 0) \), visualized by the SWIGs in Figure~\ref{fig:SWIG_X_eq_0}.

We observe that both \( Y_0 \) and \( Y_1 \) are deterministic functions of \( U \):
\[
\begin{aligned}
Y_0 &= f_Y(0, U) \\
Y_1 &= f_Y(1, U)
\end{aligned}
\]

Thus, once \( U \) is fixed, the random vector \( (Y_0, Y_1) \) and the random variable $X$ are fully determined. This implies the conditional independence:
\[
(Y_0, Y_1) \perp X \mid U
\]
which implies the Markov chain:
\[
X \to U \to (Y_0, Y_1)
\]

Applying the Data Processing Inequality (DPI; see Section~\ref{sec:alg-entropy-ATE}) to this chain gives the upper bound:
\[
I(X; (Y_0, Y_1)) \leq I(X; U)
\]

From information theory, we also have \( I(X; U) \leq H(U) \), so:
\begin{equation}
\label{eq:entropy_bound_pns}
I(X; (Y_0, Y_1)) \leq H(U)
\end{equation}

\paragraph{Rewriting the Mutual Information.}

We now express \( I(X; (Y_0, Y_1)) \) in terms of our decision vector \( q \).

Mutual information is related to the Kullback–Leibler divergence:
\[
I(X; (Y_0, Y_1)) = D_{\mathrm{KL}}\left( P((Y_0, Y_1), X) \,\middle\|\, P((Y_0, Y_1)) P(X) \right)
\]

Using the associativity of conjunction, we can write:
\[
I(X; (Y_0, Y_1)) = D_{\mathrm{KL}}\left( P(Y_0, Y_1, X) \,\middle\|\, P(Y_0, Y_1) P(X) \right)
\]

Recognizing that \( P(Y_0, Y_1, X) \) corresponds to \( q \), and \( P(Y_0, Y_1) P(X) \) to \( r \), we substitute into the KL divergence definition:
\[
I(X; (Y_0, Y_1)) = D_{\mathrm{KL}}(q \,\|\, r) = \sum_{i=0}^7 q_i \log\left( \frac{q_i}{r_i} \right)
\]

Combining this with inequality~\ref{eq:entropy_bound_pns} and our parameter $\theta$ yields the entropy constraint for \texttt{PNS\_entropybounds}:
\[
\sum_{i=0}^7 q_i \log\left( \frac{q_i}{r_i} \right) \leq \theta
\]

%% file: chapters/ch-methods/zaffalon.tex
\section{\texttt{zaffalonbounds}}\label{sec:zaffalonbounds}

The method \texttt{zaffalonbounds} implements the \emph{Expectation Maximisation for Causal Computation} (EMCC) algorithm proposed by Zaffalon et al.\ in 2021~\cite{zaffalon_causal_2021}.

EMCC repeatedly \emph{samples} possible structural causal models (SCMs, see Section~\ref{prelim-SCM}) that could fit our observations and then, similar to the regular EM algorithm, alternates between \textit{filling in} the unknown exogenous variables for every data point (E-step) and re-estimating their probability tables to maximize the data likelihood (M-step). 

Restarting this procedure many times from different random seeds lands on different, equally plausible worlds, and the spread of the resulting query values (which are easily computed from a fully specified SCM) forms our causal bound.

Specifically, in our implementation we let the algorithm sample \texttt{runs}~$=30$ SCMs, limiting each EMCC run to at most \texttt{maxiter}~$=100$ iterations (i.e., 100 E- and M-steps).  \footnote{For this we use the Credici and Crema libraries: \\ 
\url{https://github.com/IDSIA/credici} \\\url{https://github.com/IDSIA/crema}}
These hyperparameters are chosen arbitrarily but are based on examples provided by the authors.\footnote{\url{https://github.com/IDSIA-papers/2022-PGM-selection}}

For the remainder of this section, we provide a high-level overview of how the EMCC algorithm works (specific to how it is used in this study):

\begin{enumerate}[leftmargin=*]
    \item \textbf{Define the causal model template.}  
          We begin with a known causal structure — for example, \( U \rightarrow X \rightarrow Y \) and \( U \rightarrow Y \), where all variables are binary and \( U \) is unobserved.  
          Using the \texttt{CausalBuilder} from the Credici library, we build a \emph{partially specified structural causal model}.  
          This model keeps the causal structure fixed but deliberately allows all logically possible mechanisms (e.g., all deterministic functions \( f_Y \colon (X, U) \mapsto Y \)) for how variables influence each other. (This is known as a \emph{conservative} specification.)
 
          At this point, the structural equations are fixed and expressive — what remains unknown is how likely each mechanism is.

    \item \textbf{Prepare multiple starting points.}  
          To explore the space of plausible explanations, we run the EM algorithm multiple times from different initial guesses.  
          Each run starts with a random probability distribution over the latent variable \( U \).  

    \item \textbf{EM loop (up to \texttt{maxiter} = 100).}  
          Each EM run iteratively refines its guess for the distribution of \( U \) to better match the data.
          \begin{itemize}[nosep]
              \item \emph{E-step:}  
                    For every observation in the dataset, the model computes how likely each latent state \( u \) is to have produced that data point, given the current guess for \( P(U) \) and the structural equations.  
                    This computation uses the fixed structure of the model and is handled by the Credici inference engine.  
                    Only those latent states that are consistent with the observed outcome receive non-zero probability.

              \item \emph{M-step:}  
                    The model then updates the distribution over \( U \) by aggregating these soft assignments across all observations.  
                    In effect, it re-weights how likely each mechanism is, based on how well it explains the data.
          \end{itemize}
          This process continues until the update becomes negligible or the maximum number of steps is reached.  
          The result is a fully specified SCM: a fixed structure with a learned distribution over \( U \) that makes the model exactly reproduce the observed data.

    \item \textbf{Collect compatible models.}  
          After 30 runs (each from a different random starting point), we obtain 30 different fully specified models.\footnote{Zaffalon et al.\ show that around 20 EMCC runs are typically sufficient to approximate the sharp bounds with high accuracy~\cite{zaffalon_causal_2021}.}  
          All of them are \emph{model-compatible} — meaning they share the same causal graph and can reproduce the empirical distribution of the observed variables.  
          They differ only in how they distribute probability across the possible latent mechanisms.  
          Together, they form a representative sample of all plausible models given the data and assumptions.

    \item \textbf{Compute bounds.}  
          For each of these models, we compute the causal quantity of interest — either the Average Treatment Effect (ATE) or the Probability of Necessity and Sufficiency (PNS).  
          Since each model is now fully specified, the causal query can be evaluated exactly.  
          The final output is the range of these values across all runs: the smallest and largest values define the lower and upper bounds.  

\end{enumerate}

We refer to this algorithm as \texttt{ATE\_zaffalonbounds} and \texttt{PNS\_zaffalonbounds} when using it to compute bounds for ATE and PNS, respectively.

%% file: chapters/ch-methods/zhangbareinboim.tex
\section{\texttt{zhangbareinboim}}

The method \texttt{ATE\_zhangbareinboim} implements the approach of Zhang and Bareinboim, proposed in their 2021 work~\cite{zhang_bounding_2021}, which derives \emph{sharp nonparametric bounds} on the Average Treatment Effect in the presence of unobserved confounding and bounded continuous outcomes \( Y \in [0,1] \). Unlike the other algorithms introduced, which apply only to discrete outcomes, this method directly accommodates continuous-valued data under a specific instrumental variable (IV) structure.

The assumed causal model is the classical IV setting:
\begin{center}
\begin{tikzpicture}[->, node distance=2cm, thick]
  \node[draw, circle] (Z) {\( Z \)};
  \node[draw, circle, right of=Z] (X) {\( X \)};
  \node[draw, circle, right of=X] (Y) {\( Y \)};
  \node[draw, circle, dashed, above of=X] (U) {\( U \)};
  \draw (Z) -- (X);
  \draw (X) -- (Y);
  \draw (U) -- (X);
  \draw (U) -- (Y);
\end{tikzpicture}
\end{center}

Here, \( Z \) is a randomized instrument affecting the treatment \( X \), which in turn influences the outcome \( Y \). The unobserved confounder \( U \) affects both \( X \) and \( Y \), rendering the ATE non-identifiable from \( P(X, Y \mid Z) \) alone. While \( Y \in [0,1] \), $Z$ and $X$ are still assumed to be categorical variables.

The key innovation of this method is that it avoids discretizing the outcome variable. Rather than modeling the full joint distribution over potential outcomes (which would be infinite-dimensional for continuous \( Y \)), the method bounds the expected potential outcomes \( \mathbb{E}[Y_x] \) directly.

To do so, the SCM is reformulated in terms of a finite latent distribution \( P(x_Z, y_X) \), where:
\begin{itemize}
    \item \( x_Z \) encodes a unit’s treatment response type: a vector specifying the value that treatment \( X \) would take under each possible value of the instrument \( Z \). For example, if \( Z \in \{0,1\} \), then \( x_Z = (x_0, x_1) \) describes what treatment the unit would receive under \( Z=0 \) and under \( Z=1 \),
    \item \( y_X \) encodes a unit’s outcome response type: a vector specifying the value that outcome \( Y \) would take under each possible value of the treatment \( X \), e.g., \( y_X = (y_0, y_1) \).
\end{itemize}

These vectors play the role of latent response types (see, e.g., Section~\ref{alg-autobound}). Importantly, even though \( Y \) is continuous, the optimization is over the distribution of these finite types, making the problem tractable.

The method then constructs a linear program to compute sharp bounds on \( \mathbb{E}[Y_x] \) by optimizing over all joint distributions \( P(x_Z, y_X) \) that are consistent with the observed data. Specifically, the observed conditional distribution \( P(x, y \mid z) \) must satisfy:
\[
P(x, y \mid z) = \sum_{x_Z, y_X} \mathbb{I}[x_Z[z] = x] \cdot \mathbb{I}[y_x = y] \cdot P(x_Z, y_X)
\]
Here, the notation \( x_Z[z] \) refers to the treatment that a unit with response type \( x_Z \) would receive if the instrument took value \( z \). The indicator function \( \mathbb{I}[\cdot] \) evaluates to 1 if the condition is true, and 0 otherwise.

The objective of the linear program is to minimize or maximize the expected counterfactual outcome:
\[
\mathbb{E}[Y_x] = \sum_{x_Z, y_X} y_x \cdot P(x_Z, y_X)
\]
subject to the above consistency constraint, along with normalization and non-negativity:
\[
\sum_{x_Z, y_X} P(x_Z, y_X) = 1, \quad P(x_Z, y_X) \geq 0.
\]

Because outcomes are bounded in \([0,1]\), each term \( y_x \cdot P(x_Z, y_X) \) lies within a finite range. As a result, the entire program remains finite-dimensional and can be solved efficiently, without requiring any discretization of the outcome. This yields valid and sharp bounds on \( \mathbb{E}[Y_x] \), and therefore on the ATE:
\[
\mathrm{ATE} = \mathbb{E}[Y_1] - \mathbb{E}[Y_0].
\]

\noindent
Putting everything together, the bounds on \( \mathbb{E}[Y_x] \) are obtained by solving the following linear program:
\begin{alignat*}{2}
\text{min / max} \quad & \sum_{x_Z, y_X} y_x \cdot P(x_Z, y_X) \\
\text{s.t.} \quad
& \sum_{x_Z, y_X} \mathbb{I}[x_Z[z] = x] \cdot \mathbb{I}[y_x = y] \cdot P(x_Z, y_X) = P(x, y \mid z) \quad \forall x, y, z \\
& \sum_{x_Z, y_X} P(x_Z, y_X) = 1 \\
& P(x_Z, y_X) \geq 0 \quad \forall x_Z, y_X
\end{alignat*}

\noindent
We refer to this implementation as \texttt{ATE\_zhangbareinboim}, and apply it in IV-type scenarios with continuous outcomes, such as \texttt{ContIV} (see Chapter~\ref{ch-simulation_and_experiment}).

%% file: chapters/ch-simulation_and_experiment.tex
\chapter{Simulation and Experiment}
\label{ch-simulation_and_experiment}

This chapter introduces the simulation-based evaluation of the bounding algorithms presented in Chapter~\ref{ch-considered_algorithms}. We begin by formalizing a general framework for simulating causal models and evaluating bounding algorithms. Building on this foundation, we define a set of evaluation metrics and construct a suite of synthetic scenarios, each derived from a structural causal model (SCM). Specifically, we consider four core scenarios: \textit{Binary Confounding}, \textit{Binary Instrumental Variable}, \textit{Continuous Confounding}, and \textit{Continuous Instrumental Variable}. 

Each scenario gives rise to a large number of randomized simulations, where each simulation represents a dataset on which the algorithms are tested. Simulated experiments provide full control over the data-generating process, enabling systematic analysis and direct comparison against known ground truth values. Our goal is to assess each algorithm’s informativeness, validity, and robustness across a range of causal conditions.

In total, the evaluation comprises 188{,}000 algorithm runs.

\section{Simulation Framework and Notation}
\label{sec:simulation-framework-and-notation}

For each scenario in this chapter, we generate a dataset \(\mathcal{D}\) consisting of \(N = 2000\) independently simulated causal models, indexed by \(j \in \{1, \dots, N\}\). Each simulation \(\mathcal{S}_j\) samples its own structural parameters (e.g., coefficients, intercepts), which remain fixed within \(\mathcal{S}_j\) but vary across simulations.

Each simulation then produces \(n = 500\) unit-level observations:
\[
\mathcal{S}_j = \{ r^{(j)}_1, \dots, r^{(j)}_n \},
\]
where each row \(r^{(j)}_i\) represents a single instance drawn from the model defined by \(\mathcal{S}_j\).

The full dataset \(\mathcal{D}\) is thus:
\[
\mathcal{D} = \{ \mathcal{S}_1, \dots, \mathcal{S}_N \}, \qquad \text{with} \quad |\mathcal{S}_j| = n.
\]

We define \(\mathcal{A}\) as the set of bounding algorithms considered in our evaluation. Each algorithm \(a \in \mathcal{A}\) is treated as a mapping:
\[
a : \mathcal{S}_j \longrightarrow \mathbb{R}^2,
\]
where \(a(\mathcal{S}_j) = \bigl( \texttt{lower}(j), \texttt{upper}(j) \bigr)\) denotes the bound returned by algorithm \(a\) when applied to simulation \(j\).

This assumes the algorithm executes successfully. In cases of numerical failure or violated assumptions, the algorithm may produce no output. Such cases are treated as failures and will be formally defined in Section~\ref{evaluation_metrics}.

\section{Trivial Ceilings}
\label{sec:trivialceilings}

To enable a fair comparison, all algorithm outputs are \emph{clipped to lie within} their respective \emph{trivial ceilings}. This ensures that no algorithm returns bounds that exceed the logically possible range of the target causal query.

We define the trivial ceiling interval for each causal query as follows:
\begin{itemize}
    \item Average Treatment Effect (ATE): \( [-1, 1] \) (assuming binary treatment and outcome bounded in \([0, 1]\))
    \item Probability of Necessity and Sufficiency (PNS): \( [0, 1] \)
\end{itemize}

Let \( a \in \mathcal{A} \) be a bounding algorithm, and let \( a(\mathcal{S}_j) = (\texttt{lower}_a(j), \texttt{upper}_a(j)) \in \mathbb{R}^2 \) be the bound produced for simulation \( \mathcal{S}_j \in \mathcal{D} \).

If the algorithm fails to return a bound, or if the bound lies outside the trivial range, we apply the following clipping rule:
\begin{align*}
\texttt{lower}_a(j) &:= \max(\texttt{lower}_a(j),\ \texttt{trivialLower}) \\
\texttt{upper}_a(j) &:= \min(\texttt{upper}_a(j),\ \texttt{trivialUpper})
\end{align*}

This step prevents algorithms that return out-of-range bounds from being unfairly penalized during comparison — for example, when computing the average distance between the upper and lower bound.

\input{chapters/ch-simulation_and_experiment/evaluation_metrics}

\section{Naming Convention}
\label{sec:naming-convention}

Throughout this study, algorithm names are written in lowercase (e.g., \texttt{causaloptim}), while scenario names follow CamelCase (e.g., \texttt{BinaryIV}).

When referring to a specific application of an algorithm to a causal query (such as the ATE or PNS), we prepend the query name as a prefix. For instance, \texttt{ATE\_causaloptim} denotes the application of the \texttt{causaloptim} algorithm to the Average Treatment Effect (ATE).

If the algorithm requires a parameter, this is indicated via a suffix. For example, \texttt{ATE\_entropybounds-0.80} refers to the \texttt{entropybounds} algorithm applied to the ATE with a parameter value of \( 0.80 \). All algorithms considered in this study accept at most one parameter.

Most algorithms are restricted to discrete outcome spaces. In scenarios involving continuous outcomes, we apply a binarization step where values greater than 0.5 are mapped to 1, and values less than or equal to 0.5 to 0. If such binarization is applied, an additional suffix \texttt{--binned} is used. For instance, \texttt{ATE\_entropybounds-0.80--binned} denotes the binarized version of the same algorithm.

\paragraph{Scenario Naming.}  
We use the following naming scheme for scenarios:
\begin{itemize}
    \item Binary Confounding: \texttt{BinaryConf}
    \item Binary Instrumental Variable: \texttt{BinaryIV}
    \item Continuous Confounding: \texttt{ContConf}
    \item Continuous Instrumental Variable: \texttt{ContIV}
    \item Binary Entropy Confounding: \texttt{BinaryEntropyConf}
\end{itemize}

The term \emph{core scenarios} refers to the four primary cases: \texttt{BinaryConf}, \texttt{BinaryIV}, \texttt{ContConf}, and \texttt{ContIV}, which are directly comparable in structure and design.

\section{Parameter Selection}
\label{sec:parameter-selection}

We now briefly describe how parameters are chosen for the algorithms that require them.

\subsection{\texttt{entropybounds}}  
The \texttt{entropybounds} algorithm requires a single parameter, denoted by \(\theta\), which specifies an upper bound on the entropy of the unobserved confounder (see Section~\ref{sec:methods-entropybounds}).  
We evaluate both \texttt{ATE\_entropybounds} and \texttt{PNS\_entropybounds} using the following values:
\[
\theta \in \{0.1, 0.2, 0.8, \texttt{trueTheta}, \texttt{randomTheta}\}
\]
Here, \texttt{trueTheta} denotes the true entropy of the unobserved confounder in a given simulation, while \texttt{randomTheta} is sampled from a uniform distribution: \(\texttt{randomTheta} \sim \text{Uni}(0,1)\). The fixed values (0.1, 0.2, 0.8) and \texttt{randomTheta} are chosen independently of the true confounder entropy, allowing us to assess the algorithm's sensitivity to possible misspecifications of \(\theta\).

\subsection{\texttt{OLS} and \texttt{2SLS}}  
For both \texttt{ATE\_OLS} and \texttt{ATE\_2SLS}, we vary the confidence level, commonly denoted by \(1 - \alpha\) (see Section~\ref{sec:methods-2SLS_OLS}). The parameter is selected from the set:
\[
1 - \alpha \in \{0.95, 0.98, 0.99\}
\]

\section{Scenarios and Results}

This section presents the core experimental results of the thesis. For each scenario introduced below, we generate a dedicated simulation dataset \(\mathcal{D} = \{ \mathcal{S}_1, \dots, \mathcal{S}_N \}\), consisting of \(N = 2000\) independently simulated causal models. Each simulation \(\mathcal{S}_j\) generates \(n = 500\) unit-level observations according to a scenario-specific data-generating process.

Each simulation is evaluated using a set of bounding algorithms \(\mathcal{A}\), where each algorithm \(a \in \mathcal{A}\) maps a simulation \(\mathcal{S}_j\) to a bound \(a(\mathcal{S}_j) = (\texttt{lower}_a(j), \texttt{upper}_a(j)) \in \mathbb{R}^2\).

Each scenario is presented in two subsections:
\begin{enumerate}
    \item \textbf{Data-Generating Process:} A detailed description of how simulations \(\mathcal{S}_j\) are generated for the scenario.
    \item \textbf{Results:} A quantitative evaluation of the considered algorithms' performance using the metrics defined in Section~\ref{evaluation_metrics}.
\end{enumerate}

For the \texttt{BinaryConf} scenario, we additionally resample a specific simulation \(\mathcal{S}_j\) multiple times and conduct a variance analysis of the resulting bounds to illustrate the robustness of our findings.

In the \texttt{BinaryEntropyConf} scenario, we further perform a detailed sensitivity analysis to investigate the impact of parameter misspecification. In particular, we examine how the performance of the \texttt{entropybounds} algorithm is affected when it's parameter \(\theta\) is \emph{underspecified} (i.e., chosen too low).

A detailed interpretation and comparison of results across scenarios will be provided in Chapter~\ref{ch-discussion}.

\let\origsection\section
\let\origsubsection\subsection
\let\origsubsubsection\subsubsection

\let\section\subsection
\let\subsection\subsubsection
\let\subsubsection\paragraph
\input{chapters/ch-simulation_and_experiment/binaryConf/index}

\input{chapters/ch-simulation_and_experiment/binaryIV/index}

\input{chapters/ch-simulation_and_experiment/contConf/index}

\input{chapters/ch-simulation_and_experiment/contIV/index}

\input{chapters/ch-simulation_and_experiment/entropyConf/index}

\let\section\origsection
\let\subsection\origsubsection
\let\subsubsection\origsubsubsection


\section{Runtimes}
\label{sec:runtimes}

All experiments were executed on a Google Compute Engine (GCE) virtual machine of type \texttt{n1-standard-4}, equipped with 4 virtual CPUs and 15~GB of memory.

Table~\ref{tab:avg-runtimes} reports the average runtime per algorithm, aggregated across all queries, parameter settings, and scenarios. These values offer a high-level impression of computational efficiency across the considered methods. A more detailed breakdown of runtimes for individual configurations is provided in Appendix~\ref{appendix:runtimes}.

\textbf{Note:} Simulations involving binarized (see Section~\ref{contConf_dataGen}) outcome data are excluded from the average runtime calculations.

\textbf{Remark:} The \texttt{zaffalonbounds} algorithm is the only method for which multi-threading was implemented. Specifically, the dataset is split into chunks equal to the number of available CPU cores and processed in parallel. All other algorithms were executed in a single-threaded fashion.
\begin{table}[H]
\centering
\begin{tabular}{|l|c|}
\hline
Algorithm & Runtime (s) \\
\hline
\texttt{manski} & 1  \\ \hline
\texttt{tianpearl} & 1  \\ \hline
\texttt{OLS} & 5  \\ \hline
\texttt{2SLS} & 20  \\ \hline
\texttt{entropybounds} & 61  \\ \hline
\texttt{zhangbareinboim} & 63  \\ \hline
\texttt{autobound} & 85  \\ \hline
\texttt{causaloptim} & 373  \\ \hline
\texttt{zaffalonbounds} & 4473  \\ \hline
\end{tabular}
\caption{Average runtimes per algorithm (in seconds)}
\label{tab:avg-runtimes}
\end{table}

%% file: chapters/ch-simulation_and_experiment/evaluation_metrics.tex
\section{Evaluation Metrics}
\label{evaluation_metrics}

We now define evaluation metrics applied to the set of simulations \(\mathcal{D} = \{ \mathcal{S}_1, \dots, \mathcal{S}_N \}\), evaluated under each bounding algorithm \( a \in \mathcal{A} \). 
Each metric summarizes how well an algorithm performs across the entire dataset \(\mathcal{D}\) for a given scenario and causal query.

In general, when an algorithm is executed on a simulation, one of the following outcomes occurs:
\begin{itemize}
    \item \textbf{Failed:} The algorithm does not return a bound (e.g., due to numerical errors or violated assumptions). Denote the set of failed simulations by \( F_a \subseteq \{1, \dots, N\} \).
    \item \textbf{Invalid:} The algorithm returns a bound, but it does not contain the true value. Denote this set by \( I_a \subseteq \{1, \dots, N\} \setminus F_a \).
    \item \textbf{Valid:} The bound contains the true query value. Denote this set by \( V_a = \{1, \dots, N\} \setminus (F_a \cup I_a) \).
\end{itemize}

\paragraph{Query Range.}  
To improve comparability between the two causal queries, ATE and PNS, we normalize all metrics by the logically possible range of the respective query (cf. Section~\ref{sec:trivialceilings}).  
We define the normalization constant \( R \) as:
\[ R = 
\begin{cases}
2, & \text{for ATE} \\
1, & \text{for PNS}
\end{cases}
\]

\paragraph{Net Bound Width.}\footnote{May be referred to as \emph{Net Width} for brevity.}  
The average width computed over only the valid simulations:
\[
\text{Net Bound Width}_a = \frac{1}{|V_a|} \sum_{j \in V_a} \left( \frac{\texttt{upper}_a(j) - \texttt{lower}_a(j)}{R} \cdot 100 \right)
\]

\paragraph{Bound Width.}  
The average width computed over all simulations, with invalid and failed instances penalized.\footnote{This metric should be interpreted with caution, as in practice we typically do not know whether the bounds contain the true value.}
We first define the penalized width \( t_a(j) \) for algorithm \(a\) on simulation \(j\) as:
\[
t_a(j) =
\begin{cases}
\texttt{trivialUpper} - \texttt{trivialLower}, & j \in F_a \cup I_a \\
\texttt{upper}_a(j) - \texttt{lower}_a(j), & j \in V_a
\end{cases}
\]
Then, the bound width is given by:
\[
\text{Bound Width}_a = \frac{1}{N} \sum_{j=1}^{N} \left( \frac{t_a(j)}{R} \cdot 100 \right)
\]

\paragraph{Failure Rate.}  
\[
\text{Failure Rate}_a = \frac{|F_a|}{N} \cdot 100
\]

\paragraph{Invalid Rate.}  
\[
\text{Invalid Rate}_a = \frac{|I_a|}{N - |F_a|} \cdot 100
\]

\paragraph{Invalid \(\Delta\).}  
Let \(\texttt{true}(j) \in \mathbb{R}\) denote the ground-truth value of the query in simulation \(\mathcal{S}_j\). Define the violation amount \(\delta_a(j)\) as:
\[
\delta_a(j) =
\begin{cases}
\texttt{lower}_a(j) - \texttt{true}(j), & \texttt{true}(j) < \texttt{lower}_a(j) \\
\texttt{true}(j) - \texttt{upper}_a(j), & \texttt{true}(j) > \texttt{upper}_a(j) \\
0, & \text{otherwise}
\end{cases}
\]
Then, the average normalized violation over all invalid simulations is:
\[
\text{Invalid } \Delta_a = \frac{1}{|I_a|} \sum_{j \in I_a} \left( \frac{\delta_a(j)}{R} \cdot 100 \right)
\]

\bigskip

Together, these metrics capture three key evaluation axes: \textit{robustness} (failure and invalidity), \textit{informativeness} (bound width), and \textit{severity of errors} (invalid deviation). The normalization by the query range \(R\) improves comparability between results for the ATE and the PNS.

\begin{table}[H]
\centering
\caption{
Evaluation metrics for algorithm \(a \in \mathcal{A}\), evaluated over simulations \(j = 1, \dots, N\), where \(a(\mathcal{S}_j)\) returns a bound \((\texttt{lower}_a(j), \texttt{upper}_a(j))\). All metrics are expressed as percentages.
}
\label{tab:evaluation-metrics}
\resizebox{\textwidth}{!}{
\begin{tabular}{@{}llp{9cm}@{}}
\toprule
\textbf{Metric} & \textbf{Formula} & \textbf{Description} \\ \midrule

\textbf{Failure Rate} &
\(\displaystyle \frac{|F_a|}{N} \cdot 100\) &
Percentage of simulations where algorithm \(a\) fails to return a bound (e.g., due to numerical issues or unmet assumptions). \\

\textbf{Invalid Rate} &
\(\displaystyle \frac{|I_a|}{N - |F_a|} \cdot 100\) &
Percentage of non-failed simulations where the true value lies outside the returned bound. \\

\textbf{Bound Width} &
\(\displaystyle \frac{1}{N} \sum_{j=1}^{N} \left( \frac{t_a(j)}{R} \cdot 100 \right)\) &
Average normalized width over all simulations. Failed or invalid runs are penalized by assigning maximal (trivial) width. Attempts to measure overall performance. \\

\textbf{Net Bound Width} &
\(\displaystyle \frac{1}{|V_a|} \sum_{j \in V_a} \left( \frac{\texttt{upper}_a(j) - \texttt{lower}_a(j)}{R} \cdot 100 \right)\) &
Average normalized width over valid simulations only. Reflects the sharpness of bounds when they are known to be valid. \\

\textbf{Invalid \(\Delta\)} &
\(\displaystyle \frac{1}{|I_a|} \sum_{j \in I_a} \left( \frac{\delta_a(j)}{R} \cdot 100 \right)\) &
Average normalized deviation of invalid bounds from the true value. Quantifies the magnitude of error when bounds fail to contain the target. \\

\bottomrule
\end{tabular}
}
\end{table}

%% file: chapters/ch-simulation_and_experiment/binaryConf/index.tex
\section{Binary Confounding}
\label{sec:scenario-binConf}

This scenario introduces a basic confounding structure involving binary variables. A binary treatment variable \(X\) has a causal effect on an outcome \(Y\), while both are influenced by an unobserved binary confounder \(U\). The corresponding causal diagram is shown below:

\begin{center}
\begin{tikzpicture}[->, node distance=2cm, thick]
  \node[draw, circle] (X) {\( X \)};
  \node[draw, circle, right of=X] (Y) {\( Y \)};
  \node[draw, circle, dashed, above of=X] (U) {\( U \)};
  
  \draw (X) -- (Y);
  \draw (U) -- (X);
  \draw (U) -- (Y);
\end{tikzpicture}
\end{center}

The goal is to bound two causal quantities: the Average Treatment Effect (ATE) and the Probability of Necessity and Sufficiency (PNS), defined as
\[
\text{ATE} = P(Y_{X=1} = 1) - P(Y_{X=0} = 1), \qquad
\text{PNS} = P(Y_{X=1} = 1,\, Y_{X=0} = 0).
\]

Each simulation \(\mathcal{S}_j \in \mathcal{D}\) samples a fixed set of structural parameters and link functions (e.g., \(\alpha_X\), \(\beta_{U \to X}\), \(f_X\)), and generates \(n = 500\) unit-level observations \((X_i, Y_i, U_i)\) accordingly.

\input{chapters/ch-simulation_and_experiment/binaryConf/dataGen}

\input{chapters/ch-simulation_and_experiment/binaryConf/results}

\input{chapters/ch-simulation_and_experiment/binaryConf/varianceCheck}

%% file: chapters/ch-simulation_and_experiment/binaryConf/dataGen.tex
\subsection{Data-Generating Process}
\label{binConf_dataGen}

We simulate data from a binary structural causal model (SCM) that features \textbf{observation-level heteroskedasticity}, \textbf{simulation-specific coefficents}, and \textbf{varying link functions}.  
All variables are binary:
\[
U_i,\,X_i,\,Y_i \in \{0,1\}.
\]

\paragraph{Root variable.}  
The unobserved confounder for each unit \(i\) is generated as
\[
U_i \sim \text{Bern}(p_U),
\]
with \(p_U \sim \text{Uni}(0,1)\).
This leads to $U$ having an average entropy of roughly $0.68$.

\paragraph{Observation-level heteroskedastic noise.}  
Each unit \(i\) draws a private noise scale:
\[
\sigma_i \sim \left|\mathcal{N}(0,1)\right|,
\]
and generates two zero-mean Gaussian noise terms with unit-specific variance:
\[
\varepsilon_{X_i},\, \varepsilon_{Y_i} \sim \mathcal{N}(0,\sigma_i^{2}).
\]

\paragraph{Structural equations.}  
Each simulation draws intercepts from a standard normal distribution:
\[
\alpha_X,\ \alpha_Y \sim \mathcal{N}(0,1),
\]
and edge weights from a bimodal distribution\footnote{This is done to avoid coefficients being centered around zero.}
\[
\beta_{U\!\to\!X},\,
\beta_{U\!\to\!Y}
\;\sim\;
\tfrac12\,\mathcal{N}(1,0.5^{2})+\tfrac12\,\mathcal{N}(-1,0.5^{2}),
\]
with \(\beta_{X\!\to\!Y} \in [-5,5]\) assigned deterministically per simulation\footnote{\label{fn:grid-conf}%
For simulation \(j\in\{1,\dots,N\}\), set \(\beta_{X\!\to\!Y}^{(j)} = -5 + \tfrac{10\,(j-1)}{N-1}\), tiling \([-5,5]\) uniformly.}.  
We define the linear predictors:
\[
X_i^*= \alpha_X
          + \beta_{U\!\to\!X} U_i
          + \varepsilon_{X_i},
\qquad
Y_i^*= \alpha_Y
          + \beta_{X\!\to\!Y} X_i
          + \beta_{U\!\to\!Y} U_i
          + \varepsilon_{Y_i}.
\]

\paragraph{Squashing functions.}  
For each simulation, we sample two functions \(f_X, f_Y \in \mathcal{F}\) independently and uniformly from a fixed set of smooth, bounded functions:
\[
\mathcal{F} = \left\{
x \mapsto \frac{1}{1 + e^{-x}},\quad
x \mapsto \frac{1}{2}(1 + \tanh(x)),\quad
x \mapsto \frac{\log(1 + e^x)}{1 + \log(1 + e^x)},\quad
x \mapsto \Phi(x)
\right\},
\]
where \(\Phi(x)\) denotes the CDF of the standard normal distribution (probit).

\paragraph{Treatment and outcome sampling.}  
\[
X_i \sim \text{Bern}\!\bigl(f_X(X_i^*)\bigr),\qquad
Y_i \sim \text{Bern}\!\bigl(f_Y(Y_i^*)\bigr).
\]

\paragraph{Counterfactual outcomes.}  
The potential outcomes for each unit are defined as:
\[
\begin{aligned}
P\!\bigl(Y_{X=1,i}=1\bigr) &=
    f_Y\!\bigl(\alpha_Y+\beta_{X\!\to\!Y}+\beta_{U\!\to\!Y}U_i+\varepsilon_{Y_i}\bigr),\\
P\!\bigl(Y_{X=0,i}=1\bigr) &=
    f_Y\!\bigl(\alpha_Y+\beta_{U\!\to\!Y}U_i+\varepsilon_{Y_i}\bigr).
\end{aligned}
\]

\paragraph{Ground-truth causal quantities.}  
\[
\text{ATE}
   = P\!\bigl(Y_{X=1}=1\bigr)-P\!\bigl(Y_{X=0}=1\bigr),\qquad
\text{PNS}
   = P\!\bigl(Y_{X=1}=1,\;Y_{X=0}=0\bigr).
\]
(Note: the above refer to population-level probabilities averaged over all units.)

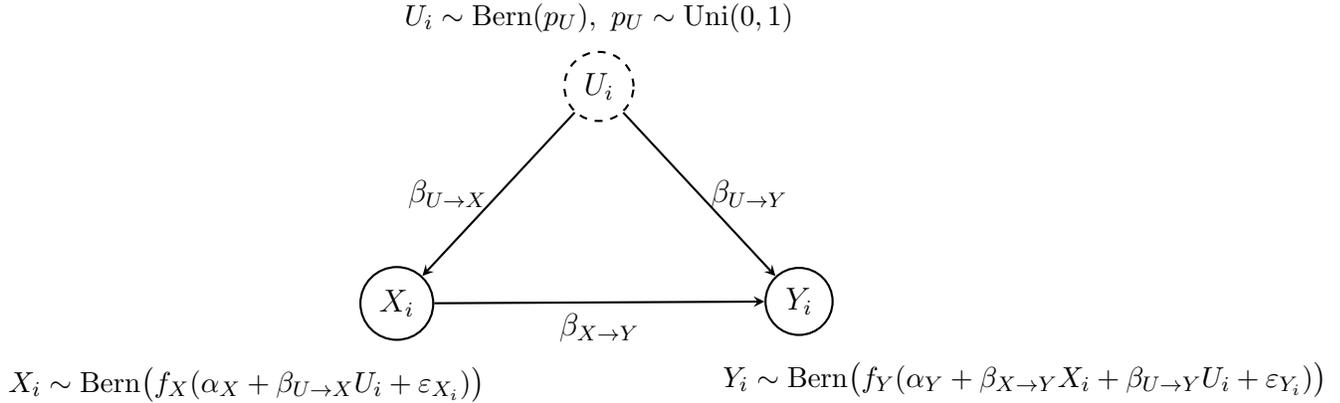
\begin{figure}[H]
\centering
\begin{tikzpicture}[>=stealth, node distance=2.8cm, thick]

\tikzstyle{latent} = [draw, circle, dashed]
\tikzstyle{obs} = [draw, circle]
\tikzstyle{vartext} = [align=center, font=\small]

\node[latent] (U) {\(U_i\)};
\node[obs, below left=2.2cm and 2.0cm of U] (X) {\(X_i\)};
\node[obs, below right=2.2cm and 2.0cm of U] (Y) {\(Y_i\)};

\node[vartext, above=0.1cm of U] {\(U_i \sim \mathrm{Bern}(p_U),\ p_U \sim \mathrm{Uni}(0,1)\)};
\node[vartext, below=0.2cm of X, xshift=-2.0cm] {\(X_i \sim \mathrm{Bern}\bigl(f_X(\alpha_X + \beta_{U \to X} U_i + \varepsilon_{X_i})\bigr)\)};
\node[vartext, below=0.2cm of Y, xshift=3.0cm] {\(Y_i \sim \mathrm{Bern}\bigl(f_Y(\alpha_Y + \beta_{X \to Y} X_i + \beta_{U \to Y} U_i + \varepsilon_{Y_i})\bigr)\)};

\draw[->] (U) -- (X) node[midway, left] {\(\beta_{U \to X}\)};
\draw[->] (U) -- (Y) node[midway, right] {\(\beta_{U \to Y}\)};
\draw[->] (X) -- (Y) node[midway, below] {\(\beta_{X \to Y}\)};

\end{tikzpicture}
\caption{Binary SCM with confounding: causal graph and data-generating process.}
\label{fig:binaryConf-dgp-annotated}
\end{figure}

%% file: chapters/ch-simulation_and_experiment/binaryConf/results.tex
\subsection{Results}

We now present the simulation results for the \texttt{BinaryConf} scenario—first for the ATE, then for the PNS.

\paragraph*{ATE}~

All ATE algorithms are applicable in the \texttt{BinaryConf} scenario, except for \texttt{2SLS}, which requires an instrument, and \texttt{zhangbareinboim}, which requires both an instrumental variable and a continuous outcome.

Table~\ref{tab:binaryconf-ate-results} reports the evaluation metrics for each applicable algorithm (see Section~\ref{evaluation_metrics} for definitions). We emphasize the \textbf{Invalid Rate} and \textbf{Net Width} columns, as they reflect the reliability and informativeness of each method.

\begin{table}[h]
\centering
\caption{Results for ATE in the \texttt{BinaryConf} scenario.}
\label{tab:binaryconf-ate-results}
\scriptsize
\renewcommand{\arraystretch}{1.1}
\begin{tabularx}{\textwidth}{Xrrrrr@{\hskip 6pt}}
\toprule
\textbf{Algorithm} & \textbf{Fail Rate} & \textbf{Invalid Rate} & \textbf{Net Width} & \textbf{Bound Width} & \textbf{Invalid $\Delta$} \\
\midrule
\texttt{OLS-0.95}                   & 0.00 & \textbf{13.40} & \textbf{8.08}  & 20.40 & 1.20 \\
\texttt{OLS-0.98}                   & 0.00 & \textbf{7.70}  & \textbf{9.55}  & 16.52 & 1.12 \\
\texttt{OLS-0.99}                   & 0.00 & \textbf{5.05}  & \textbf{10.55} & 15.07 & 1.13 \\
\texttt{entropybounds-0.10}        & 0.00 & \textbf{20.50} & \textbf{32.32} & 46.20 & 4.88 \\
\texttt{zaffalonbounds}            & 0.00 & \textbf{4.05}  & \textbf{33.05} & 35.76 & 1.04 \\
\texttt{entropybounds-0.20}        & 0.00 & \textbf{10.00} & \textbf{40.58} & 46.52 & 4.27 \\
\texttt{entropybounds-randomTheta} & 0.00 & \textbf{10.50} & \textbf{45.24} & 50.99 & 5.21 \\
\texttt{entropybounds-trueTheta}   & 0.00 & \textbf{7.75}  & \textbf{47.82} & 51.87 & 4.86 \\
\texttt{entropybounds-0.80}        & 0.00 & \textbf{6.80}  & \textbf{49.79} & 53.20 & 4.89 \\
\texttt{autobound}                 & 0.00 & \textbf{0.05}  & \textbf{50.00} & 50.02 & 0.00 \\
\texttt{causaloptim}               & 0.00 & \textbf{0.05}  & \textbf{50.00} & 50.02 & 0.00 \\
\texttt{manski}                    & 0.00 & \textbf{0.05}  & \textbf{50.00} & 50.02 & 0.00  \\
\bottomrule
\end{tabularx}
\end{table}

\begin{figure}[H]
    \centering
    \includegraphics[width=0.8\textwidth]{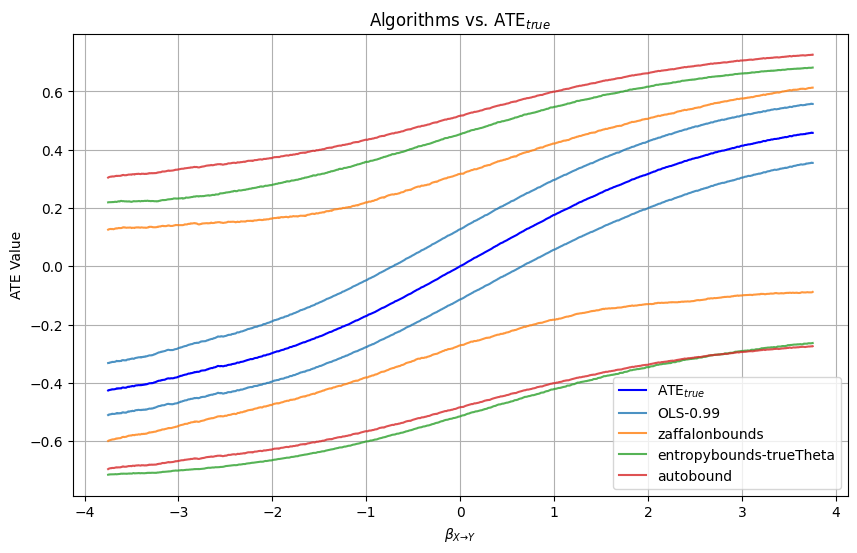}
    \caption{Upper and lower bounds for the ATE of selected algorithms in the \texttt{BinaryConf} scenario, alongside the true ATE. For visualization, lines are smoothed by averaging across 500 samples.}
    \label{fig:binaryConf_algs_vs_ATE}
\end{figure}

Figure~\ref{fig:binaryConf_algs_vs_ATE} visualizes the upper and lower bounds of selected algorithms, along with the true ATE, smoothed by averaging over 500 samples.

\paragraph*{PNS}~

All PNS algorithms under consideration are applicable in the \texttt{BinaryConf} scenario. Table~\ref{tab:binaryconf-pns-results} reports the same evaluation metrics as for the ATE, again emphasizing the \textbf{Invalid Rate} and \textbf{Net Width}.

\begin{table}[H]
\centering
\caption{Results for PNS in the \texttt{BinaryConf} scenario.}
\label{tab:binaryconf-pns-results}
\scriptsize
\renewcommand{\arraystretch}{1.1}
\begin{tabularx}{\textwidth}{Xrrrrr@{\hskip 6pt}}
\toprule
\textbf{Algorithm} & \textbf{Fail Rate} & \textbf{Invalid Rate} & \textbf{Net Width} & \textbf{Bound Width} & \textbf{Invalid $\Delta$} \\
\midrule
\texttt{zaffalonbounds}            & 0.00 & \textbf{0.80}  & \textbf{48.15} & 48.57 & 0.04 \\
\texttt{entropybounds-0.10}        & 0.00 & \textbf{0.05}  & \textbf{49.87} & 49.89 & 0.00 \\
\texttt{causaloptim}               & 0.00 & \textbf{0.00}  & \textbf{51.63} & 51.63 & N/A  \\
\texttt{autobound}                 & 0.00 & \textbf{0.00}  & \textbf{51.63} & 51.63 & N/A  \\
\texttt{tianpearl}                 & 0.00 & \textbf{0.00}  & \textbf{51.63} & 51.63 & N/A  \\
\texttt{entropybounds-0.20}        & 0.00 & \textbf{0.00}  & \textbf{57.57} & 57.57 & N/A  \\
\texttt{entropybounds-randomTheta} & 0.00 & \textbf{0.10}  & \textbf{64.27} & 64.30 & 0.77 \\
\texttt{entropybounds-trueTheta}   & 0.00 & \textbf{0.00}  & \textbf{68.75} & 68.75 & N/A  \\
\texttt{entropybounds-0.80}        & 0.00 & \textbf{0.00}  & \textbf{72.13} & 72.13 & N/A  \\
\bottomrule
\end{tabularx}
\end{table}

\begin{figure}[H]
    \centering
    \includegraphics[width=0.8\textwidth]{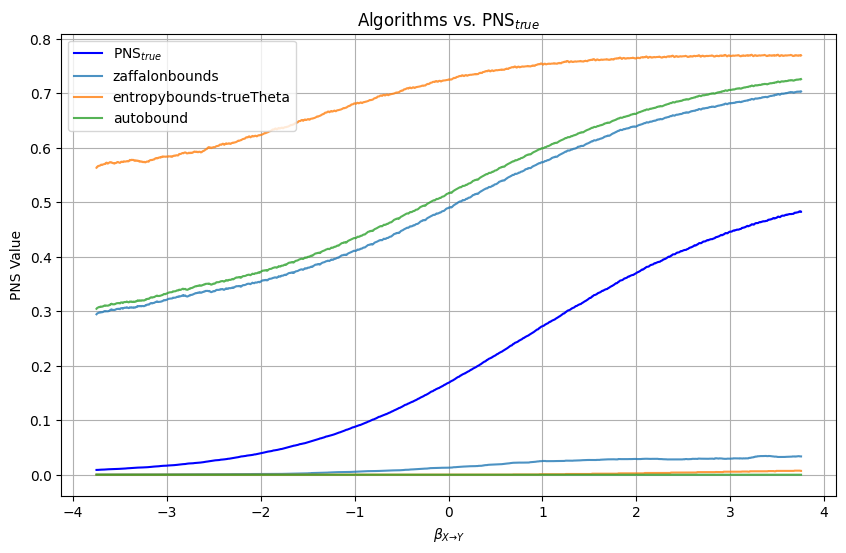}
    \caption{Upper and lower bounds of selected PNS algorithms in the \texttt{BinaryConf} scenario, together with the true PNS. Lines are smoothed by averaging across 500 samples.}
    \label{fig:binaryConf_algs_vs_PNS}
\end{figure}

Figure~\ref{fig:binaryConf_algs_vs_PNS} shows the PNS bounds of a subset of these algorithms, alongside the true PNS values, smoothed in the same manner.

%% file: chapters/ch-simulation_and_experiment/binaryConf/varianceCheck.tex
\subsection{Robustness under Resampling}

To assess the robustness of our simulation results, we perform a variance analysis on the evaluation metrics within a fixed structural causal model \(\mathcal{S}_j\).

To this end, we simulate a single instance \(\mathcal{S}_j\) from the dataset \(\mathcal{D}\), corresponding to the \texttt{BinaryConf} scenario, and replicate it \(M = 2000\) times. In each of the \(M\) repetitions:
\begin{itemize}
    \item the unit-level variables \(U_i, X_i, Y_i\), and
    \item the noise terms \(\sigma_i\) and \(\varepsilon_i\)
\end{itemize}
are independently resampled, while the structural parameters remain fixed across all repetitions.

The fixed structural parameters used throughout this analysis are:
\begin{align*}
\alpha_X &= 0.3, \\
\alpha_Y &= -0.5, \\
\beta_{U \to X} &= 1, \\
\beta_{U \to Y} &= -1, \\
\beta_{X \to Y} &= 2, \\
p_U &= 0.8, \\
f_X &= \frac{1}{1 + e^{-x}}, \\
f_Y &= \frac{1}{1 + e^{-x}}.
\end{align*}
As in all other simulations, we draw \(n = 500\) unit-level observations per repetition.

\vspace{0.5em}
\begin{table}[H]
\centering
\begin{tabular}{llrrrr}
\toprule
Query & Algorithm & Mean LB & Std LB & Mean UB & Std UB \\
\midrule
\multirow{13}{*}{ATE} & OLS-0.99 & 0.252 & \textbf{0.050} & 0.485 & \textbf{0.043} \\
 & OLS-0.98 & 0.263 & \textbf{0.049} & 0.474 & \textbf{0.043} \\
 & OLS-0.95 & 0.280 & \textbf{0.049} & 0.457 & \textbf{0.044} \\
 & causaloptim & -0.336 & \textbf{0.022} & 0.664 & \textbf{0.022} \\
 & autobound & -0.336 & \textbf{0.022} & 0.664 & \textbf{0.022} \\
 & manski & -0.336 & \textbf{0.022} & 0.664 & \textbf{0.022} \\
 & entropybounds-0.80 & -0.402 & \textbf{0.019} & 0.598 & \textbf{0.019} \\
 & entropybounds-0.20 & -0.347 & \textbf{0.024} & 0.569 & \textbf{0.019} \\
 & entropybounds-0.10 & -0.240 & \textbf{0.026} & 0.491 & \textbf{0.023} \\
 & entropybounds-trueTheta & -0.402 & \textbf{0.019} & 0.598 & \textbf{0.019} \\
 & entropybounds-randomTheta & -0.362 & \textbf{0.092} & 0.571 & \textbf{0.071} \\
 & zaffalonbounds & -0.104 & \textbf{0.088} & 0.488 & \textbf{0.057} \\
\midrule
\multirow{9}{*}{PNS} & causaloptim & 0.000 & \textbf{0.000} & 0.664 & \textbf{0.022} \\
 & autobound & 0.000 & \textbf{0.000} & 0.664 & \textbf{0.022} \\
  & tianpearl & 0.000 & \textbf{0.000} & 0.664 & \textbf{0.022} \\
 & entropybounds-0.80 & 0.000 & \textbf{0.000} & 0.924 & \textbf{0.012} \\
 & entropybounds-0.20 & 0.000 & \textbf{0.000} & 0.899 & \textbf{0.017} \\
 & entropybounds-0.10 & 0.000 & \textbf{0.000} & 0.837 & \textbf{0.022} \\
 & entropybounds-trueTheta & 0.000 & \textbf{0.000} & 0.924 & \textbf{0.012} \\
 & entropybounds-randomTheta & 0.006 & \textbf{0.036} & 0.901 & \textbf{0.053} \\
 & zaffalonbounds & 0.036 & \textbf{0.049} & 0.641 & \textbf{0.028} \\
\bottomrule
\end{tabular}
\caption{Mean and standard deviation of lower (LB) and upper (UB) bounds across \(M = 2000\) resampled runs of a fixed SCM \(\mathcal{S}_j\). Standard deviations are highlighted in bold.}
\label{tab:variance-analysis}
\end{table}

\noindent
As shown in Table~\ref{tab:variance-analysis}, both lower and upper bounds remain remarkably stable across repeated runs. The vast majority of algorithms exhibit low standard deviations, typically below 0.05 for both queries. Notable exceptions include \texttt{entropybounds-randomTheta} and \texttt{zaffalonbounds}, which show increased variation. The first still samples its entropy parameter \(\theta\) from a uniform distribution over \([0,1]\), explaining the observed variability. The latter appears to be more sensitive to sampling variation (or repeated execution) compared to other methods, but even here, deviations remain moderate.

\begin{figure}[H]
    \centering
    \includegraphics[width=0.8\textwidth]{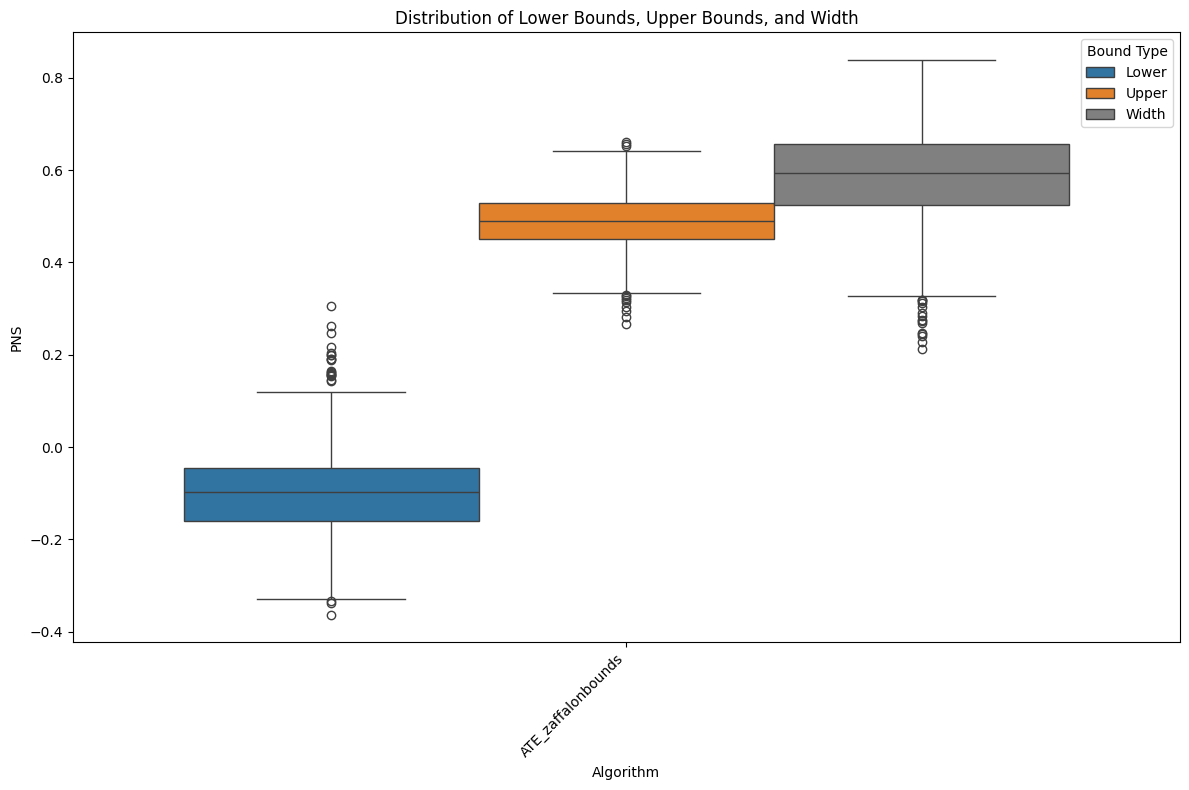}
    \caption{Distribution of lower bounds, upper bounds, and bound widths for \texttt{zaffalonbounds} in the ATE setting across \(M = 2000\) resampled runs.}
    \label{fig:boxplot-ate}
\end{figure}

\begin{figure}[H]
    \centering
    \includegraphics[width=0.8\textwidth]{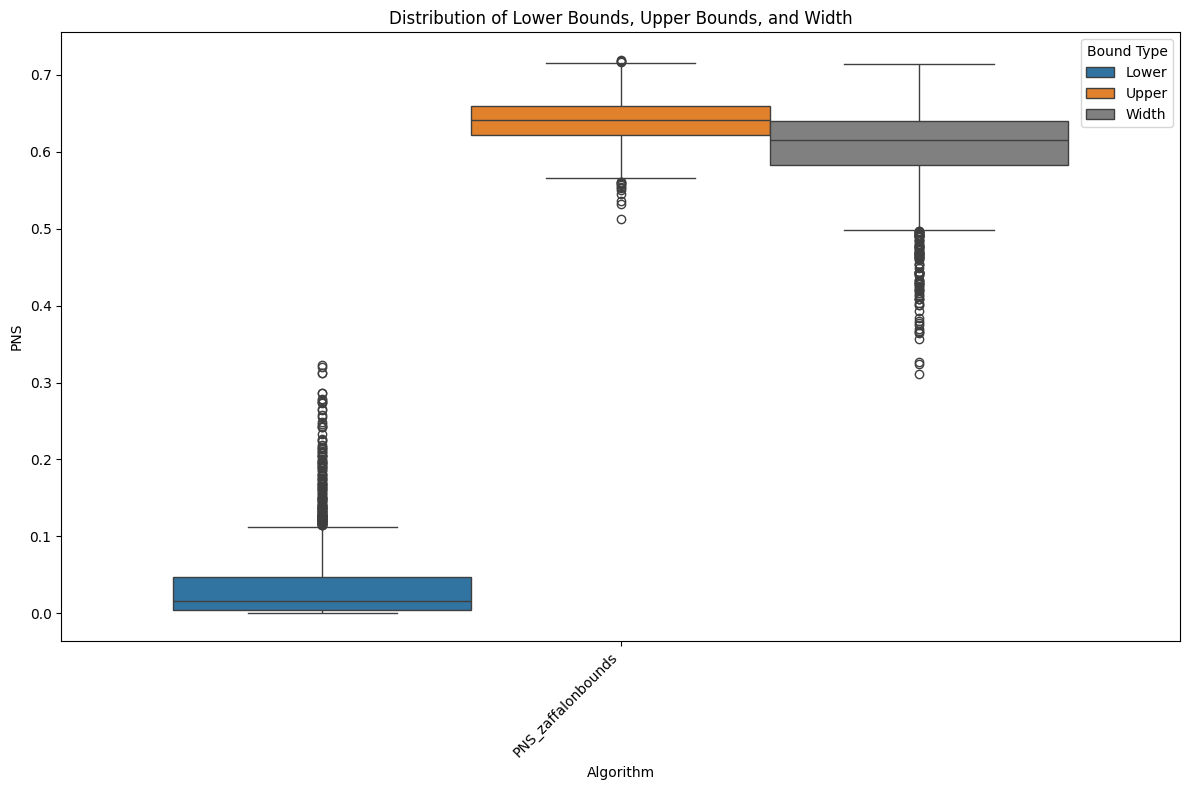}
    \caption{Distribution of lower bounds, upper bounds, and bound widths for \texttt{zaffalonbounds} in the PNS setting across \(M = 2000\) resampled runs.}
    \label{fig:boxplot-pns}
\end{figure}

To further visualize this effect, Figure~\ref{fig:boxplot-ate} and Figure~\ref{fig:boxplot-pns} show boxplots for the \texttt{zaffalonbounds} algorithm, comparing the distributions of the lower bound, upper bound, and bound width across the 2000 resampled runs. Separate plots are shown for the ATE and PNS settings.

\noindent
Even for the most sensitive algorithm, variability remains within a practically acceptable range. This analysis shows that the results of the bounding algorithms used throughout this thesis are relatively robust to random fluctuations in data, and that the reported results are not purely artifacts of simulation noise.

%% file: chapters/ch-simulation_and_experiment/binaryIV/index.tex
\section{Binary Instrumental Variable}
\label{sec:scenario-binIV}

This scenario extends the binary confounding setup by introducing an instrumental variable (IV). A treatment variable \(X\) has a causal effect on an outcome \(Y\), and both are influenced by an unobserved confounder \(U\). To mitigate this confounding, we introduce an exogenous instrument \(Z\), which affects \(X\) but has no direct effect on \(Y\) except through \(X\). All variables are binary: \(X, Y, Z, U \in \{0,1\}\). This setup is depicted by the causal diagram below:

\begin{center}
\begin{tikzpicture}[->, node distance=2cm, thick]
  \node[draw, circle] (X) {\( X \)};
    \node[draw, circle, left of=X] (Z) {\( Z \)};
  \node[draw, circle, right of=X] (Y) {\( Y \)};
  \node[draw, circle, dashed, above of=X] (U) {\( U \)};

  \draw (Z) -- (X);
  \draw (X) -- (Y);
  \draw (U) -- (X);
  \draw (U) -- (Y);
\end{tikzpicture}
\end{center}

As before, the objective is to estimate or bound the following causal quantities:
\[
\text{ATE} = P(Y_{X=1} = 1) - P(Y_{X=0} = 1), \qquad
\text{PNS} = P(Y_{X=1} = 1,\, Y_{X=0} = 0).
\]

Each simulation \(\mathcal{S}_j\) generates \(n = 500\) unit-level observations \((X_i, Y_i, U_i, Z_i)\) using fixed, simulation-specific parameters and link functions.

\input{chapters/ch-simulation_and_experiment/binaryIV/dataGen}

\input{chapters/ch-simulation_and_experiment/binaryIV/results}

%% file: chapters/ch-simulation_and_experiment/binaryIV/dataGen.tex
\subsection{Data-Generating Process}
\label{binIV_dataGen}

The \texttt{BinaryIV} scenario builds upon the confounding setup defined in Section~\ref{binConf_dataGen}, introducing a binary instrument \(Z_i \sim \text{Bern}(p_Z)\), with \(p_Z \sim \text{Uni}(0,1)\). The instrument is therefore exogenous, i.e., independent of the unobserved confounder \(U_i\).

The treatment mechanism is modified to include an additional term:
\[
X_i^* = \alpha_X
        + \beta_{Z \to X} Z_i
        + \beta_{U \to X} U_i
        + \varepsilon_{X_i}, \qquad
X_i \sim \text{Bern}\bigl(f_X(X_i^*)\bigr)
\]
where \(\beta_{Z \to X} \sim \tfrac{1}{2} \mathcal{N}(1, 0.5^2) + \tfrac{1}{2} \mathcal{N}(-1, 0.5^2)\).

All other components — including heteroskedastic noise, squashing functions \(f_X, f_Y \in \mathcal{F}\), and counterfactual definitions — remain unchanged from Section~\ref{binConf_dataGen}.

\begin{figure}[H]
\centering
\begin{tikzpicture}[>=stealth, node distance=2.8cm, thick]

\tikzstyle{latent} = [draw, circle, dashed]
\tikzstyle{obs} = [draw, circle]
\tikzstyle{vartext} = [align=center, font=\small]

\node[latent] (U) {\(U_i\)};
\node[obs, below left=2.2cm and 1.5cm of U] (X) {\(X_i\)};
\node[obs, below right=2.2cm and 1.5cm of U] (Y) {\(Y_i\)};
\node[obs, left=2.5cm of X] (Z) {\(Z_i\)};

\node[vartext, above=0.1cm of U] {\(U_i \sim \text{Bern}(p_U)\)};
\node[vartext, above=0.1cm of Z, xshift=-0.5cm] {\(Z_i \sim \text{Bern}(p_Z)\)};
\node[vartext, below=0.2cm of X, xshift=2.0cm] 
      {\(X_i \sim \text{Bern}\bigl(f_X(\alpha_X + \beta_{Z \to X} Z_i + \beta_{U \to X} U_i + \varepsilon_{X_i})\bigr)\)};
\node[vartext, above=0.0cm of Y, xshift=2.0cm, yshift=-0.5] 
      {\(Y_i \sim \text{Bern}\bigl(f_Y(Y_i^*)\bigr)\)};

\draw[->] (U) -- (X) node[midway, left] {\(\beta_{U \to X}\)};
\draw[->] (U) -- (Y) node[midway, right] {\(\beta_{U \to Y}\)};
\draw[->] (X) -- (Y) node[midway, below] {\(\beta_{X \to Y}\)};
\draw[->] (Z) -- (X) node[midway, above] {\(\beta_{Z \to X}\)};

\end{tikzpicture}
\caption{Binary SCM with confounding and instrument: causal graph and data-generating process.}
\label{fig:binaryIV-dgp-annotated}
\end{figure}
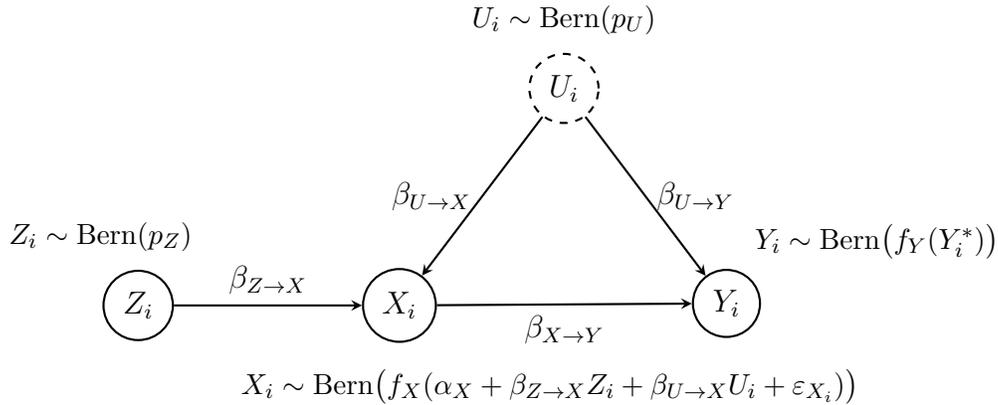

%% file: chapters/ch-simulation_and_experiment/binaryIV/results.tex
\subsection{Results}

We now present the simulation results for the \texttt{BinaryIV} scenario—first for the ATE, then for the PNS.

Note that in this setting, not all algorithms make use of the instrumental variable \(Z\). Tables~\ref{tab:binaryiv-ate-instrument} and~\ref{tab:binaryiv-pns-instrument} provide an overview of whether each algorithm incorporates the instrument.

\begin{table}[H]
\centering
\caption{Considered algorithms for the ATE in the \texttt{BinaryIV} scenario. Not all of them utilize the instrument \(Z\).}
\label{tab:binaryiv-ate-instrument}
\renewcommand{\arraystretch}{1.1}
\begin{tabularx}{0.6\textwidth}{Xc}
\toprule
\textbf{Algorithm} & \textbf{Uses Instrument?} \\
\midrule
\texttt{ATE\_manski}         & No  \\
\texttt{ATE\_2SLS}           & Yes \\
\texttt{ATE\_causaloptim}    & Yes \\
\texttt{ATE\_autobound}      & Yes \\
\texttt{ATE\_zaffalonbounds} & Yes \\
\texttt{ATE\_entropybounds}  & No  \\
\bottomrule
\end{tabularx}
\end{table}

\begin{table}[H]
\centering
\caption{Considered algorithms for the PNS in the \texttt{BinaryIV} scenario. Not all of them utilize the instrument \(Z\).}
\label{tab:binaryiv-pns-instrument}
\renewcommand{\arraystretch}{1.1}
\begin{tabularx}{0.6\textwidth}{Xc}
\toprule
\textbf{Algorithm} & \textbf{Uses Instrument?} \\
\midrule
\texttt{PNS\_tianpearl}         & No  \\
\texttt{PNS\_causaloptim}       & Yes \\
\texttt{PNS\_autobound}         & Yes \\
\texttt{PNS\_zaffalonbounds}    & Yes \\
\texttt{PNS\_entropybounds}     & No  \\
\bottomrule
\end{tabularx}
\end{table}

All introduced algorithms are applicable to this scenario, except for \texttt{zhangbareinboim}, which requires a continuous outcome, and \texttt{OLS}.

\paragraph*{ATE}~

Table~\ref{tab:binaryiv-ate-results} reports the evaluation metrics for all applicable algorithms (see Section~\ref{evaluation_metrics} for definitions). We highlight the \textbf{Invalid Rate} and \textbf{Net Width} columns for emphasis.

\begin{table}[H]
\centering
\caption{Results for ATE in the \texttt{BinaryIV} scenario.}
\label{tab:binaryiv-ate-results}
\scriptsize
\renewcommand{\arraystretch}{1.1}
\begin{tabularx}{\textwidth}{Xrrrrr@{\hskip 6pt}}
\toprule
\textbf{Algorithm} & \textbf{Fail Rate} & \textbf{Invalid Rate} & \textbf{Net Width} & \textbf{Bound Width} & \textbf{Invalid $\Delta$} \\
\midrule
\texttt{zaffalonbounds}            & 0.00 & \textbf{5.70}  & \textbf{28.47} & 32.55 & 1.08 \\
\texttt{entropybounds-0.10}        & 0.00 & \textbf{23.40} & \textbf{33.61} & 49.14 & 4.69 \\
\texttt{causaloptim}               & 0.90 & \textbf{5.65}  & \textbf{37.29} & 41.37 & 2.59 \\
\texttt{autobound}                 & 4.05 & \textbf{0.99}  & \textbf{40.08} & 43.07 & 2.51 \\
\texttt{entropybounds-0.20}        & 0.00 & \textbf{11.10} & \textbf{41.26} & 47.78 & 4.70 \\
\texttt{entropybounds-randomTheta} & 0.00 & \textbf{12.50} & \textbf{46.02} & 52.77 & 5.88 \\
\texttt{entropybounds-trueTheta}   & 0.00 & \textbf{9.35}  & \textbf{48.62} & 53.42 & 5.44 \\
\texttt{entropybounds-0.80}        & 0.00 & \textbf{8.50}  & \textbf{49.81} & 54.07 & 5.32 \\
\texttt{manski}                    & 0.00 & \textbf{0.05}  & \textbf{50.00} & 50.02 & 0.92  \\
\texttt{2SLS-0.95}                 & 0.15 & \textbf{4.21}  & \textbf{56.49} & 58.39 & 7.71 \\
\texttt{2SLS-0.98}                 & 0.15 & \textbf{1.10}  & \textbf{61.09} & 61.58 & 9.27 \\
\texttt{2SLS-0.99}                 & 0.15 & \textbf{0.80}  & \textbf{64.12} & 64.46 & 8.41 \\
\bottomrule
\end{tabularx}
\end{table}

\begin{figure}[H]
    \centering
    \includegraphics[width=0.8\textwidth]{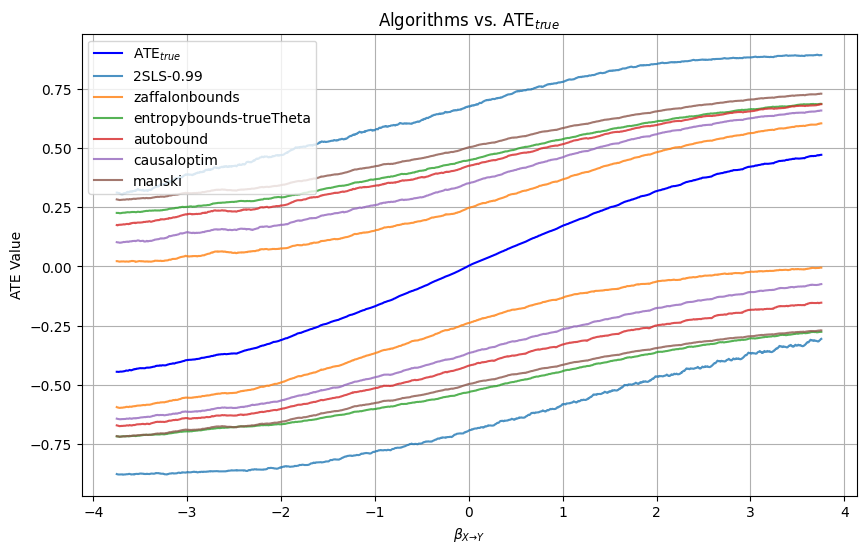}
    \caption{Upper and lower bounds for the ATE of selected algorithms in the \texttt{BinaryIV} scenario, alongside the true ATE. For visualization, lines are smoothed by averaging across 500 samples.}
    \label{fig:binaryIV_algs_vs_ATE}
\end{figure}

Figure~\ref{fig:binaryIV_algs_vs_ATE} shows the upper and lower bounds of selected algorithms alongside the true ATE, smoothed over 500 samples.

\paragraph*{PNS}~

Table~\ref{tab:binaryiv-pns-results} summarizes the performance of PNS algorithms. As before, we emphasize the \textbf{Invalid Rate} and \textbf{Net Width} columns.

\begin{table}[H]
\centering
\caption{Results for PNS in the \texttt{BinaryIV} scenario.}
\label{tab:binaryiv-pns-results}
\scriptsize
\renewcommand{\arraystretch}{1.1}
\begin{tabularx}{\textwidth}{Xrrrrr@{\hskip 6pt}}
\toprule
\textbf{Algorithm} & \textbf{Fail Rate} & \textbf{Invalid Rate} & \textbf{Net Width} & \textbf{Bound Width} & \textbf{Invalid $\Delta$} \\
\midrule
\texttt{zaffalonbounds}            & 0.00 & \textbf{0.60}  & \textbf{41.96} & 42.31 & 0.93 \\
\texttt{causaloptim}               & 5.15 & \textbf{17.13} & \textbf{43.23} & 55.38 & 7.40 \\
\texttt{autobound}                 & 4.05 & \textbf{1.72}  & \textbf{43.69} & 46.90 & 1.96 \\
\texttt{entropybounds-0.10}        & 0.00 & \textbf{0.15}  & \textbf{48.90} & 48.98 & 0.00 \\
\texttt{tianpearl}                 & 0.00 & \textbf{0.00}  & \textbf{50.50} & 50.50 & N/A  \\
\texttt{entropybounds-0.20}        & 0.00 & \textbf{0.05}  & \textbf{56.78} & 56.80 & 0.00 \\
\texttt{entropybounds-randomTheta} & 0.00 & \textbf{0.05}  & \textbf{62.86} & 62.88 & 0.00 \\
\texttt{entropybounds-trueTheta}   & 0.00 & \textbf{0.05}  & \textbf{68.51} & 68.53 & 0.00 \\
\texttt{entropybounds-0.80}        & 0.00 & \textbf{0.00}  & \textbf{70.57} & 70.57 & N/A  \\
\bottomrule
\end{tabularx}
\end{table}

\begin{figure}[H]
    \centering
    \includegraphics[width=0.8\textwidth]{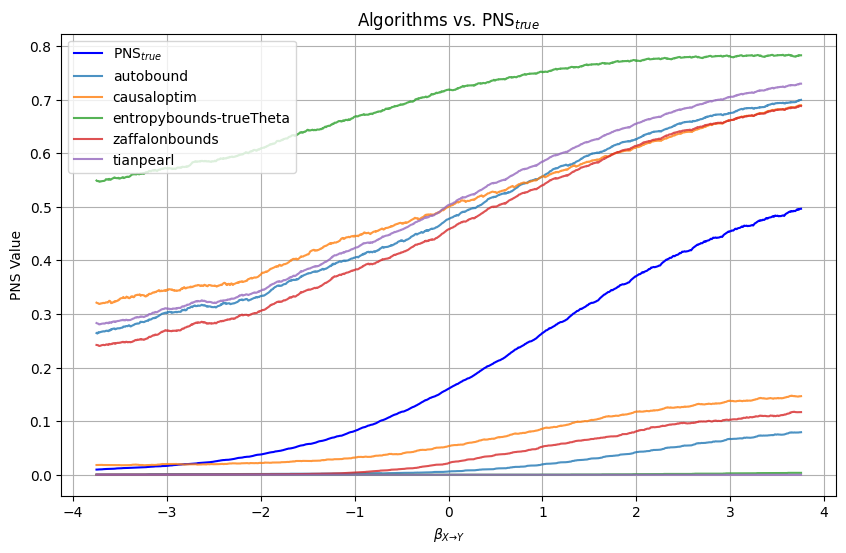}
    \caption{Upper and lower bounds of selected PNS algorithms in the \texttt{BinaryIV} scenario, together with the true PNS. Lines are smoothed by averaging across 500 samples.}
    \label{fig:binaryIV_algs_vs_PNS}
\end{figure}

%% file: chapters/ch-simulation_and_experiment/contConf/index.tex
\section{Continuous Confounding}
\label{sec:scenario-contConf}

This scenario introduces a structural causal model involving a binary treatment \(X \in \{0,1\}\) and a continuous, bounded outcome \(Y \in [0,1]\), both influenced by an unobserved continuous confounder \(U \in \mathbb{R}\). The objective is to estimate the Average Treatment Effect (ATE), defined as:
\[
\text{ATE} = \mathbb{E}[Y(1) - Y(0)]
\]

Each simulation \(\mathcal{S}_j\) generates \(n = 500\) unit-level observations \((X_i, Y_i, U_i)\) using the data-generating process described below.

\input{chapters/ch-simulation_and_experiment/contConf/dataGen}

\input{chapters/ch-simulation_and_experiment/contConf/results}

%% file: chapters/ch-simulation_and_experiment/contConf/dataGen.tex
\subsection{Data-Generating Process}
\label{contConf_dataGen}

This scenario simulates a continuous, bounded outcome \(Y \in [0,1]\) influenced by a binary treatment \(X\) and a continuous unobserved confounder \(U \in \mathbb{R}\). Each simulation \(\mathcal{S}_j\) draws fixed simulation-specific parameters and generates \(n = 500\) unit-level observations.

\paragraph{Latent confounder.}
Each unit draws a latent confounder:
\[
U_i \sim \mathcal{N}(0,\sigma_U^2), \qquad \sigma_U \sim |\mathcal{N}(0,1)|.
\]

\paragraph{Simulation-specific parameters.}
Each simulation samples:
\[
\alpha_X,\ \alpha_Y \sim \mathcal{N}(0,1)
\]
\[
\beta_{U \to X},\ \beta_{U \to Y} \sim \tfrac{1}{2}\, \mathcal{N}(1,0.5^2) + \tfrac{1}{2}\, \mathcal{N}(-1,0.5^2)
\]
with \(\beta_{X \to Y}\) deterministically swept across \([-5, 5]\) over simulations.

\paragraph{Observation-level heteroskedastic noise.}
We first draw global noise scales:
\[
\sigma_X, \sigma_Y \sim |\mathcal{N}(0,1)|
\]

Then for each unit $i$, we sample:
\[
\sigma_{X_i} \sim |\mathcal{N}(0,\sigma_X)|, \qquad
\sigma_{Y_i} \sim |\mathcal{N}(0,\sigma_Y)|
\]
\[
\varepsilon_{X_i} \sim \mathcal{N}(0,\sigma_{X_i}^2), \qquad
\varepsilon_{Y_i} \sim \mathcal{N}(0,\sigma_{Y_i}^2)
\]

\paragraph{Squashing functions and feature transforms.}
For each simulation, we independently sample:
\begin{itemize}
    \item Squashing functions \(f_X, f_Y \in \mathcal{F}\), with:
    \[
    \mathcal{F} = \left\{
        \frac{1}{1 + e^{-x}},\ 
        \frac{1}{2}(1 + \tanh(x)),\ 
        \frac{\log(1 + e^x)}{1 + \log(1 + e^x)},\ 
        \Phi(x)
    \right\}
    \]
    where \(\Phi(x)\) denotes the CDF of the standard normal distribution.
    
    \item Feature transforms \(g_{U \to X},\ g_{U \to Y} \in \mathcal{G}\), with:
    \begin{align*}
    \mathcal{G} = \{\, 
    & x,\ \sin(x),\ \cos(x),\ \tanh(x),\ \log(1 + |x|),\ e^{-x^2},\ \frac{1}{1 + e^{-x}}, \\
    & e^{\mathrm{clip}(x, -5, 5)},\ \tanh(x),\ \frac{1}{1 + e^{-(x - \mu)}},\ \sin(\pi x), \\
    & \mathrm{clip}\left(\frac{x}{5}, -1, 1\right),\ \frac{x}{1 + |x|} \,\}
    \end{align*}
    with \(\mu = \frac{1}{n} \sum_{i=1}^{n} x_i\) denoting the sample mean.
\end{itemize}
All functions are drawn uniformly from their respective sets.

\paragraph{Treatment mechanism.}
\[
X_i^* = \alpha_X + \beta_{U \to X} \cdot g_{U \to X}(U_i) + \varepsilon_{X_i}, \qquad
X_i \sim \text{Bernoulli}(f_X(X_i^*))
\]

\paragraph{Outcome mechanism.}
\[
Y_i^* = \alpha_Y + \beta_{X \to Y} \cdot X_i + \beta_{U \to Y} \cdot g_{U \to Y}(U_i) + \varepsilon_{Y_i}, \qquad
Y_i = f_Y(Y_i^*) \in [0,1]
\]

\paragraph{Counterfactual outcomes.}
The potential outcomes for each unit are:
\[
\begin{aligned}
Y_{X=1,i} &= f_Y\bigl(\alpha_Y + \beta_{X \to Y} + \beta_{U \to Y} \cdot g_{U \to Y}(U_i) + \varepsilon_{Y_i}\bigr) \\
Y_{X=0,i} &= f_Y\bigl(\alpha_Y + \beta_{U \to Y} \cdot g_{U \to Y}(U_i) + \varepsilon_{Y_i}\bigr)
\end{aligned}
\]

\paragraph{Causal estimand.}
\[
\text{ATE} = \mathbb{E}[Y_{X=1} - Y_{X=0}]
\]

\begin{figure}[H]
\centering
\begin{tikzpicture}[>=stealth, node distance=2.8cm, thick]

\tikzstyle{latent} = [draw, circle, dashed]
\tikzstyle{obs} = [draw, circle]
\tikzstyle{vartext} = [align=center, font=\small]

\node[latent] (U) {\(U_i\)};
\node[obs, below left=2.2cm and 1.8cm of U] (X) {\(X_i\)};
\node[obs, below right=2.2cm and 1.8cm of U] (Y) {\(Y_i\)};

\node[vartext, above=0.1cm of U] 
  {\(U_i \sim \mathcal{N}(0,\sigma_U^2),\quad \sigma_U \sim |\mathcal{N}(0,1)|\)};
\node[vartext, below=0.2cm of X, xshift=-2.3cm] 
  {\(X_i \sim \mathrm{Bern}\bigl(f_X(\alpha_X + \beta_{U \to X} \cdot g_{U \to X}(U_i) + \varepsilon_{X_i})\bigr)\)};
\node[vartext, below=0.2cm of Y, xshift=2.3cm] 
  {\(Y_i = f_Y(\alpha_Y + \beta_{X \to Y} X_i + \beta_{U \to Y} \cdot g_{U \to Y}(U_i) + \varepsilon_{Y_i}) \in [0,1]\)};

\draw[->] (U) -- (X) node[midway, left] {\(\beta_{U \to X},\ g_{U \to X}\)};
\draw[->] (U) -- (Y) node[midway, right] {\(\beta_{U \to Y},\ g_{U \to Y}\)};
\draw[->] (X) -- (Y) node[midway, below] {\(\beta_{X \to Y}\)};

\end{tikzpicture}
\caption{Causal DAG with structural coefficients and feature transforms for continuous confounding.}
\label{fig:dag-cont-confounding-annotated}
\end{figure}

\paragraph{Binarization.}
As noted in Section~\ref{sec:naming-convention}, we additionally create a separate version of the dataset in which a binarization step is applied to the observed variables. Specifically, we use a threshold of 0.5: values less than or equal to this threshold are mapped to 0, and values above are mapped to 1. If we were to also binarize the unobserved confounder \(U\) using the same threshold, it would yield an average entropy of approximately 0.61 across simulations.

%% file: chapters/ch-simulation_and_experiment/contConf/results.tex
\subsection{Results}

We now present the simulation results for the \texttt{ContConf} scenario.

In this setting, most algorithms are not directly applicable due to the continuous outcome \(Y \in [0,1]\). As a workaround, we apply a binarization step (see Section~\ref{contConf_dataGen}) to create a version of the dataset with binary outcomes. The only algorithm that operates on the continuous outcome directly is \texttt{OLS}.

Table~\ref{tab:contconf-ate-algorithms} provides an overview of the algorithms used for estimating the ATE in this scenario and whether they were applied to the binarized data.

\begin{table}[H]
\centering
\caption{Considered algorithms for estimating the ATE in the \texttt{ContConf} scenario. All except \texttt{OLS} operate on binarized outcomes.}
\label{tab:contconf-ate-algorithms}
\renewcommand{\arraystretch}{1.1}
\begin{tabularx}{0.5\textwidth}{Xc}
\toprule
\textbf{Algorithm} & \textbf{Binarized?} \\
\midrule
\texttt{ATE\_manski}             & Yes \\
\texttt{ATE\_OLS}                & No  \\
\texttt{ATE\_causaloptim}        & Yes \\
\texttt{ATE\_autobound}          & Yes \\
\texttt{ATE\_zaffalonbounds}     & Yes \\
\texttt{ATE\_entropybounds}      & Yes \\
\bottomrule
\end{tabularx}
\end{table}

Additionally, we omit the \texttt{2SLS} algorithm, as it requires an instrumental variable, and the \texttt{entropybounds-trueTheta} algorithm, since the true empirical Shannon entropy of the continuous confounder is not straightforward to compute in this setting.

Table~\ref{tab:contconf-ate-results} shows the evaluation metrics for each applicable algorithm. The \textbf{Invalid Rate} and \textbf{Net Width} are again highlighted to compare reliability and informativeness.

\begin{table}[H]
\centering
\caption{Results for ATE in the \texttt{ContConf} scenario.}
\label{tab:contconf-ate-results}
\scriptsize
\renewcommand{\arraystretch}{1.1}
\begin{tabularx}{\textwidth}{Xrrrrr@{\hskip 6pt}}
\toprule
\textbf{Algorithm} & \textbf{Fail Rate} & \textbf{Invalid Rate} & \textbf{Net Width} & \textbf{Bound Width} & \textbf{Invalid $\Delta$} \\
\midrule
\texttt{OLS-0.95}                            & 0.00 & \textbf{72.85} & \textbf{4.70}  & 74.13 & 6.03 \\
\texttt{OLS-0.98}                            & 0.00 & \textbf{69.70} & \textbf{5.52}  & 71.37 & 6.06 \\
\texttt{OLS-0.99}                            & 0.00 & \textbf{67.65} & \textbf{6.00}  & 69.59 & 6.09 \\
\texttt{entropybounds-0.10--binned}         & 0.20 & \textbf{16.28} & \textbf{29.20} & 40.85 & 6.73 \\
\texttt{entropybounds-0.20--binned}         & 0.20 & \textbf{7.97}  & \textbf{37.73} & 42.80 & 6.55 \\
\texttt{zaffalonbounds--binned}             & 0.00 & \textbf{16.90} & \textbf{38.16} & 48.61 & 7.89 \\
\texttt{entropybounds-randomTheta--binned}  & 0.20 & \textbf{9.02}  & \textbf{43.67} & 48.85 & 6.34 \\
\texttt{entropybounds-0.80--binned}         & 0.20 & \textbf{4.61}  & \textbf{49.49} & 51.91 & 5.57 \\
\texttt{autobound--binned}                  & 0.00 & \textbf{9.25}  & \textbf{50.00} & 54.62 & 6.80 \\
\texttt{causaloptim--binned}                & 0.00 & \textbf{9.25}  & \textbf{50.00} & 54.62 & 6.80 \\
\texttt{manski--binned}                     & 0.00 & \textbf{9.25}  & \textbf{50.00} & 54.62 & 6.80 \\
\bottomrule
\end{tabularx}
\end{table}

\begin{figure}[H]
    \centering
    \includegraphics[width=0.8\textwidth]{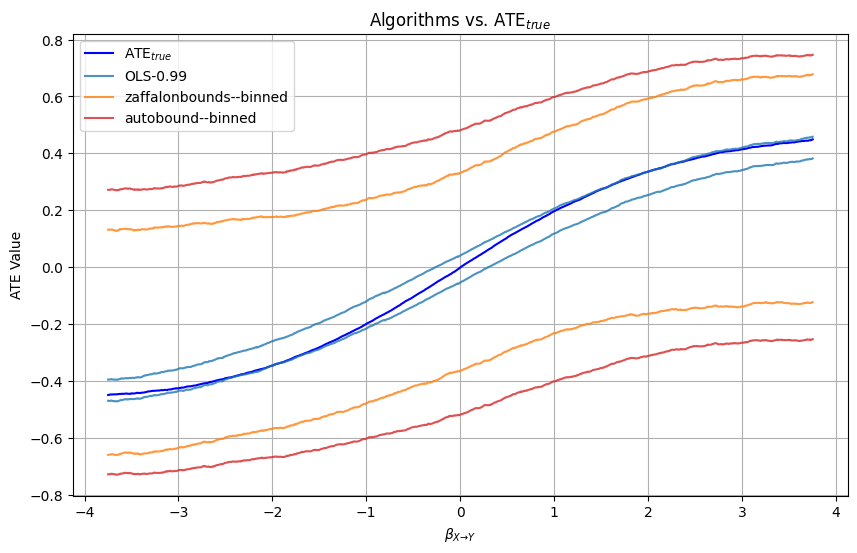}
    \caption{Upper and lower bounds of selected ATE algorithms in the \texttt{ContConf} scenario together with the true ATE. Lines are smoothed by averaging across 500 samples.}
    \label{fig:contConf_algs_vs_ATE}
\end{figure}

Figure~\ref{fig:contConf_algs_vs_ATE} displays the upper and lower bounds of selected algorithms in this scenario, smoothed by averaging over 500 samples.

%% file: chapters/ch-simulation_and_experiment/contIV/index.tex
\section{Continuous Instrumental Variable}
\label{sec:scenario-contIV}

This scenario extends the continuous confounding setup by introducing a binary instrumental variable \(Z \in \{0,1\}\). The treatment \(X \in \{0,1\}\) has a causal effect on a continuous, bounded outcome \(Y \in [0,1]\), while both are also influenced by a continuous unobserved confounder \(U \in \mathbb{R}\).

The instrument \(Z\) affects the treatment but is assumed to be independent of the confounder \(U\).

Our goal is to bound the Average Treatment Effect (ATE), defined as:
\[
\text{ATE} = \mathbb{E}[Y_{X=1} - Y_{X=0}]
\]

Each simulation \(\mathcal{S}_j\) generates \(n = 500\) unit-level observations \((Z_i, X_i, Y_i, U_i)\) using the data-generating process described below.

\input{chapters/ch-simulation_and_experiment/contIV/dataGen}

\input{chapters/ch-simulation_and_experiment/contIV/results}

%% file: chapters/ch-simulation_and_experiment/contIV/dataGen.tex
\subsection{Data-Generating Process}
\label{contIV_dataGen}

This simulation extends the continuous confounding scenario defined in Section~\ref{contConf_dataGen} by introducing a binary instrument \(Z_i \sim \mathrm{Bern}(p_Z)\), with \(p_Z \sim \mathrm{Uni}(0,1)\). The instrument \(Z_i\) is assumed to be independent of the confounder \(U_i\) and has no direct effect on \(Y_i\) other than through \(X_i\).

\paragraph{Modified treatment mechanism.}
The structural equation for treatment is updated to include an instrument term:
\[
X_i^* = \alpha_X
      + \beta_{Z \to X} \cdot   Z_i
      + \beta_{U\to X} \cdot g_{U\to X}(U_i)
      + \varepsilon_{X_i}, \qquad
X_i \sim \text{Bern}(f_X(X_i^*))
\]
The coefficient \(\beta_{Z \to X}\) is sampled from the same bimodal mixture as the other structural coefficients:
\[
\beta_{Z \to X} \sim \tfrac{1}{2} \mathcal{N}(1, 0.5^2) + \tfrac{1}{2} \mathcal{N}(-1, 0.5^2)
\]

\paragraph{Other components.}
All other components — including the outcome mechanism, noise generation, squashing functions \(f_X, f_Y \in \mathcal{F}\), feature transforms \(g_{U \to X}, g_{U \to Y} \in \mathcal{G}\), and counterfactual logic — remain unchanged from Section~\ref{contConf_dataGen}.

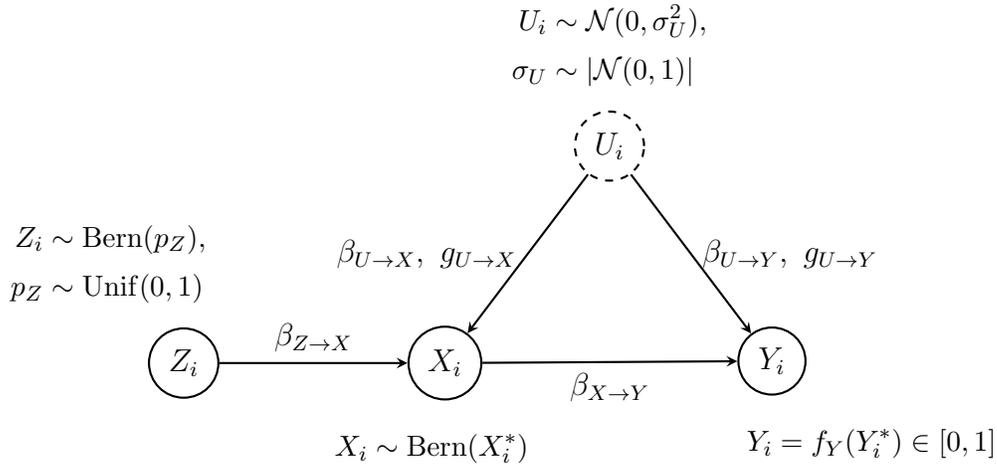
\begin{figure}[H]
\centering
\begin{tikzpicture}[>=stealth, node distance=2.8cm, thick]

\tikzstyle{latent} = [draw, circle, dashed]
\tikzstyle{obs} = [draw, circle]
\tikzstyle{vartext} = [align=center, font=\small]

\node[latent] (U) {\(U_i\)};
\node[obs, below left=2.2cm and 1.5cm of U] (X) {\(X_i\)};
\node[obs, below right=2.2cm and 1.5cm of U] (Y) {\(Y_i\)};
\node[obs, left=2.5cm of X] (Z) {\(Z_i\)};

\node[vartext, above=0.1cm of U] 
  {\( 
    \begin{aligned}
      U_i &\sim \mathcal{N}(0,\sigma_U^2), \\
      \sigma_U &\sim |\mathcal{N}(0,1)|
    \end{aligned}
  \)};

\node[vartext, above=0.1cm of Z, xshift=-1.0cm] 
  {\(
    \begin{aligned}
      Z_i &\sim \mathrm{Bern}(p_Z), \\
      p_Z &\sim \mathrm{Unif}(0,1)
    \end{aligned}
  \)};

\node[vartext, below=0.2cm of X, xshift=-0.0cm] 
  {\(
    \begin{aligned}
      X_i &\sim \mathrm{Bern}(X_i^*)
    \end{aligned}
  \)};

\node[vartext, below=0.2cm of Y, xshift=1.5cm] 
  {\(
    \begin{aligned}
      Y_i &= f_Y(Y_i^*) \in [0,1]
    \end{aligned}
  \)};

\draw[->] (Z) -- (X) node[midway, above] {\(\beta_{Z \to X}\)};
\draw[->] (U) -- (X) node[midway, left] {\(\beta_{U \to X},\ g_{U \to X}\)};
\draw[->] (U) -- (Y) node[midway, right] {\(\beta_{U \to Y},\ g_{U \to Y}\)};
\draw[->] (X) -- (Y) node[midway, below] {\(\beta_{X \to Y}\)};

\end{tikzpicture}
\caption{
Causal DAG with structural coefficients, instrument, and feature transforms for the continuous IV scenario. This extends the continuous confounding setup by adding an exogenous instrument \(Z\).
}
\label{fig:dag-cont-iv-annotated}
\end{figure}

%% file: chapters/ch-simulation_and_experiment/contIV/results.tex
\subsection{Results}

We now present the simulation results for the \texttt{ContIV} scenario.

As in the \texttt{ContConf} scenario (see Section~\ref{sec:scenario-contConf}), most algorithms are not directly applicable to the continuous outcome. We therefore apply a binarization step to the data, as described in Section~\ref{contConf_dataGen}. The only exceptions are \texttt{2SLS} and \texttt{zhangbareinboim}, which are capable of handling continuous outcomes directly.

We omit the \texttt{OLS} algorithm, as it does not match the instrumental variable DAG assumed in this setting, and the \texttt{entropybounds-trueTheta} algorithm, since exact entropy computation for the continuous confounder is not performed here.

Unlike \texttt{ContConf}, this scenario is particularly interesting because it allows us to evaluate \texttt{zhangbareinboim}—the only genuine causal bounding algorithm in our study that supports continuous outcomes.

Table~\ref{tab:contiv-ate-algorithms} summarizes which ATE algorithms are used in this scenario and whether they are applied to binarized data.

\begin{table}[H]
\centering
\caption{Considered algorithms for the ATE in the \texttt{ContIV} scenario. All except \texttt{2SLS} and \texttt{zhangbareinboim} operate on binarized outcomes.}
\label{tab:contiv-ate-algorithms}
\renewcommand{\arraystretch}{1.1}
\begin{tabularx}{0.5\textwidth}{Xc}
\toprule
\textbf{Algorithm} & \textbf{Binarized?} \\
\midrule
\texttt{ATE\_manski}             & Yes \\
\texttt{ATE\_2SLS}                & No \\
\texttt{ATE\_causaloptim}        & Yes \\
\texttt{ATE\_autobound}          & Yes \\
\texttt{ATE\_zaffalonbounds}     & Yes \\
\texttt{ATE\_entropybounds}      & Yes \\
\texttt{ATE\_zhangbareinboim}    & No \\
\bottomrule
\end{tabularx}
\end{table}

Table~\ref{tab:contiv-ate-results} reports the evaluation metrics for all applicable algorithms. As in earlier scenarios, we emphasize the \textbf{Invalid Rate} and \textbf{Net Width} columns.

\begin{table}[H]
\centering
\caption{Results for ATE in the \texttt{ContIV} scenario.}
\label{tab:contiv-ate-results}
\scriptsize
\renewcommand{\arraystretch}{1.1}
\begin{tabularx}{\textwidth}{Xrrrrr@{\hskip 6pt}}
\toprule
\textbf{Algorithm} & \textbf{Fail Rate} & \textbf{Invalid Rate} & \textbf{Net Width} & \textbf{Bound Width} & \textbf{Invalid $\Delta$} \\
\midrule
\texttt{entropybounds-0.10--binned}        & 0.00  & \textbf{16.50} & \textbf{28.98} & 40.69 & 6.20 \\
\texttt{zaffalonbounds--binned}            & 0.00  & \textbf{23.35} & \textbf{33.36} & 48.92 & 7.44 \\
\texttt{2SLS-0.95}                          & 0.35  & \textbf{30.41} & \textbf{35.90} & 55.55 & 6.53 \\
\texttt{causaloptim--binned}               & 5.20  & \textbf{21.04} & \textbf{36.40} & 52.40 & 8.86 \\
\texttt{entropybounds-0.20--binned}        & 0.00  & \textbf{7.50}  & \textbf{37.50} & 42.19 & 5.89 \\
\texttt{2SLS-0.98}                          & 0.35  & \textbf{26.19} & \textbf{38.71} & 54.92 & 6.69 \\
\texttt{autobound--binned}                 & 11.40 & \textbf{11.17} & \textbf{39.77} & 52.60 & 7.27 \\
\texttt{zhangbareinboim}                   & 1.20  & \textbf{7.69}  & \textbf{40.11} & 45.38 & 5.20 \\
\texttt{2SLS-0.99}                          & 0.35  & \textbf{24.79} & \textbf{40.77} & 55.61 & 6.54 \\
\texttt{entropybounds-randomTheta--binned} & 0.00  & \textbf{7.00}  & \textbf{43.89} & 47.82 & 6.43 \\
\texttt{entropybounds-0.80--binned}        & 0.00  & \textbf{3.35}  & \textbf{49.46} & 51.16 & 5.51 \\
\texttt{manski--binned}                    & 0.00  & \textbf{7.55}  & \textbf{50.00} & 53.77 & 6.62  \\
\bottomrule
\end{tabularx}
\end{table}

\begin{figure}[H]
    \centering
    \includegraphics[width=0.8\textwidth]{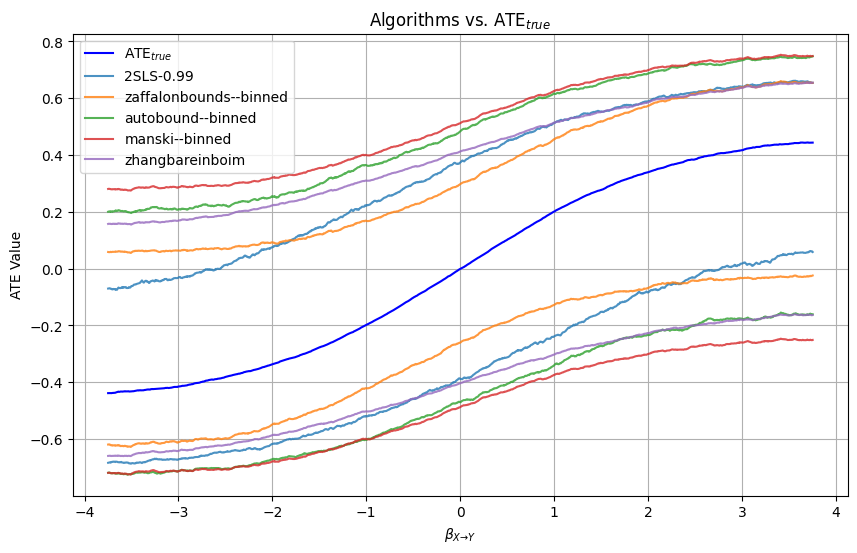}
    \caption{Upper and lower bounds of selected ATE algorithms in the \texttt{ContIV} scenario together with the true ATE. Lines are smoothed by averaging across 500 samples.}
    \label{fig:contIV_algs_vs_ATE}
\end{figure}

Figure~\ref{fig:contIV_algs_vs_ATE} shows the upper and lower bounds of selected algorithms together with the true ATE, smoothed over 500 samples.

%% file: chapters/ch-simulation_and_experiment/entropyConf/index.tex
\section{Binary Entropy Confounding}
\label{sec:scenario-binaryentropyconf}

This scenario builds on the binary confounding setup from Section~\ref{sec:scenario-binConf}, but is specifically designed to evaluate the \texttt{entropybounds} algorithm. To this end, the data-generating process has been modified in several key ways.

All variables are binary: \(X, Y, U \in \{0,1\}\).

Our goal is to compute bounds for both the Average Treatment Effect (ATE) and the Probability of Necessity and Sufficiency (PNS):
\[
\text{ATE} = P(Y_{X=1} = 1) - P(Y_{X=0} = 1), \qquad
\text{PNS} = P(Y_{X=1} = 1,\; Y_{X=0} = 0)
\]

\input{chapters/ch-simulation_and_experiment/entropyConf/dataGen}
\input{chapters/ch-simulation_and_experiment/entropyConf/results}

\input{chapters/ch-simulation_and_experiment/entropyConf/underspecTheta}

%% file: chapters/ch-simulation_and_experiment/entropyConf/dataGen.tex
\subsection{Data-Generating Process}
\label{entropyConf_dataGen}

The \texttt{BinaryEntropyConf} scenario modifies the binary confounding setup from Section~\ref{binConf_dataGen} along the following three dimensions:

\begin{enumerate}
    \item \textbf{Controlled entropy of the confounder:}  
    Rather than freely sampling the success probability \( p_U \in [0,1] \), we enforce a fixed entropy level \( H_{\mathrm{target}} \in [0,1] \) for the binary confounder \( U \sim \mathrm{Bern}(p_U) \).  
    Specifically, we numerically invert the binary entropy function
    \[
    H(p) = -p \log_2(p) - (1-p) \log_2(1-p)
    \]
    to find a value \( p_U \in [0, 0.5] \) such that \( H(p_U) = H_{\mathrm{target}} \), and then reflect it with 50\% probability\footnote{Since the binary entropy function is symmetric around \( p = 0.5 \), both \( p \) and \( 1 - p \) yield the same entropy. After solving for \( p_U \in [0, 0.5] \), we randomly replace it with \( 1 - p_U \) with 50\% probability to ensure that values are uniformly distributed across the entire interval \( [0, 1] \).} over the interval \( [0,1] \).  
    This ensures that \( U \sim \text{Bern}(p_U) \) has entropy approximately equal to \( H_{\mathrm{target}} \), while maintaining a uniform distribution over entropy levels.
        
    \item \textbf{No heteroskedastic noise:}  
    To isolate the effect of the confounder entropy, we disable observation-level noise:
    \[
    \varepsilon_{X_i} = \varepsilon_{Y_i} = 0 \quad \forall i,
    \]
    making the system fully deterministic once the latent variables are fixed.

    \item \textbf{Grid over entropy levels:}  
    To assess how identifiability varies with the informativeness of the confounder, we simulate datasets across a grid of entropy values:
    \[
    H_{\mathrm{target}} \in \{0.05,\; 0.15,\; 0.25,\; \dots,\; 0.95\}
    \]
    Each level corresponds to a different concentration of \(U\), ranging from nearly deterministic to nearly uniform.
    Given a total of \(N = 2000\) simulations and 10 distinct entropy levels, this yields exactly 200 simulations per level.
    
\end{enumerate}

%% file: chapters/ch-simulation_and_experiment/entropyConf/results.tex
\subsection{Results}

In this scenario, we restrict our evaluation to two algorithms per query:
\begin{itemize}
\item For the ATE, we examine \texttt{ATE\_entropybounds-trueTheta} and \texttt{ATE\_manski}.
\item For the PNS, we consider \texttt{PNS\_entropybounds-trueTheta} and \texttt{PNS\_tianpearl}.
\end{itemize}

All results are compared across a range of confounder entropy levels, denoted by \(H_{\text{target}}\) (see Section~\ref{entropyConf_dataGen}).

\paragraph*{ATE}~

Table~\ref{tab:binaryentropyconf-ate-results-long} shows detailed results for the ATE across ten entropy levels. As before, we highlight the \textbf{Invalid Rate} and \textbf{Net Width} to facilitate comparison.

\begin{table}[H]
\centering
\caption{Results for ATE in the \texttt{BinaryEntropyConf} scenario across varying values of \(H_{\text{target}}\).}
\label{tab:binaryentropyconf-ate-results-long}
\scriptsize
\renewcommand{\arraystretch}{1.1}
\begin{tabularx}{\textwidth}{c l c c c c c}
\toprule
\textbf{$H_{\text{target}}$} & \textbf{Algorithm} & \textbf{Fail Rate} & \textbf{Invalid Rate} & \textbf{Net Width} & \textbf{Bound Width} & \textbf{Invalid $\Delta$} \\
\midrule
\multirow{2}{*}{0.05} & \texttt{entropybounds-trueTheta} & 0.50 & \textbf{50.25} & \textbf{31.00} & 65.84 & 7.68 \\
                      & \texttt{manski}                  & 0.00 & \textbf{1.00}  & \textbf{50.00} & 50.50 & 0.51 \\
\arrayrulecolor{black!20}\midrule
\multirow{2}{*}{0.15} & \texttt{entropybounds-trueTheta} & 0.50 & \textbf{26.63} & \textbf{38.10} & 54.81 & 6.54 \\
                      & \texttt{manski}                  & 0.00 & \textbf{1.00}  & \textbf{50.00} & 50.50 & 0.39 \\
\arrayrulecolor{black!20}\midrule
\multirow{2}{*}{0.25} & \texttt{entropybounds-trueTheta} & 1.00 & \textbf{16.16} & \textbf{43.62} & 53.21 & 6.10 \\
                      & \texttt{manski}                  & 0.00 & \textbf{0.00}  & \textbf{50.00} & 50.00 & N/A  \\
\arrayrulecolor{black!20}\midrule
\multirow{2}{*}{0.35} & \texttt{entropybounds-trueTheta} & 0.50 & \textbf{11.06} & \textbf{46.10} & 52.30 & 8.22 \\
                      & \texttt{manski}                  & 0.00 & \textbf{0.50}  & \textbf{50.00} & 50.25 & 0.56 \\
\arrayrulecolor{black!20}\midrule
\multirow{2}{*}{0.45} & \texttt{entropybounds-trueTheta} & 0.00 & \textbf{9.00}  & \textbf{47.49} & 52.21 & 6.23 \\
                      & \texttt{manski}                  & 0.00 & \textbf{0.00}  & \textbf{50.00} & 50.00 & N/A  \\
\arrayrulecolor{black!20}\midrule
\multirow{2}{*}{0.55} & \texttt{entropybounds-trueTheta} & 0.50 & \textbf{15.58} & \textbf{48.29} & 56.57 & 6.09 \\
                      & \texttt{manski}                  & 0.00 & \textbf{0.00}  & \textbf{50.00} & 50.00 & N/A  \\
\arrayrulecolor{black!20}\midrule
\multirow{2}{*}{0.65} & \texttt{entropybounds-trueTheta} & 0.50 & \textbf{10.05} & \textbf{49.17} & 54.51 & 4.72 \\
                      & \texttt{manski}                  & 0.00 & \textbf{0.50}  & \textbf{50.00} & 50.25 & 0.76 \\
\arrayrulecolor{black!20}\midrule
\multirow{2}{*}{0.75} & \texttt{entropybounds-trueTheta} & 0.00 & \textbf{12.00} & \textbf{49.60} & 55.65 & 5.29 \\
                      & \texttt{manski}                  & 0.00 & \textbf{0.00}  & \textbf{50.00} & 50.00 & N/A  \\
\arrayrulecolor{black!20}\midrule
\multirow{2}{*}{0.85} & \texttt{entropybounds-trueTheta} & 0.00 & \textbf{10.00} & \textbf{49.86} & 54.87 & 7.98 \\
                      & \texttt{manski}                  & 0.00 & \textbf{0.50}  & \textbf{50.00} & 50.25 & 1.52 \\
\arrayrulecolor{black!20}\midrule
\multirow{2}{*}{0.95} & \texttt{entropybounds-trueTheta} & 0.50 & \textbf{10.05} & \textbf{49.98} & 55.23 & 4.09 \\
                      & \texttt{manski}                  & 0.00 & \textbf{0.50}  & \textbf{50.00} & 50.25 & 0.17 \\
\arrayrulecolor{black}
\bottomrule
\end{tabularx}
\end{table}

\begin{figure}[H]
    \centering
    \includegraphics[width=0.8\textwidth]{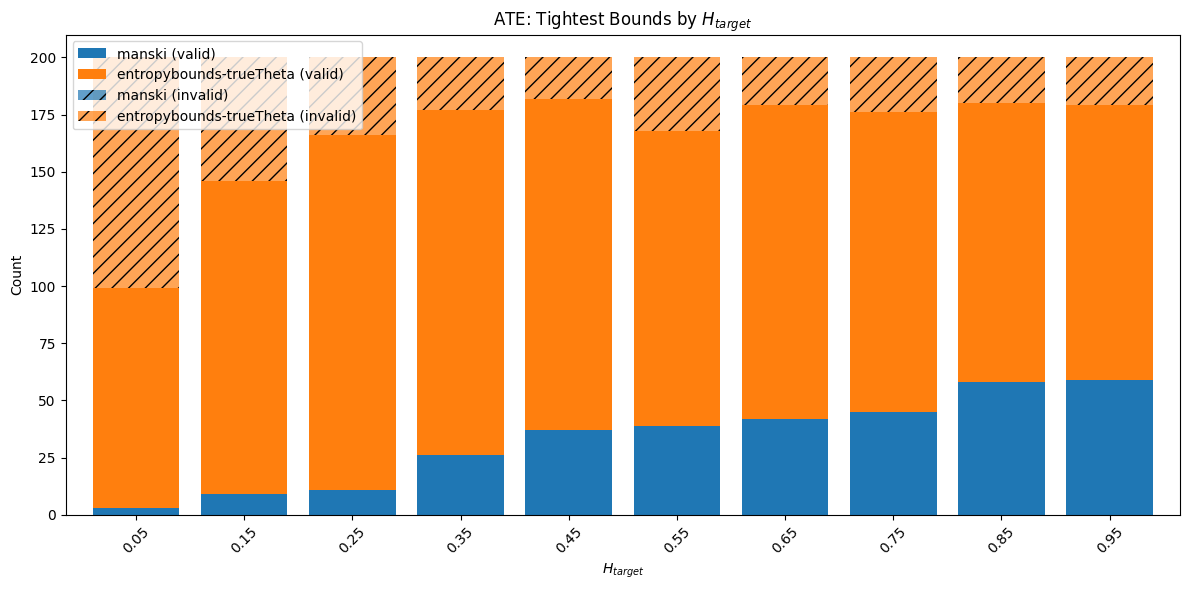}
    \caption{Tightest bounds for the ATE across varying values of \(H_{\text{target}}\) in the \texttt{BinaryEntropyConf} scenario. Dashed segments indicate invalid bounds.}
    \label{fig:ATE_entropybounds_winner}
\end{figure}

Figure~\ref{fig:ATE_entropybounds_winner} visualizes, for each entropy level, the proportion of simulations where each algorithm achieved the tightest bounds. The bars are further split into valid and invalid outcomes, with dashed areas indicating invalid bounds.

\paragraph*{PNS}~

Table~\ref{tab:binaryentropyconf-pns-results-long} shows the same structure of results for the PNS estimand. In contrast to the ATE case, both algorithms yield (almost) fully valid bounds across all entropy levels.

\begin{table}[H]
\centering
\caption{Results for PNS in the \texttt{BinaryEntropyConf} scenario across varying values of \(H_{\text{target}}\).}
\label{tab:binaryentropyconf-pns-results-long}
\scriptsize
\renewcommand{\arraystretch}{1.1}
\begin{tabularx}{\textwidth}{c l c c c c c}
\toprule
\textbf{$H_{\text{target}}$} & \textbf{Algorithm} & \textbf{Fail Rate} & \textbf{Invalid Rate} & \textbf{Net Width} & \textbf{Bound Width} & \textbf{Invalid $\Delta$} \\
\midrule
\multirow{2}{*}{0.05} & \texttt{entropybounds-trueTheta} & 0.00 & \textbf{2.50} & \textbf{36.19} & 37.79 & 2.83 \\
                      & \texttt{tianpearl}               & 0.00 & \textbf{1.00} & \textbf{48.28} & 48.80 & 0.50 \\
\arrayrulecolor{black!20}\midrule
\multirow{2}{*}{0.15} & \texttt{tianpearl}               & 0.00 & \textbf{1.50} & \textbf{49.97} & 50.72 & 1.15 \\
                      & \texttt{entropybounds-trueTheta} & 0.00 & \textbf{2.00} & \textbf{50.34} & 51.34 & 0.81 \\
\arrayrulecolor{black!20}\midrule
\multirow{2}{*}{0.25} & \texttt{tianpearl}               & 0.00 & \textbf{0.00} & \textbf{48.68} & 48.68 & N/A \\
                      & \texttt{entropybounds-trueTheta} & 0.00 & \textbf{0.00} & \textbf{54.16} & 54.16 & N/A \\
\arrayrulecolor{black!20}\midrule
\multirow{2}{*}{0.35} & \texttt{tianpearl}               & 0.00 & \textbf{0.50} & \textbf{50.36} & 50.60 & 1.60 \\
                      & \texttt{entropybounds-trueTheta} & 0.00 & \textbf{1.00} & \textbf{58.97} & 59.38 & 0.80 \\
\arrayrulecolor{black!20}\midrule
\multirow{2}{*}{0.45} & \texttt{tianpearl}               & 0.00 & \textbf{0.00} & \textbf{49.03} & 49.03 & N/A \\
                      & \texttt{entropybounds-trueTheta} & 0.00 & \textbf{1.00} & \textbf{59.60} & 60.00 & 0.00 \\
\arrayrulecolor{black!20}\midrule
\multirow{2}{*}{0.55} & \texttt{tianpearl}               & 0.00 & \textbf{0.50} & \textbf{50.04} & 50.29 & 0.24 \\
                      & \texttt{entropybounds-trueTheta} & 0.00 & \textbf{0.00} & \textbf{65.19} & 65.19 & N/A \\
\arrayrulecolor{black!20}\midrule
\multirow{2}{*}{0.65} & \texttt{tianpearl}               & 0.00 & \textbf{0.00} & \textbf{48.94} & 48.94 & N/A \\
                      & \texttt{entropybounds-trueTheta} & 0.00 & \textbf{0.00} & \textbf{65.14} & 65.14 & N/A \\
\arrayrulecolor{black!20}\midrule
\multirow{2}{*}{0.75} & \texttt{tianpearl}               & 0.00 & \textbf{0.00} & \textbf{49.93} & 49.93 & N/A \\
                      & \texttt{entropybounds-trueTheta} & 0.00 & \textbf{0.00} & \textbf{66.68} & 66.68 & N/A \\
\arrayrulecolor{black!20}\midrule
\multirow{2}{*}{0.85} & \texttt{tianpearl}               & 0.00 & \textbf{0.50} & \textbf{50.00} & 50.25 & 3.15 \\
                      & \texttt{entropybounds-trueTheta} & 0.00 & \textbf{0.50} & \textbf{69.54} & 69.69 & 3.15 \\
\arrayrulecolor{black!20}\midrule
\multirow{2}{*}{0.95} & \texttt{tianpearl}               & 0.00 & \textbf{1.00} & \textbf{49.39} & 49.89 & 0.18 \\
                      & \texttt{entropybounds-trueTheta} & 0.00 & \textbf{0.50} & \textbf{68.35} & 68.51 & 0.00 \\
\arrayrulecolor{black}
\bottomrule
\end{tabularx}
\end{table}

\begin{figure}[H]
    \centering
    \includegraphics[width=0.8\textwidth]{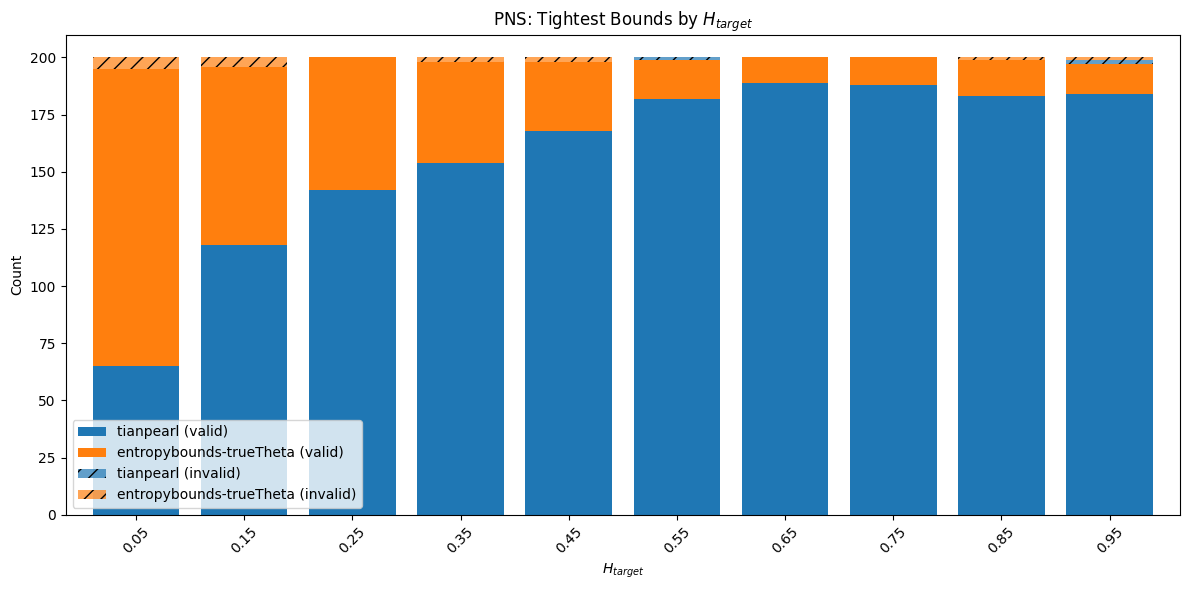}
    \caption{Tightest bounds for the PNS across varying values of \(H_{\text{target}}\) in the \texttt{BinaryEntropyConf} scenario. Dashed segments indicate invalid bounds.}
    \label{fig:PNS_entropybounds_winner}
\end{figure}

Figure~\ref{fig:PNS_entropybounds_winner} summarizes which algorithm achieved the tightest bounds for the PNS across the different confounder entropy levels, again distinguishing between valid and invalid bounds via dashed segments.

%% file: chapters/ch-simulation_and_experiment/entropyConf/underspecTheta.tex
\subsection{Underspecifying $\theta$}
\label{sec:underspec_theta}

We now investigate the consequences of choosing the parameter
\(\theta\)—which should represent an upper bound on the confounder entropy \(H(U)\)
in the \texttt{entropybounds} algorithm—too low, i.e.\ when \(\theta < H(U)\).

To do so, we introduce a dedicated parametrisation of
\texttt{entropybounds}, denoted
\texttt{entropybounds--underspecifyTheta}, which draws
\(\theta \sim \text{Uni}(0, H(U))\) 
for every simulation $\mathcal{S}_j \in \mathcal{D}$ of the
\texttt{BinEntropyConf} scenario.
The degree of misspecification is

\[
\texttt{thetaerror} \;=\; H(U) - \theta \;\;\;(\text{bits}),
\]

so larger values correspond to more severe underspecification.
Our target variable is the indicator
\(\texttt{invalid} \in \{0,1\}\), which equals~1 when the resulting
\texttt{entropybounds} interval fails to cover the ground-truth ATE.
\footnote{We restrict our analysis to the ATE, as \texttt{PNS\_entropybounds--underspecifyTheta} does not provide enough invalid observations.}

\paragraph{Model.}
To quantify the association between \texttt{thetaerror} and the
likelihood of invalid bounds—while adjusting for the true confounder entropy $H(U)$—we fit a logistic regression:

\begin{equation}
  \label{eq:logit_underspec}
  \operatorname{logit}
  \bigl\{P(\texttt{invalid}=1)\bigr\}
  \;=\;
  \beta_0
  + \beta_1\,\texttt{thetaerror}
  + \beta_2\,H(U).
\end{equation}

Adding an interaction term
\(\texttt{thetaerror}\!\times\!H(U)\) did not improve model fit
(\(p = 0.52\)), so the simpler additive specification
\eqref{eq:logit_underspec} is retained for parsimony.

\paragraph{Results.}
Table~\ref{tab:logit_underspec} summarises the maximum-likelihood
estimates. Each additional \emph{0.1\,bit} of underspecification
increases the odds of producing an invalid bound by a factor of
\(\exp(0.1\beta_1) \approx 1.45\) (95\%~CI: 1.24–1.70), while higher
latent entropy substantially decreases the risk. Overall the model
explains about 12\,\% of the deviance (pseudo-\(R^2 = 0.12\)).

\begin{table}[ht]
  \centering
  \caption{Logistic regression results for the probability that 
         \texttt{entropybounds} fails to cover the true ATE when 
         \(\theta\) is underspecified. The table reports 
         maximum-likelihood estimates for each predictor, along with 
         standard errors (SE), Wald \( z \)-statistics, and 
         associated \( p \)-values for testing the null hypothesis 
         that each coefficient equals zero.}
  \label{tab:logit_underspec}
  \sisetup{table-format = 1.3, table-number-alignment = center}
  \begin{tabular}{lllll}
    \toprule
    & {Coef.} & {SE} & {$z$} & {$p$} \\
    \midrule
    Intercept          &  0.188 & 0.098 &  1.92 & 0.054 \\
    \texttt{thetaerror} &  4.187 & 0.404 & 10.37 & \ensuremath{< 0.001}\\
    $H(U)$             & -4.769 & 0.344 & -13.88 & \ensuremath{< 0.001} \\
    \bottomrule
  \end{tabular}
\end{table}

\begin{figure}[H]
    \centering
    \includegraphics[width=0.8\textwidth]{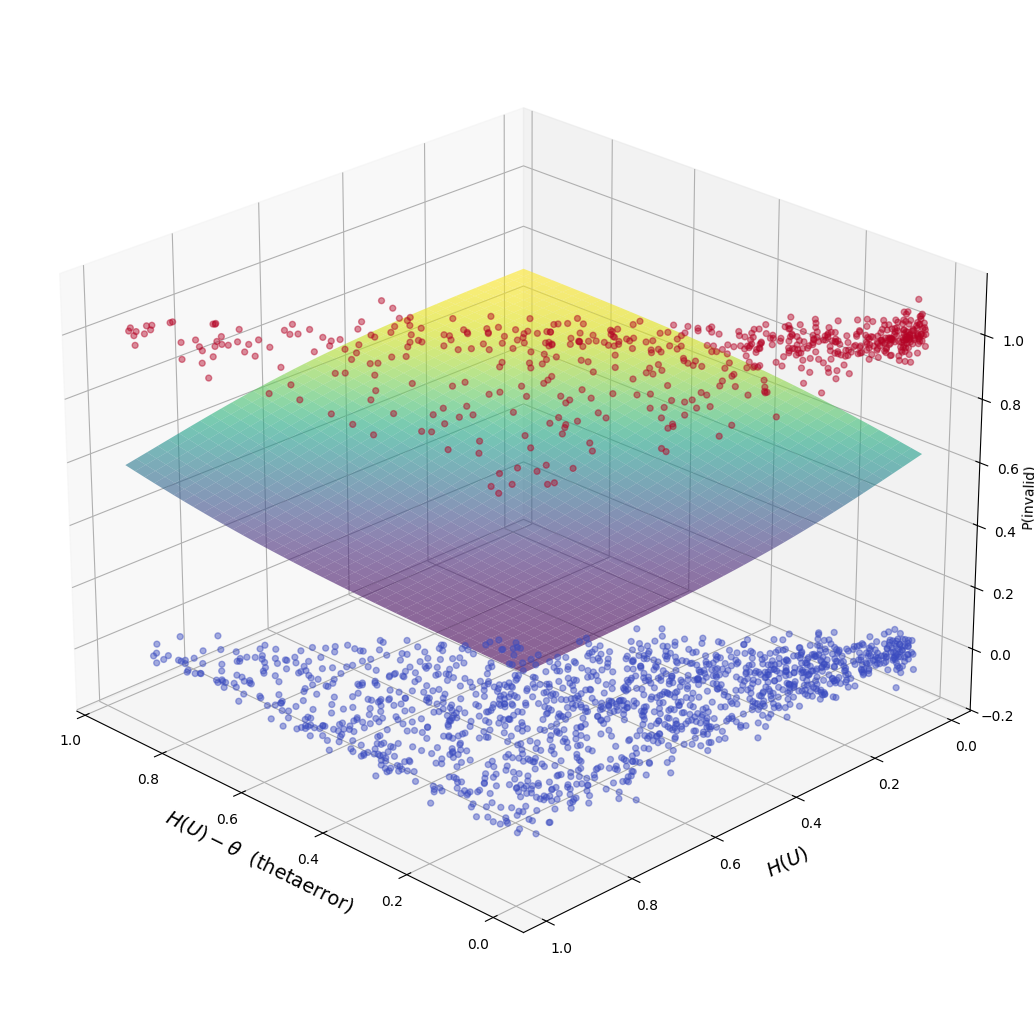}
    \caption{Predicted probability of \texttt{entropybounds} returning invalid bounds as a function of underspecification and true confounder entropy. Red dots represent simulations with invalid bounds (\texttt{invalid} = 1), blue dots represent valid ones. The surface illustrates fitted probabilities from the logistic regression model in Equation~\ref{eq:logit_underspec}.}
    \label{fig:invalid_surface}
\end{figure}

Figure~\ref{fig:invalid_surface} visualizes the fitted model surface along with the raw simulation data. Each dot corresponds to one simulation instance from the \texttt{BinEntropyConf} scenario, with the \(x\)-axis representing \texttt{thetaerror} and the \(y\)-axis the true confounder entropy \(H(U)\). The \(z\)-axis indicates whether the resulting ATE interval was invalid (\(\texttt{invalid} = 1\)) or not. Red dots represent invalid simulations (i.e., where the bound failed), and blue dots valid ones. The colored surface shows the predicted probability of invalidity from the logistic model \eqref{eq:logit_underspec}, increasing sharply in regions of high underspecification and low entropy.

\paragraph{Conclusion.}
The analysis highlights the sensitivity of \texttt{entropybounds} to misspecification of the entropy cap $\theta$. Translating the odds into probabilities, we find that at a moderate underspecification of, for example, \(\texttt{thetaerror} = 0.2\), corresponding to \(\theta = 0.5\) and a true entropy level \(H(U) = 0.7\), the model predicts a failure probability of approximately

\[
\text{logit}^{-1}\bigl(0.188 + 4.187 \cdot 0.2 - 4.769 \cdot 0.7\bigr)
\;\approx\;
\text{logit}^{-1}(-2.313)
\;\approx\;
0.09.
\]

For comparison, if $\theta$ had been correctly specified (\(\texttt{thetaerror} = 0\)), the predicted failure probability would be

\[
\text{logit}^{-1}\bigl(0.188 - 4.769 \cdot 0.7\bigr)
\;\approx\;
\text{logit}^{-1}(-3.141)
\;\approx\;
0.04.
\]

Thus, a modest 0.2-bit underspecification may more than double the risk of not capturing the ground truth—from 4\% to 9\%—despite using the same underlying data. This underscores the importance of choosing $\theta$ conservatively when the true entropy is unknown, particularly if the goal is to compute reliable bounds rather than risk shifting toward a potentially incorrect point estimate.

%% file: chapters/ch-discussion.tex
\chapter{Discussion}
\label{ch-discussion}

This chapter interprets the results presented in Chapter~\ref{ch-simulation_and_experiment}. We begin by analyzing how each algorithm performed across the different simulation scenarios. Next, we attempt to summarize these findings in the form of a decision tree aimed at guiding practitioners.

We then introduce a unified notion of the \emph{best} algorithm for a given simulation, which combines multiple evaluation metrics into a single summary measure. Using this definition, we employ machine-learning techniques to examine whether the best-performing algorithm can be predicted from observable data characteristics.

The chapter concludes with a summary of key insights and a discussion of potential directions for future work.

\input{chapters/ch-discussion/algorithms}

\input{chapters/ch-discussion/generalDecisionTree}

\input{chapters/ch-discussion/bestalg}

\section{Conclusion}

Finally, we summarize our contributions and give an outlook for potential future research.

\subsection{Contributions}
In this thesis, we assembled a diverse suite of bounding algorithms and evaluated their performance across multiple causal scenarios, each generated by a randomized synthetic data-generating process based on structural causal models (SCMs). Performance was assessed with a consistent set of quantitative metrics that captured informativeness, reliability, and error severity.

All implementations were executed in Python. Some algorithms were readily available in this language (\texttt{autobound}, \texttt{ATE\_entropybounds}), while others were implemented by us (\texttt{manski}, \texttt{tianpearl}, \texttt{PNS\_entropybounds}, \texttt{zhangbareinboim}). For methods originally developed in other programming environments, such as \texttt{causaloptim} and \texttt{zaffalonbounds}, we created Python wrappers or re-implementations.

As a result, we developed the \texttt{CausalBoundingEngine} Python package\footnote{\url{https://github.com/tmaringgele/CausalBoundingEngine}}, which enables users to apply a broad range of bounding methods from a unified interface—eliminating the need for extensive preprocessing or tool-specific data formatting.

Additionally, we proposed an extension of the \texttt{entropybounds} algorithm, enabling it to handle counterfactual queries on the third rung of the causal ladder—most notably the Probability of Necessity and Sufficiency (PNS).

To aid practitioners, we distilled the empirical results into a practical decision tree, offering an intuitive guide for algorithm selection based on scenario characteristics. We also trained a Random Forest classifier to predict the empirically best-performing algorithm from easily observable statistics, such as entropy and mutual information of the observed variables.\footnote{Code for both running the classifier and recreating the experiments is available under \\ \url{https://github.com/tmaringgele/causal-bounds}}

\subsection{Potential Future Research}
Our study revealed several promising directions for future work:

\begin{itemize}
    \item While we benchmarked partial identification against classical point estimators like \texttt{OLS} and \texttt{2SLS} using confidence intervals, more sophisticated point estimation techniques should be included in future comparisons for a more balanced evaluation.
    
    \item Recent years have seen the emergence of neural network-based methods for bounding causal effects (e.g., \cite{melnychuk_partial_2024}). Due to their high computational cost, we did not include them in our experiments. It would be valuable to assess their performance relative to traditional methods.
    
    \item Several algorithms in our study (\texttt{autobound}, \texttt{causaloptim}, and asymptotically \texttt{zaffalonbounds}) claim to yield sharp bounds. However, they did not always produce identical results—potentially due to simulation noise or violated assumptions. Further empirical investigation is needed to assess the practical consistency and robustness of these sharpness claims.
    
    \item The EM-based strategy employed by \texttt{zaffalonbounds} showed promise. Future work could explore hybrid methods, such as integrating entropy constraints into the EM framework or combining it with symbolic bounding techniques.

    \item Our version of the \texttt{entropybounds} algorithm, which computes bounds for the PNS, turned out to be less informative than, for example, \texttt{tianpearl}, when the confounder was not \emph{weak}—that is, when its entropy was large. A possible direction for future work may be to modify the program in a way that yields tight bounds even when the weak confounding assumption does not hold.
    
    \item Most existing algorithms are tailored to discrete outcomes. While \texttt{zhangbareinboim} supports continuous outcomes through linear programming, its applicability is restricted to a specific instrumental variable setup. A general-purpose algorithm that bridges \texttt{autobound} and \texttt{zhangbareinboim} could mark a significant step toward automated causal inference in continuous domains.
\end{itemize}

\medskip
\noindent
Taken together, our contributions offer both practical tools and conceptual insights for the field of causal inference under partial identification. By systematically comparing existing methods, proposing novel extensions, and offering actionable guidance to practitioners, this thesis contributes to bridging the gap between theoretical development and empirical application. We hope it encourages further exploration into interpretable, scalable, and general-purpose tools for causal reasoning in the presence of uncertainty.

%% file: chapters/ch-discussion/algorithms.tex
\section{Algorithms}
\label{ch-discussion-sec:algorithms}

In this section, we revisit each algorithm from our evaluation suite and provide a high-level summary of its empirical performance across the different scenarios.

\subsection{\texttt{manski}}\label{sec:findings:manski}

The \texttt{manski} approach behaved as expected in our experiments, producing conservative bounds on the ATE that were almost always valid. In other words, Manski’s minimal-assumption bounds consistently contained the true effect across nearly all simulation runs.

In the \texttt{BinaryConf} scenario, the bounds returned by \texttt{manski} were identical to those of \texttt{autobound} and \texttt{causaloptim}, which is unsurprising given that all three rely on the same underlying assumptions.

In settings with standard confounding and no additional assumptions available, \texttt{manski} may be the preferable choice due to its simplicity, transparency, and ease of computation.

\subsection{\texttt{tianpearl}}\label{sec:findings:tianpearl}

The \texttt{tianpearl} bounds can be seen as the counterpart to the \texttt{manski} bounds for the PNS query. As with \texttt{manski}, the bounds produced by \texttt{tianpearl} were identical to those of \texttt{causaloptim} and \texttt{autobound} in the \texttt{BinaryConf} scenario, reflecting their shared assumption set.

In such cases, the main advantage of \texttt{tianpearl} lies in its use of symbolic computation, which enables clear and interpretable expressions for the resulting bounds.

\subsection{\texttt{OLS} and \texttt{2SLS}}\label{sec:findings:ols-2sls}
As anticipated, the naive Ordinary Least Squares estimator (treating the OLS coefficient’s confidence interval as a “bound”) gave overly narrow intervals that often failed to contain the true ATE. Because OLS assumes no unobserved confounding, its ATE estimates were biased in scenarios with confounders, resulting in a high invalid rate despite the seemingly tight intervals.

The two-stage least squares (2SLS) method, which uses an instrumental variable, also performed poorly in these simulations. Not only were the 2SLS-based intervals relatively wide, but they still had a high rate of invalid bounds. One reason might be that 2SLS rests on a linearity assumption which was violated in our nonlinear data-generating processes. In practice, more sophisticated IV techniques (e.g., those allowing nonlinear relationships) might improve performance in such settings.

\subsection{\texttt{causaloptim}}\label{sec:findings:causaloptim}

In the \texttt{BinaryConf} scenario, \texttt{causaloptim} proved reliable: it consistently produced valid (or near-valid) bounds for both the ATE and the PNS.  
By contrast, when an instrument was introduced in the \texttt{BinaryIV} scenario, the method became markedly less dependable—especially for the PNS.

However, a key strength of \texttt{causaloptim}, compared to other optimization-based methods, is its ability to derive symbolic bounds. This provides analytical insight into how the bounds are constructed and enhances reusability: if parts of the observational data change or are yet to be collected, the symbolic expressions can simply be re-evaluated without rerunning the entire procedure.

\subsection{\texttt{autobound}}\label{sec:findings:autobound}

The \texttt{autobound} algorithm displayed identical performance to \texttt{causaloptim} in the \texttt{BinaryConf} scenario, which is likely due to their shared theoretical foundations for sharp bounding.  
Unlike \texttt{causaloptim}, however, \texttt{autobound} remained reliable across all binary scenarios, and even in continuous settings—where we applied it to binarized data—it showed a lower invalid rate than most competing algorithms.

These results, coupled with its broad applicability, make \texttt{autobound} one of the most promising candidates in our study.

\subsection{\texttt{entropybounds}}

Our analysis of \texttt{entropybounds} focuses primarily on the \texttt{BinaryEntropyConf} scenario (see Section~\ref{sec:scenario-binaryentropyconf}), which stratifies simulations into subgroups with varying levels of confounding, indexed by the target entropy \( H_{\text{target}} \).

As expected, in strata with \emph{weaker} confounding (i.e., smaller \( H_{\text{target}} \)), both the ATE and PNS variants of \texttt{entropybounds}, when given the optimal parameter choice \(\theta = H(U)\), produced on average tighter bounds than the benchmark algorithms \texttt{manski} and \texttt{tianpearl}.

However, this improvement came at a cost in the case of the ATE variant (i.e., \texttt{ATE\_entropybounds--trueTheta}): the invalid rate increased substantially. This trade-off was less pronounced for \texttt{PNS\_entropybounds}, which generally yielded more reliable bounds. Still, this observation should be interpreted with caution, as PNS bounds are inherently wider and thus more conservative.

When the parameter \(\theta\) is chosen heuristically—especially when underestimated—\texttt{entropybounds} continued to produce relatively narrow bounds. Yet this narrowing appears to reflect a controlled trade-off: smaller values of \(\theta\) reduce the feasible set and thus the bound width, but increase the risk of invalid estimates. We explored this relationship in more detail in Section~\ref{sec:underspec_theta}.

Overall, our simulations suggest that \texttt{entropybounds} should be applied with care. While entropy-based constraints can tighten bounds considerably, they also raise the likelihood of invalid inference—particularly when the assumptions do not align with the true data-generating process.

Finally, we emphasize that our evaluation is based on SCM-driven simulations, where structural parameters and causal mechanisms are explicitly modeled. This differs from the approach used in the original work, which—like most algorithm papers considered in this thesis—validates performance by sampling directly from compatible joint distributions. Determining which approach better reflects real-world data-generating processes is an open and important question, but one that lies beyond the scope of this thesis.

\subsection{\texttt{zaffalonbounds}}\label{sec:findings:zaffalonbounds}

Among all methods evaluated, \texttt{zaffalonbounds} was the top performer in the binary-outcome scenarios, consistently yielding the tightest bounds for both the ATE and the PNS.

Although its invalid rate was slightly higher than that of \texttt{autobound}, the magnitude of each miss was small—that is, it exhibited a low invalid~$\Delta$. This indicates that its expectation–maximization (EM) strategy can offer practical advantages over more traditional bounding techniques.

Compared with the other algorithms, \texttt{zaffalonbounds} required noticeably longer runtimes. Its computational cost, however, is tunable through hyperparameters such as the number of EM restarts and the maximum number of iterations.

A potential limitation is that, unlike \texttt{autobound} or (in our settings) \texttt{causaloptim}, the bounds produced by \texttt{zaffalonbounds} are not theoretically guaranteed to be sharp. Instead, they represent approximations to the sharp bounds. Nevertheless, both prior work and our simulation results suggest that, given a sufficient number of EM restarts, the method can closely approximate sharp solutions in practice~\cite{zaffalon_causal_2021}.

\subsection{\texttt{zhangbareinboim}}\label{sec:findings:zhangbareinboim}

The \texttt{zhangbareinboim} algorithm was only applicable in the \texttt{ContIV} scenario, where it targets the ATE in the presence of continuous outcomes and an instrumental variable. In this setting, it performed comparably well in terms of validity, achieving one of the lowest invalid rates. While its bounds were not the tightest, they were among the most reliable.

Overall, when dealing with continuous outcomes and a valid instrument, \texttt{zhangbareinboim} appears to offer a principled option. However, due to the limited number of algorithms directly applicable in this specific setting, we are unable to fully contextualize its performance against direct competitors.

%% file: chapters/ch-discussion/generalDecisionTree.tex
\section{Decision Tree}

The results of our experiments demonstrate that partial identification is not a binary concept but a continuum—especially when violated assumptions (such as random noise in the data-generating process) come into play. While some algorithms consistently performed well across all metrics in specific scenarios, there was often a clear trade-off between the informativeness of the resulting bounds and their reliability. In particular, narrower bounds tended to come with a higher risk of being invalid.

To help practitioners navigate this landscape, we now distill our findings into a practical decision aid: a decision tree. This tree (Figure~\ref{fig:decision-tree-general}) guides the user in selecting a suitable bounding algorithm based on the characteristics of their problem and their tolerance for uncertainty.

At several points in the tree, the user must decide how conservative they want the bounds to be. We distinguish between three levels of conservativeness, indicated by the following symbols:
\begin{itemize}
    \item \textcolor{blue}{\(\boldsymbol{\uparrow}\)}: \textbf{Conservative} — safe but wide bounds, with strong theoretical guarantees
    \item \textcolor{black}{\(\boldsymbol{-}\)}: \textbf{Moderate} — a balanced trade-off between tightness and robustness
    \item \textcolor{red}{\(\boldsymbol{\downarrow}\)}: \textbf{Aggressive} — tighter bounds, but at a higher risk of invalidity
\end{itemize}

We briefly describe the meaning of the other decision points used in the tree:

\begin{itemize}
    \item \textbf{Outcome:} Indicates whether the outcome variable is binary or continuous. In continuous settings, values lie in the interval \([0, 1]\), as in the \texttt{ContConf} and \texttt{ContIV} scenarios.

    \item \textbf{Query:} Specifies the causal estimand of interest—such as the Average Treatment Effect (ATE) or the Probability of Necessity and Sufficiency (PNS)—which determines the type of causal relationship to be bounded.

    \item \textbf{Instrument:} Denotes whether an exogenous instrumental variable is available; that is, a variable that affects the treatment but has no direct effect on the outcome. This applies to scenarios like \texttt{BinaryIV} and \texttt{ContIV}.

    \item \boldmath\( H(U) \leq \theta \)\unboldmath: Indicates whether an upper bound on the entropy of the unobserved confounder \(U\) is known or assumed.

    \item \textbf{Symbolic bounds:} Reflects whether the user requires symbolic expressions for the bounds (as produced by methods like \texttt{causaloptim} or \texttt{tianpearl}), or whether numeric estimates are sufficient.
\end{itemize}

\afterpage{\clearpage}
\begin{sidewaysfigure}
\centering
\begin{adjustbox}{width=\textwidth,center}
\begin{forest} mytree
[Outcome
  [Query, edge label={node[font=\Huge,midway,left,yshift=10pt]{binary}}%
    [Instrument, edge label={node[font=\Huge,midway,left,yshift=10pt]{ATE}}%
      [Conservativeness, edge label={node[font=\Huge,midway,left,yshift=10pt]{unavailable}}%
        [Use Confidence Intervals, edge label={node[font=\Huge,midway,right]{\textcolor{red}{\(\boldsymbol{\downarrow}\)}}}]%
        [\texttt{zaffalonbounds}, edge label={node[font=\Huge,midway,right,yshift=15pt,xshift=5pt]{\textcolor{black}{\(\boldsymbol{-}\)}}}]%
        [Symbolic bounds, edge label={node[font=\Huge,midway,left,yshift=-10pt]{\textcolor{blue}{\(\boldsymbol{\uparrow}\)}}}%
          [\texttt{autobound}, edge label={node[font=\Huge,midway,left,yshift=10pt]{don't care}}]%
          [\texttt{manski}, edge label={node[font=\Huge,midway,right,yshift=10pt]{wanted}}]%
        ]%
      ]%
      [Conservativeness, edge label={node[font=\Huge,midway,left,yshift=-10pt]{available}}%
        [\texttt{zaffalonbounds}, edge label={node[font=\Huge,midway,left,yshift=10pt]{\textcolor{black}{\(\boldsymbol{-}\)}}}]%
        [\texttt{autobound}, edge label={node[font=\Huge,midway,right,yshift=15pt,xshift=5pt]{\textcolor{blue}{\(\boldsymbol{\uparrow}\)}}}]%
      ]%
    ]%
    [Instrument, edge label={node[font=\Huge,midway,right,yshift=10pt]{PNS}}%
      [Conservativeness, edge label={node[font=\Huge,midway,left,yshift=10pt]{unavailable}}%
        [$H(U)\leq \theta$, edge label={node[font=\Huge,midway,left,yshift=10pt]{\textcolor{black}{\(\boldsymbol{-}\)}}}%
          [\texttt{zaffalonbounds}, edge label={node[font=\Huge,midway,left,yshift=10pt]{unknown}}]%
          [\texttt{entropybounds}, edge label={node[font=\Huge,midway,right,yshift=10pt]{known}}]%
        ]%
        [Symbolic bounds, edge label={node[font=\Huge,midway,left,yshift=-10pt]{\textcolor{blue}{\(\boldsymbol{\uparrow}\)}}}%
          [\texttt{autobound}, edge label={node[font=\Huge,midway,left,yshift=10pt]{don't care}}]%
          [\texttt{tianpearl} or \texttt{causaloptim}, edge label={node[font=\Huge,midway,right,yshift=10pt]{wanted}}]%
        ]%
      ]%
      [Conservativeness, edge label={node[font=\Huge,midway,left,yshift=-10pt]{available}}%
        [\texttt{zaffalonbounds}, edge label={node[font=\Huge,midway,left,yshift=10pt]{\textcolor{black}{\(\boldsymbol{-}\)}}}]%
        [\texttt{tianpearl}, edge label={node[font=\Huge,midway,right,yshift=15pt,xshift=5pt]{\textcolor{blue}{\(\boldsymbol{\uparrow}\)}}}]%
      ]%
    ]%
  ]%
  [Query, edge label={node[font=\Huge,midway,right,yshift=10pt]{continuous}}%
    [Instrument, edge label={node[font=\Huge,midway,right]{ATE}}%
      [\texttt{autobound} on binarized data, edge label={node[font=\Huge,midway,left,yshift=10pt]{unavailable}}]%
      [\texttt{zhangbareinboim}, edge label={node[font=\Huge,midway,right,yshift=10pt]{available}}]%
    ]%
  ]%
]
\end{forest}
\end{adjustbox}
\caption{Decision tree for selecting a bounding algorithm based on problem characteristics and desired level of conservativeness. The tree covers both binary and continuous settings, and guides the user through key decision points such as the target query, availability of an instrumental variable, and whether symbolic bounds are preferred. Color-coded arrows indicate the conservativeness of the returned bounds.}
\label{fig:decision-tree-general}
\end{sidewaysfigure}

%% file: chapters/ch-discussion/bestalg.tex
\section{The \textit{Best} Algorithm}
\label{sec:best-alg}

\subsection{Definition}
We now aim to further compress the empirical results from our simulations into a practical guide for selecting the best bounding algorithm in a given setting.

To do so, we define what it means for an algorithm to be the \textit{best} for a specific simulation instance \(\mathcal{S}_j \in \mathcal{D}\) (see Section~\ref{sec:simulation-framework-and-notation} for notation). Our definition balances two criteria: bound validity and tightness.

\begin{enumerate}
    \item Let \( \mathcal{A}'_j \subseteq \mathcal{A} \) be the subset of algorithms that either return valid bounds on \(\mathcal{S}_j\), or—if invalid—exhibit an Invalid \(\Delta\) below 1\%.
    \item Among these, select the algorithm with the smallest bound width.
\end{enumerate}

Formally, let \(a \in \mathcal{A}\) return bounds \((\texttt{lower}_a(j), \texttt{upper}_a(j))\) on simulation \(\mathcal{S}_j\). The best algorithm for \(\mathcal{S}_j\) is therefore given by:
\[
\arg\min_{a \in \mathcal{A}'_j} ( \texttt{upper}_a(j) - \texttt{lower}_a(j))
\]

Table~\ref{tab:best_algs} lists the algorithms that were selected as the \emph{best} most frequently for a given query across the $N=2000$ simulations in each scenario.

\begin{table}[H]
  \centering
  \begin{tabular}{|c|c|c|c|}
    \hline
    Scenario & Query & \emph{Best} Algorithm & How often \\ \hline
    binConf   & ATE   & OLS-0.95   & 1709   \\ \hline
    binIV   & ATE   & zaffalonbounds   & 755   \\ \hline
    contConf   & ATE  & entropybounds-0.10--binned  & 806  \\ \hline
    contIV & ATE & 2SLS-0.95 & 736 \\ \hline
    binConf   & PNS   & zaffalonbounds & 1194   \\ \hline
    binIV   & PNS   & zaffalonbounds   & 725   \\ \hline

  \end{tabular}
    \caption{Most frequently selected \emph{best} algorithm per scenario and query. Out of $N=2000$ simulations each.}
  \label{tab:best_algs}
\end{table}

\subsection{Prediction}

We now investigate whether the \emph{best} algorithm for a given dataset can be predicted \emph{a priori}.
To this end, we train a Random Forest classifier and compare its accuracy to that of a simple \emph{null model}, which always selects the algorithm most frequently identified as best for the given scenario–query combination (see Table~\ref{tab:best_algs}).\footnote{For simplicity, we restrict our analysis to scenarios with binary outcomes.}

\vspace{0.3em}\noindent
\textbf{Features.}  Only statistics that are observable from the data are
used:
\begin{itemize}\setlength{\itemsep}{2pt}
    \item Entropies of the observed variables: $H(Z),\,H(X),\,H(Y)$.
    \item Pairwise mutual informations: $I(Z;X),\,I(Z;Y),\,I(X;Y)$.
\end{itemize}
If no instrument $Z$ is present, the corresponding features are omitted.
The analysis is restricted to the two binary scenarios.  Table
\ref{tab:prediction-accuracies} summarises the out-of-sample performance.

\begin{table}[H]
  \centering
  \caption{Classification accuracy of the Random-Forest meta-selector
  versus the no-information (null) baseline.}
  \label{tab:prediction-accuracies}
  \begin{tabular}{lccc}
    \toprule
    Scenario & Query & Null acc. & RF acc. \\
    \midrule
    \texttt{BinaryConf} & ATE & 0.85 & \textbf{0.90} \\
    \texttt{BinaryConf} & PNS & 0.60 & \textbf{0.81} \\
    \texttt{BinaryIV}   & ATE & 0.38 & \textbf{0.63} \\
    \texttt{BinaryIV}   & PNS & 0.36 & \textbf{0.63} \\
    \bottomrule
  \end{tabular}
\end{table}

\noindent
The Random Forest consistently outperforms the null baseline, increasing accuracy by up to $0.27$ depending on the scenario.

We now take a closer look at the feature importance (SHAP values) for two selected scenario-query combinations.

\paragraph{BinaryIV, ATE target.}
Table~\ref{tab:shap_binaryiv_ate} displays signed mean SHAP values for the
two dominant winners.%
\footnote{Positive numbers indicate that large feature values increase the
probability of selecting the algorithm; negative numbers have the opposite
effect.}

\begin{table}[H]
\centering
\caption{Signed mean SHAP values (BinaryIV, ATE).}
\label{tab:shap_binaryiv_ate}
\begin{tabular}{lrrrrrr}
\toprule
Algorithm & $H(Z)$ & $H(X)$ & $H(Y)$ & $I(Z;X)$ & $I(Z;Y)$ & $I(X;Y)$\\
\midrule
\texttt{zaffalonbounds} &  0.0006 & \textbf{\phantom{$-$}--0.0087} & --0.0014 & --0.0008 & --0.0003 &  0.0004\\
\texttt{manski}      & \textbf{--0.0034} &  0.0015 &  0.0036 & --0.0015 & \textbf{--0.0033} & --0.0105\\
\bottomrule
\end{tabular}
\end{table}

\noindent
\textbf{Interpretation.}  
\texttt{zaffalonbounds} is favoured when the instrument is informative and the
treatment is \emph{skewed} (low $H(X)$); weak or noisy instruments (low
$H(Z)$, low $I(Z;X)$) swing the odds towards \texttt{manski}, which does
not rely on IV assumptions. The classifier therefore tells us that if the IV assumptions are possibly violated, it may be better to drop the instrument entirely.

\paragraph{BinaryConf, PNS target.}
In this scenario, all algorithms utilize the same causal structure. The signed mean SHAP values are reported in Table~\ref{tab:shap_binaryconf_pns}.

\begin{table}[H]
\centering
\caption{Signed mean SHAP values (BinaryConf, PNS).}
\label{tab:shap_binaryconf_pns}
\begin{tabular}{lrrr}
\toprule
Algorithm & $H(X)$ & $H(Y)$ & $I(X;Y)$\\
\midrule
\texttt{zaffalonbounds}            &  \textbf{\phantom{$-$}0.0148} & 0.0005  & 0.0004\\
\texttt{entropybounds-0.10}        & \textbf{--0.0145} & --0.0026 & --0.0037\\
\texttt{entropybounds-randomTheta} & --0.0007 & --0.0001 &  0.0027\\
\texttt{entropybounds-trueTheta}   &  \phantom{$-$}0.0003 & 0.0023  & 0.0006\\
\texttt{causaloptim}               &  \phantom{$-$}0.0003 & 0.0000  & --0.0001\\
\texttt{autobound}                 & --0.0001 & --0.0001 &  0.0000\\
\bottomrule
\end{tabular}
\end{table}

\noindent\textbf{Interpretation.}  
Balanced treatment ($H(X)$ high) points to \texttt{zaffalonbounds}; a
skewed treatment, maybe implying low confounder entropy, favours the 
\texttt{entropybounds} variants.

\paragraph{Conclusion.}
Across all binary scenarios, the meta-selector outperforms the null baseline. The SHAP analysis suggests that a weak instrument may be more detrimental than having no instrument at all. While the best algorithm—according to our definition—can be predicted from observable data in our simulations, one should be very cautious in generalizing these results beyond the synthetic settings studied here.

%% file: chapters/appendix.tex
\appendix

\chapter{Detailed Runtimes}
\label{appendix:runtimes}

This appendix provides the full runtime details for each algorithm, broken down by causal query, scenario, and parameter setting. All runtimes are measured in seconds and correspond to executions on a \texttt{n1-standard-4} Google Compute Engine VM (4 vCPUs, 15~GB RAM). These values complement the averaged results reported in Section~\ref{sec:runtimes}.

\begin{table}[H]
  \centering
  \caption{Detailed per-scenario runtimes (in seconds) for each algorithm variant. The final column reports the average across all applicable scenarios.}
  \label{tab:detailed-runtimes}
  \begin{tabular}{|l|c|c|c|c|c|}
\hline
\textbf{Algorithm} & \texttt{BinaryConf} & \texttt{BinaryIV} & \texttt{ContConf} & \texttt{ContIV} & \textbf{Average} \\
\hline
\texttt{ATE\_2SLS-0.95} &  & 20 &  & 21 & \textbf{20} \\ \hline
\texttt{ATE\_2SLS-0.98} &  & 20 &  & 21 & \textbf{20} \\ \hline
\texttt{ATE\_2SLS-0.99} &  & 20 &  & 21 & \textbf{20} \\ \hline
\texttt{ATE\_OLS-0.95} & 6 &  & 6 &  & \textbf{6} \\ \hline
\texttt{ATE\_OLS-0.98} & 6 &  & 6 &  & \textbf{6} \\ \hline
\texttt{ATE\_OLS-0.99} & 6 &  & 6 &  & \textbf{6} \\ \hline
\texttt{ATE\_autobound} & 74 & 96 &  &  & \textbf{85} \\ \hline
\texttt{ATE\_autobound--binned} &  &  & 76 & 87 & \textbf{82} \\ \hline
\texttt{ATE\_causaloptim} & 238 & 532 &  &  & \textbf{385} \\ \hline
\texttt{ATE\_causaloptim--binned} &  &  & 327 & 576 & \textbf{451} \\ \hline
\texttt{ATE\_entropybounds-0.10} & 91 & 87 &  &  & \textbf{89} \\ \hline
\texttt{ATE\_entropybounds-0.10--binned} &  &  & 171 & 174 & \textbf{172} \\ \hline
\texttt{ATE\_entropybounds-0.20} & 87 & 84 &  &  & \textbf{85} \\ \hline
\texttt{ATE\_entropybounds-0.20--binned} &  &  & 152 & 154 & \textbf{153} \\ \hline
\texttt{ATE\_entropybounds-0.80} & 71 & 67 &  &  & \textbf{69} \\ \hline
\texttt{ATE\_entropybounds-0.80--binned} &  &  & 93 & 95 & \textbf{94} \\ \hline
\texttt{ATE\_entropybounds-randomTheta} & 78 & 73 &  &  & \textbf{76} \\ \hline
\texttt{ATE\_entropybounds-trueTheta} & 73 & 75 &  &  & \textbf{74} \\ \hline
\texttt{ATE\_manski} & 1 & 1 &  &  & \textbf{1} \\ \hline
\texttt{ATE\_zaffalonbounds} & 1919 & 6984 &  &  & \textbf{4452} \\ \hline
\texttt{ATE\_zaffalonbounds--binned} &  &  & 1718 & 6139 & \textbf{3929} \\ \hline
\texttt{ATE\_zhangbareinboim} &  &  &  & 63 & \textbf{63} \\ \hline
\texttt{PNS\_autobound} & 75 & 95 &  &  & \textbf{85} \\ \hline
\texttt{PNS\_causaloptim} & 269 & 455 &  &  & \textbf{362} \\ \hline
\texttt{PNS\_entropybounds-0.10} & 51 & 49 &  &  & \textbf{50} \\ \hline
\texttt{PNS\_entropybounds-0.20} & 48 & 45 &  &  & \textbf{46} \\ \hline
\texttt{PNS\_entropybounds-0.80} & 38 & 36 &  &  & \textbf{37} \\ \hline
\texttt{PNS\_entropybounds-randomTheta} & 45 & 41 &  &  & \textbf{43} \\ \hline
\texttt{PNS\_entropybounds-trueTheta} & 40 & 39 &  &  & \textbf{40} \\ \hline
\texttt{PNS\_tianpearl} & 1 & 1 &  &  & \textbf{1} \\ \hline
\texttt{PNS\_zaffalonbounds} & 1918 & 7068 &  &  & \textbf{4493} \\ \hline
  \end{tabular}
\end{table}